\documentclass[conference]{IEEEtran}
\IEEEoverridecommandlockouts
\usepackage{cite}
\usepackage{amsmath,amssymb,amsfonts}
\usepackage{algorithmic}
\usepackage{graphicx}
\usepackage{textcomp}
\usepackage{xcolor}
\usepackage{multirow} 
\def\BibTeX{{\rm B\kern-.05em{\sc i\kern-.025em b}\kern-.08em
    T\kern-.1667em\lower.7ex\hbox{E}\kern-.125emX}}
\begin{document}

\title{Exploiting Fairness to Enhance Sensitive Attributes Reconstruction}

\author{\IEEEauthorblockN{1\textsuperscript{st} Julien Ferry}
\IEEEauthorblockA{
\textit{LAAS-CNRS, Universit\'{e} de Toulouse, CNRS}\\
Toulouse, France \\
jferry@laas.fr}
\and
\IEEEauthorblockN{2\textsuperscript{nd} Ulrich A{\"i}vodji}
\IEEEauthorblockA{\textit{\'Ecole de Technologie Sup\'erieure} \\
Montr\'eal, Canada \\
Ulrich.Aivodji@etsmtl.ca}
\and
\IEEEauthorblockN{3\textsuperscript{rd} S\'ebastien Gambs}
\IEEEauthorblockA{\textit{Universit\'e du Qu\'ebec \`a Montr\'eal} \\
Montr\'eal, Canada \\
gambs.sebastien@uqam.ca}
\and
\IEEEauthorblockN{4\textsuperscript{th} Marie-Jos\'e Huguet}
\IEEEauthorblockA{
\textit{LAAS-CNRS, Universit\'{e} de Toulouse, CNRS, INSA}\\
Toulouse, France \\
huguet@laas.fr}
\and
\IEEEauthorblockN{5\textsuperscript{th} Mohamed Siala}
\IEEEauthorblockA{
\textit{LAAS-CNRS, Universit\'{e} de Toulouse, CNRS, INSA}\\
Toulouse, France \\
msiala@laas.fr}
}

\def\dataset{{D}}
\def\groundtruth{{Y}}
\def\onegroundtruth{{y}}
\def\predictions{\hat{Y}}
\def\oneprediction{\hat{y}}
\def\sensattr{{S}}
\def\onesensattr{{s}}
\def\unsensattr{{X}}
\def\oneunsensattr{{x}}
\def\example{{e}}

\def\jointdistribution{\mathcal{D}}
\def\groundtruthdistribution{\mathcal{Y}}
\def\sensattrdistribution{\mathcal{S}}
\def\unsensattrdistribution{\mathcal{X}}
\def\unsensattrdistribution#1{\mathcal{X}_{#1}}

\def\learningalgo{\mathcal{L}}
\def\classifier{h}

\def\nattributes{M}
\def\nexamples{N}

\def\tol{\epsilon}

\def\attacker#1{\mathcal{1}_{#1}}
\def\confidence{p}
\def\variables{{\onesensattr{}}^*}
\def\attackersix{\mathcal{A}}
\def\attackerseven{\mathcal{A'}}
\def\confidencevector{P}

\def\generalmodel{\mathcal{RC}(\hat{\sensattr{}}, \confidencevector{},\predictions{},\tol)}
\def\efficientmodel{\mathcal{RC}_\mathcal{E}(\hat{\sensattr{}}, \confidencevector{},\predictions{},\tol)}


\def\mjo#1{\textcolor{blue}{#1}}
\def\mo#1{\textcolor{red}{#1}}
\def\ulr#1{\textcolor{violet}{#1}}
\def\seb#1{\textcolor{cyan}{#1}}
\def\jul#1{\textcolor{orange}{#1}} 

\maketitle

\begin{abstract}
In recent years, a growing body of work has emerged on how to learn machine learning models under fairness constraints, often expressed with respect to some \emph{sensitive attributes}. 
In this work, we consider the setting in which an adversary has black-box access to a target model and show that information about this model's fairness can be exploited by the adversary to enhance his reconstruction of the sensitive attributes of the training data.
More precisely, we propose a generic reconstruction correction method, which takes as input an initial guess made by the adversary and corrects it to comply with some user-defined constraints (such as the fairness information) while minimizing the changes in the adversary's guess.
The proposed method is agnostic to the type of target model, the fairness-aware learning method as well as the auxiliary knowledge of the adversary.
To assess the applicability of our approach, we have conducted a thorough experimental evaluation on two state-of-the-art fair learning methods, using four different fairness metrics with a wide range of tolerances and with three datasets of diverse sizes and sensitive attributes.
The experimental results demonstrate the effectiveness of the proposed approach to improve the reconstruction of the sensitive attributes of the training set.
\end{abstract}


\section{Introduction}
The growing use of machine learning models in high-stakes decision-making raises several ethical issues such as the risk of discrimination. 
To address this issue, a growing body of work has emerged on how to learn machine learning models under fairness constraints, often expressed with respect to some \emph{sensitive attributes}~\cite{barocas-hardt-narayanan,caton2020fairness,DBLP:journals/csur/MehrabiMSLG21}.
These \emph{sensitive attributes} correspond to characteristics such as gender, age or race~\cite{DBLP:conf/nips/DingHMS21}, which should not be taken into account in decision-making processes impacting individuals, for legal, ethical, social or philosophical reasons~\cite{barocas-hardt-narayanan}. 
While fair models usually do not use such sensitive attributes at inference time to avoid disparate treatment~\cite{10.2307/24758720}, they still require access to them at training time~\cite{DBLP:journals/ail/ZliobaiteC16}.
The fact that these models are learnt with the objective to meet specific constraints regarding these sensitive attributes indicates that fair models intrinsically contain information about them.

Another fundamental aspect of responsible machine learning is the protection of privacy. 
Indeed, machine learning models are often trained on large amounts of personal data. 
Here, the main challenge is ensuring that these models learn useful generic patterns without leaking private information about individuals. 
In this context, \emph{inference attacks}~\cite{doi:10.1146/annurev-statistics-060116-054123,DBLP:journals/corr/abs-2007-07646,DBLP:journals/corr/abs-2005-08679} aim at leveraging the output of a computation (\emph{e.g.}, a trained model) to retrieve information regarding its inputs (\emph{e.g.}, a training dataset). 
Our work belongs to the category of \emph{dataset reconstruction attacks}, in which an adversary tries to recover part of a model's training data~\cite{DBLP:journals/corr/abs-2005-08679}. 
More precisely, we study the setting in which an adversary aims at retrieving the entire column of sensitive attributes of the training set.

Depending on the available \emph{auxiliary knowledge}, several strategies can be adopted by an adversary to reconstruct the sensitive attributes of the training set. 
The proposed approach is a post-processing method that we coin as \emph{reconstruction correction}, which takes as input an initial reconstruction performed by an adversary, optionally associated with confidence scores for each guess. 
The reconstruction correction method then minimally updates the adversary's initial guess to satisfy some user-defined constraints. 
Our work focuses on the scenario in which these are fairness constraints and the adversary leverages the fact that a model is known to be fair to improve his initial reconstruction. 
Such \emph{fairness information} can for instance be the results of legal requirements, such as the ``$80$ percent rule" for Statistical Parity~\cite{DBLP:conf/kdd/FeldmanFMSV15} stated by the US Equal Employment Opportunity Commission (EEOC)~\cite{uniformguidelinesemployeeselection}.

The tensions between fairness and privacy in machine learning have been studied in recent years, mainly through the theoretical~\cite{10.1145/3314183.3323847,1548832} and technical~\cite{DBLP:conf/nips/BagdasaryanPS19,DBLP:conf/eurosp/ChangS21,DBLP:journals/corr/abs-2202-08187} conflicts existing between statistical fairness metrics and Differential Privacy (DP). 
For instance, it was proved theoretically impossible to learn models under fairness constraints while respecting DP~\cite{10.1145/3314183.3323847,1548832}. 
Furthermore, DP was shown to have unfair effects on the model's performances~\cite{DBLP:conf/nips/BagdasaryanPS19} and it was observed that fairness led to an increased privacy risk~\cite{DBLP:conf/eurosp/ChangS21}.
We refer the interested reader to a recent survey~\cite{DBLP:journals/corr/abs-2202-08187} summarizing the different causes and consequences of this conflict. 
Our work takes a different direction but also demonstrates that 
enforcing statistical fairness can endanger the privacy of sensitive attributes.
More precisely, our contributions are as follows:
\begin{itemize}
    \item We propose a novel reconstruction attack pipeline, in which a \emph{reconstruction correction} is applied as post-processing to an initial adversary's guess to enforce some user-defined constraints (\emph{e.g.}, fairness constraints).
    \item We show that declarative programming approaches can be applied to implement a generic reconstruction correction. 
    The proposed integer programming model includes statistical fairness constraints but is general enough to also work for a wide range of user-defined constraints.
    \item We derive an efficient reconstruction correction model 
    with polynomial search space, suitable to formulate any rate constraints (such as statistical fairness constraints).
    \item We empirically demonstrate the effectiveness of the proposed reconstruction correction method for two fairness-enhancing techniques that intervene at different stages of the learning pipeline, three datasets with diverse characteristics and sensitive attributes, four statistical fairness metrics as well as a wide range of unfairness tolerances.
    \item We discuss possible countermeasures to mitigate the proposed reconstruction correction method. 
    In particular, we show that even when the fairness information is not disclosed, the adversary 
    can
    estimate it and that the performance of reconstruction correction remains high.
\end{itemize}

The outline of the paper is as follows. 
First, we introduce in Section~\ref{sec:background} the necessary background notions and review the related work on reconstruction attacks.
Afterwards, we describe in Section~\ref{sec:theory} our proposed reconstruction correction strategy before evaluating its empirical effectiveness in Section~\ref{sec:expes}.
Finally, we discuss possible countermeasures in Section~\ref{sec:countermeasures} before concluding.

\section{Background and Related Work}
\label{sec:background}

In this section, we first introduce the considered supervised machine learning setup and the associated notations. Then, we explain how fairness can be quantified in machine learning before reviewing related work on reconstruction attacks.

\subsection{Supervised Machine Learning \& Fairness}


Let $\nattributes$ be the number of \emph{non-sensitive attributes} characterizing an example. For $j\in \{1..\nattributes\}$, $\unsensattrdistribution{j}$ denotes the domain of possible values for attribute $j$, which can be either categorical or numerical, and $\unsensattrdistribution{}~=~\unsensattrdistribution{1} \times \unsensattrdistribution{2} \times \ldots \times \unsensattrdistribution{\nattributes}$. 
Similarly, let $\sensattrdistribution$ (respectively $\groundtruthdistribution$) be the domain of 
a (categorical) \emph{sensitive attribute} (respectively \emph{label}). 
Such sensitive attribute corresponds to personal information such as age, gender or race, which should not be used for a decision-making process due to legal, ethical, social or philosophical reasons~\cite{barocas-hardt-narayanan}.

$\dataset = (\unsensattr{},\sensattr{},\groundtruth{})$ is a dataset drawn from the true (unknown) distribution over $\unsensattrdistribution{} \times \sensattrdistribution \times \groundtruthdistribution$. 
Let $\nexamples{}$ be the number of \emph{examples} (\emph{i.e.}, datapoints) in $\dataset{}$, with $\example_{i \in \{1..\nexamples\}} = (\oneunsensattr{}, \onesensattr{}, \onegroundtruth{}) \in \unsensattrdistribution{} \times \sensattrdistribution \times \groundtruthdistribution$. 
The objective of a supervised machine learning algorithm is to learn a \emph{classifier} $\learningalgo{}(\dataset{}) = \classifier{}$ mapping the attributes space to the label space. 
The explicit use of a sensitive attribute (such as gender, age or race~\cite{DBLP:conf/nips/DingHMS21}) is usually prohibited by law to avoid \emph{disparate treatment}~\cite{10.2307/24758720}. 
Thus, we assume that the sensitive attribute is not used for inference, which means that $\classifier{}: \unsensattrdistribution{} \mapsto \groundtruthdistribution{}$, with $\predictions{} = \classifier{}(\unsensattr{})$ being the predictions of the machine learning model.
In line with the fairness literature, we consider the task of binary classification in this work: $\groundtruthdistribution{} = \{0,1\}$. 
Nonetheless, our framework could easily be extended to non-binary classification provided that fairness constraints are formulated in this more general setting.


\begin{table*}[hbt]
\caption{Summary of the considered statistical fairness metrics}
\label{tab:metrics}
\centering
\begin{tabular}{@{}cccc@{}}
\hline
\textbf{Ref.}                & \textbf{Metric}     & \textbf{Equalized Measure}                 & \textbf{Constraint Expression}                                                                                                                                                   \\ \hline
\cite{dwork2012fairness}    & Statistical Parity (SP)  & Probability of positive prediction         & $\forall \onesensattr,~ \lvert \mathbb{P}(\oneprediction = 1) - \mathbb{P}(\oneprediction = 1 \mid \onesensattr) \rvert \leq \tol$                                               \\
\cite{chouldechova2017fair} & Predictive Equality (PE) & False Positive Rate                        & $\forall \onesensattr,~ \lvert \mathbb{P}(\oneprediction = 1 \mid \onegroundtruth = 0) - \mathbb{P}(\oneprediction = 1 \mid \onesensattr,~\onegroundtruth = 0) \rvert \leq \tol$ \\
\cite{hardt2016equality}    & Equal Opportunity (EO)   & True Positive Rate                         & $\forall \onesensattr,~ \lvert \mathbb{P}(\oneprediction = 1 \mid \onegroundtruth = 1) - \mathbb{P}(\oneprediction = 1 \mid \onesensattr,~\onegroundtruth = 1) \rvert \leq \tol$ \\
\cite{hardt2016equality}    & Equalized Odds (EOdds)      & False Positive Rate and True Positive Rate & Conjunction of Predictive Equality and Equal Opportunity                                                                                                                         \\ \hline
\end{tabular}
\end{table*}

To ensure that machine learning algorithms do not reproduce or create \emph{undesirable biases} (\emph{e.g.}, leading to discrimination), different fairness notions have been proposed in the literature~\cite{DBLP:journals/csur/MehrabiMSLG21}. 
Three main approaches have emerged~\cite{10.1145/3194770.3194776}, namely statistical fairness metrics, individual fairness and causal fairness. 
\emph{Statistical fairness measures}~\cite{dwork2012fairness} aim at equalizing some value (function of the confusion matrix of a classifier, \emph{e.g.}, true positive rate) between several \emph{protected groups} (usually defined by the sensitive attributes). 
\emph{Individual fairness} has as rationale that similar examples should be treated similarly~\cite{dwork2012fairness}. 
Finally, \emph{causal fairness} approaches analyze the causal relationships between the different attributes and the outcome of a classifier, possibly mitigating those deemed as discriminatory.

Many methods have been proposed in recent years~\cite{barocas-hardt-narayanan,caton2020fairness,DBLP:journals/csur/MehrabiMSLG21} to produce \emph{fair} models, which can be divided into three categories depending on which step of the machine learning pipeline they intervene~\cite{DBLP:journals/corr/abs-1810-01943}. \emph{Pre-processing} methods aim at removing undesired correlations from the training data before applying standard learning techniques on the sanitized data~\cite{kamiran2012data} while \emph{in-processing} techniques directly adapt the learning procedure to produce inherently fair models. 
Finally, \emph{post-processing} techniques~\cite{hardt2016equality} modify the outputs of a trained classifier to achieve fairness.

In this paper, we consider the setting in which fairness is expressed using statistical fairness notions. 
Our framework is agnostic to the type of fairness-enhancing technique used. 
This means that the step of the machine learning pipeline in which the fairness intervention occurs does not impact our attack as the latter simply relies on the predictions vector of the model along with the fairness information. 
For our experiments, we consider four metrics widely used in the literature, namely Statistical Parity~\cite{dwork2012fairness}, Predictive Equality~\cite{chouldechova2017fair}, Equal Opportunity~\cite{hardt2016equality} and Equalized Odds~\cite{hardt2016equality}. 
Table~\ref{tab:metrics} provides a summary of these statistical fairness metrics, along with the measure being equalized across the different protected groups and the corresponding mathematical expression.

\subsection{Reconstruction Attacks}
One fundamental objective in privacy protection is to ensure that the output of a computation over a dataset $\dataset{}$ cannot be used to retrieve private information about this dataset~\cite{DBLP:conf/pods/DinurN03}.
Our proposed framework lies in the category of inference attacks~\cite{doi:10.1146/annurev-statistics-060116-054123,DBLP:journals/corr/abs-2007-07646}, which precisely aim at retrieving information regarding the dataset $\dataset{}$ by only observing the outputs of the computation.
In the machine learning field, the computation being performed is usually a learning algorithm whose output is a trained model. 

Different types of inference attacks have been proposed against machine learning models~\cite{DBLP:journals/corr/abs-2005-08679}. 
For instance, membership inference attacks~\cite{DBLP:conf/sp/ShokriSSS17,DBLP:journals/corr/abs-2103-07853} try to infer whether individuals whose profiles are known from the adversary were present in the training set of the model.
Our proposed inference attack is rather a \emph{reconstruction attack}~\cite{doi:10.1146/annurev-statistics-060116-054123,DBLP:journals/corr/abs-2007-07646,DBLP:journals/corr/abs-2005-08679}, sometimes called \emph{model inversion attack}. 
Inference attacks against machine learning often consider two distinct adversarial settings~\cite{DBLP:journals/corr/abs-2005-08679,DBLP:journals/corr/abs-2007-07646}. 
In the \emph{black-box setting}, the adversary does not know the actual trained model's parameters and can only query it through an API. 
In contrast, in the \emph{white-box} setting, the adversary has full knowledge of the model parameters. 
Between these two scenarios, different \emph{gray-box} settings are possible. 
Our attack only requires black-box access to the trained fair model and is agnostic to the actual type of the model, the training algorithm and the fairness mitigation procedure.

Reconstruction attacks have been studied in the context of database access mechanisms since the early 2000s. 
In the considered setup, a database contains records about individuals, which each record being composed of non-private information along with a private bit (one per individual)~\cite{doi:10.1146/annurev-statistics-060116-054123}. 
The adversary performs queries to a database access mechanism, whose outputs are aggregate and noisy statistics about private bits of individuals in the database.
Such reconstruction attacks were introduced and formalized in~\cite{DBLP:conf/pods/DinurN03}, along with some fundamental reconstruction results based on the adversary's capabilities. 
An efficient linear program for reconstructing private bits of a database leveraging counting queries was also proposed.
This linear program was later improved and extended to handle different query types~\cite{10.1145/1250790.1250804}. 
The practical effectiveness of the proposed attacks was demonstrated by a large-scale study carried out by the US Census Bureau in 2018~\cite{10.1145/3291276.3295691} and was part of its motivation to adopt differential privacy for future data releases. 
The linear reconstruction program was also used successfully to break the Diffix commercial database access mechanism~\cite{DBLP:journals/jpc/CohenN20}.
Pursuing the same goal, another attack~\cite{DBLP:conf/uss/GadottiHRLM19} exploited Diffix's data-dependent noise (\emph{i.e.}, sticky noise as well as the addition of static and dynamic noise) to infer private attributes of individuals in a dataset.

One fundamental difference between this line of work and ours lies is the nature of the mechanism accessing the private data. 
In the machine learning (respectively, database access) setup, such mechanism is the learning algorithm (respectively, database access mechanism), and its output is the trained model (respectively, answers to queries). 
Indeed, database access mechanisms use the private information to compute the answer to each query. 
On the contrary, in our setup, the training set sensitive attributes are not accessed anymore at inference time, and all the information regarding them is released at once (with the model itself or its predictions).
However, our objective is similar to these works: we aim at retrieving a \emph{column} of the dataset by leveraging the output of some computation involving this column (query answers in the previously depicted works, trained fair model in ours).

Other previous works have also tackled reconstruction problems in various settings. 
For example in the context of online learning, a reconstruction attack was proposed to infer the \emph{updating set} (newly-collected data used to re-train the deployed model) information using a generative adversarial network leveraging the difference between the model before and after its update~\cite{DBLP:conf/uss/0001B0F020}. 
In collaborative deep learning, it was also shown that an adversarial server can exploit the collected gradient updates to recover parts of the participants' data~\cite{DBLP:conf/atis/PhongA0WM17}.
In the pharmacogenetics field, machine learning models are learnt to propose medical treatments specific to a patient's genotype and background. 
In this sensitive context, a reconstruction attack was proposed, taking advantage of the correlation between the sensitive attributes, the non-sensitive ones, and the output of a trained model. More precisely, the attack takes as input a trained model and some demographic (non-private) information about a patient whose records were used for training and predicts the patient's sensitive attributes~\cite{DBLP:conf/uss/FredriksonLJLPR14}.
Subsequent work proposed model inversion attacks leveraging confidence values output by several ML models to infer private information about training examples given some information about them~\cite{DBLP:conf/ccs/FredriksonJR15}. 
The attack has been shown to be effective against several models and applications, namely decision trees for lifestyle surveys and neural networks for facial recognition.
In the white-box setting, an attack was introduced that exploits the structure of an interpretable machine learning model to reconstruct a probabilistic (uncertain) version of a database~\cite{DBLP:conf/dbsec/GambsGH12}.
While being different both in terms of techniques and objectives, such inference attack still lies in the category of reconstruction attacks.
Finally, other works have studied the intended~\cite{DBLP:conf/ccs/SongRS17} and unintended~\cite{DBLP:conf/uss/Carlini0EKS19} training data memorization of machine learning models, along with different ways to exploit it in a white-box or black-box setting.


\begin{figure*}[htb]
    \centering
    \includegraphics[width=0.87\textwidth]{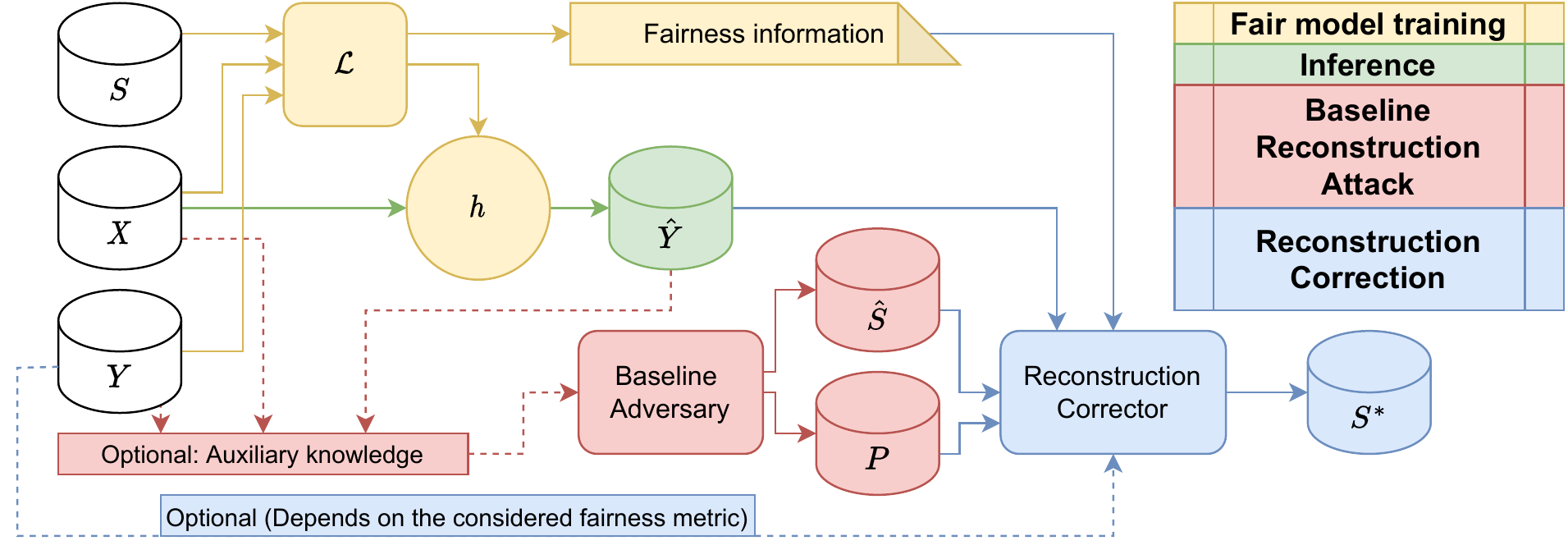} 
    \caption{The proposed attack framework. A model $h$ is learnt by the fair learning procedure $\mathcal{L}$ and used for inference. 
    Then, a \emph{Baseline Adversary} tries to reconstruct the sensitive attributes $\sensattr{}$ of $h$'s training set. 
    Our contribution lies in the \emph{Reconstruction Corrector} component, which takes as input the Baseline Adversary's guess $\hat{\sensattr{}}$ and corrects it to satisfy some fairness property.}
    \label{fig:framework}
\end{figure*}

More closely related to this paper are the works of~\cite{aalmoes2022dikaios} and~\cite{hu2020inference}. 
On the one side, \cite{aalmoes2022dikaios} proposes an attack to infer the sensitive attribute of an example given the model's output for this example. It is the only attack considering the scenario in which the sensitive attribute is not used for inference (what we also assume in this paper). 
In a nutshell, the adversary trains a machine learning model using a separate \emph{attack set} for which the sensitive attributes are known. 
This attack roughly corresponds to the baseline adversaries introduced in section~\ref{subsubsection:attackers}.
On the other side,~\cite{hu2020inference} proposes a mechanism whose principle is related to ours, but considers a very particular setup~\cite{hu2020inference}.
The fair training process is done in a distributed manner, with a learner wanting to build a fair model on some training dataset for which it does not know the sensitive attributes, and a third-party which owns them. 
The learner iteratively sends models parameters to the third-party, which then tells him whether the current model is fair.
The learner then knows, for an entire set of models, whether they satisfy the fairness constraint or not.
Afterwards, he uses Integer Programming techniques to encode this information and perform the reconstruction of the training set sensitive attributes.
While the intuition is similar, our work covers a more general setting (with no assumption on the underlying fairness-enhancing method) in a considerably less favourable attack setup (as the adversary only knows that the final model satisfies the fairness constraint).

\section{Leveraging Fairness to Improve Sensitive Attributes Reconstruction}
\label{sec:theory}

In this section, we first introduce our proposed framework to enhance the reconstruction of sensitive attributes by leveraging 
the 
information about the target model's fairness.
Afterwards, we describe a general model that can be used to correct any adversary's guess about the sensitive attributes vector, given some knowledge expressed as constraints over this vector.
We show how this model can be reformulated to improve scalability in the case of statistical fairness metrics. 
Finally, we discuss how the proposed models can be generalized to handle other metrics and sensitive attribute values. 

\subsection{Attack Pipeline}

Fig.~\ref{fig:framework} illustrates the different components of the considered framework. 
Given a training dataset $\dataset{} = (\unsensattr{}, \sensattr{}, \groundtruth{})$, a model $\classifier{}$ is trained using a fair learning algorithm $\learningalgo{}$, which ensures that $\classifier{}$ is fair on $\dataset{}$ according to some statistical fairness metric with respect to the sensitive attribute $\sensattr{}$.
Note that $\classifier{}$ does not use the sensitive attribute $\sensattr{}$ for inference to prevent disparate treatment~\cite{10.2307/24758720}.
Thus once trained, $\classifier{}$ can be used for inference based only on non-sensitive attributes $\unsensattr{}$.
Our approach does not make any assumption on the underlying fairness-enhancing technique $\learningalgo{}$ used. 
Indeed, the only requirement of our attack is the knowledge of the fairness information. 

The attack itself aims at retrieving the training set sensitive attributes vector $\sensattr{}$. 
In the considered pipeline, $\sensattr{}$ is only used by $\learningalgo{}$ to ensure $\classifier{}$'s fairness (and never used again). 
In the first step of the attack, a \emph{Baseline Adversary} 
makes a \emph{guess} $\hat{\sensattr{}}$ 
on $\sensattr{}$, based on some auxiliary knowledge.
The adversary also outputs a probability vector $\confidencevector$, illustrating his confidence for each component of the guess vector $\hat{\sensattr{}}$. 
Our attack does not assume anything about the form of the auxiliary knowledge. 
If the adversary does not compute confidence scores, the confidence vector can simply be set to the identity vector.

In the second step of the attack, a \emph{Reconstruction Corrector} component takes as input the baseline adversary's guess and confidence vectors ($\hat{\sensattr{}}$ and $\confidencevector$). 
It outputs a new reconstruction guess $\sensattr{}^*$ minimizing the (confidence-weighted) changes to the adversary's guess while satisfying some given properties, such as statistical fairness constraints. 
To ensure the respect of such constraints, the \emph{Reconstruction Corrector} component also needs as input the fairness information, the target model's predictions on the training set $\predictions{}$ as well as (depending on the particular statistical fairness metric at hand, \emph{cf.} Table~\ref{tab:metrics}) the true labels $\groundtruth{}$. Importantly, if the actual fairness information is unknown, it can still be estimated as discussed later in Section~\ref{subsec:hiding_fairness}. 
As stated previously, our attack does not make any assumptions about the target model $\classifier{}$, which can be seen as a black-box as it only requires access to its predictions.

The success of the attack pipeline can be evaluated as the \emph{reconstruction accuracy} of $\sensattr{}^*$ (\emph{i.e.}, proportion of elements of $\sensattr{}$ correctly predicted in $\sensattr{}^*$). 
The core contribution of our 
attack lies in the Reconstruction Corrector component, which, by incorporating solely the fairness information, is able to significantly improve the quality of the reconstruction of the sensitive attribute. 
Such improvement can be quantified by comparing the reconstruction accuracy of the initial adversary's guess $\hat{\sensattr{}}$ and that of the corrected one $\sensattr{}^*$.

\subsection{General Reconstruction Correction Model}


We now introduce $\generalmodel$, a general Integer Programming model implementing the Reconstruction Corrector component of Fig.~\ref{fig:framework}, for the binary sensitive attributes setting. 
Its objective is to modify the adversary's guess for the sensitive attributes of the training examples to satisfy some constraints while minimizing the (confidence-weighted) changes to the adversary's original guess. 
Here, the constraints implement the fairness information.

\paragraph{Inputs} 
\begin{itemize}
    \item $\hat{\onesensattr{}}_i \in \{0, 1\},~i = 1,\ldots,\nexamples$ (adversary's initial guesses)
    \item $\confidence_i \in \{0, 1\},~i = 1,\ldots,\nexamples$ (adversary's confidence for $\hat{\onesensattr{}}_i$)
    \item $\oneprediction_i \in \{0, 1\},~i = 1,\ldots,\nexamples$ (target model $\classifier$'s predictions)
    \item Fairness information: $\classifier$ satisfies fairness constraints for some metric (\emph{e.g.}, SP) and some tolerance $\tol$
\end{itemize}

\paragraph{Decision variables} 
\begin{itemize}
    \item $\variables_i \in \{0, 1\},~i = 1,\ldots,\nexamples$ (corrected guess for the sensitive attributes vector)
\end{itemize}

\paragraph{Model $\generalmodel$}
    \begin{align}
        \min~ &\sum\limits_{i=1}^{\nexamples}{(\confidence_i \cdot (1 - \hat{\onesensattr{}}_i) \cdot \variables_i)} + \sum\limits_{i=1}^{\nexamples}{(\confidence_i \cdot \hat{\onesensattr{}}_i \cdot (1 - \variables_i) )}\label{line1}\\
        s.t.:~&\sum\limits_{i=1}^{\nexamples}{\variables_i} > 0\label{line2}\\
        &\sum\limits_{i=1}^{\nexamples}{(1 - \variables_i)} > 0\label{line3}\\
        &- \tol \leq \frac{\sum_{i=1}^{\nexamples}{\oneprediction{}_i}}{\nexamples} - \frac{\sum_{i=1}^{\nexamples}{\oneprediction{}_i \cdot \variables_i }}{\sum_{i=1}^{\nexamples}{\variables_i}} \leq \tol\label{line4}\\
        &- \tol \leq \frac{\sum_{i=1}^{\nexamples}{\oneprediction{}_i}}{\nexamples} - \frac{\sum_{i=1}^{\nexamples}{\oneprediction{}_i \cdot (1-\variables_i)}}{\sum_{i=1}^{\nexamples}{(1-\variables_i)}} \leq \tol\label{line5}
    \end{align}

The objective~(\ref{line1}) aims at minimizing the confidence-weighted changes to the original adversary's guess $\hat{\sensattr{}}$. 
Each modification of a component $\hat{\onesensattr{}}_i$ of the original adversary's guess is penalized with cost $\confidence_i$ and the model minimizes the total cost.
Constraints~(\ref{line2}) and~(\ref{line3}) simply ensure that the reconstruction contains at least one example from each protected group. 
Finally, constraints~(\ref{line4}) and~(\ref{line5}) encode the fairness constraint for the Statistical Parity metric. 
Here, constraint~(\ref{line4}) (respectively, constraint~(\ref{line5})) ensures that the Positive Prediction Rate (PPR) on group $1$ (respectively, group $0$) is no further than $\tol$ from the PPR on the overall dataset. 

The key idea here is that fairness is ensured by modifying the reconstruction of the sensitive attributes. 
This differs from the typical case of fair model training, in which the sensitive attributes are known and fairness is ensured by modifying the model's predictions $\oneprediction{}_i$ (which, in turn, are fixed here, and exploited to build the sensitive attributes $\variables_i$).

Finally, an optimal solution to our general reconstruction correction model $\generalmodel$ is an assignment of the binary variables $\variables_i$ that minimizes~(\ref{line1}) while satisfying constraints~(\ref{line2}) to~(\ref{line5}). 
This assignment $\sensattr{}^*$ corresponds to the minimum (confidence-weighted) changes to the original adversary guess $\hat{\sensattr{}}$ in order to meet the fairness requirement. 
If the performed changes are correct most of the time (which is to be expected if the adversary provides good confidence scores), then the overall reconstruction accuracy will be improved. 
In any case, the algorithm is guaranteed to find a solution satisfying the fairness constraint - which is not the case of the baseline adversary.
Indeed, as it is able to modify the sensitive attributes guess of all training examples, the model could actually set any fairness value regarding the sensitive attributes corrected reconstruction. 
Thus, the knowledge of the exact training unfairness value (rather than a simple upper bound) could easily be used to reduce the set of acceptable reconstructions and enhance the performance of the reconstruction correction.
Finally, because it explicitly encodes each training example's sensitive attribute, $\generalmodel$ can be used to formulate \emph{any} constraint using such attributes. 


\subsection{Efficient Model for Statistical Fairness}\label{sec:polynomialmodel}

The search space of the reconstruction correction model $\generalmodel$ grows exponentially with the number of training examples $\nexamples$. 
As each element of the sensitive attributes vector $\sensattr{}$ is considered independently from the others (and represented as a binary decision variable), the search space of this model is $O(2^\nexamples{})$, which limits its scalability. 
However, when considering statistical fairness metrics, one does not need such granularity. 
More precisely to satisfy the fairness constraint, the reconstruction corrector may consider exactly four different moves: flipping an element of the reconstructed sensitive attributes $\hat{\onesensattr{}}_i$ from $0$ to $1$ (or the contrary), for an example with prediction $\oneprediction{}=1$ (or $0$). 
Then, for the chosen move, the model will always select the example with the lowest confidence score (and then, eventually, the second lower and so on), which drastically reduces the size of the search space as we explain below.  

Let $n_1^+$ be the number of training examples positively predicted by the target model and assigned to group $1$ by the initial adversary's guess: $n_1^+ = \sum_{i=1}^{\nexamples}{\hat{\onesensattr{}}_i \cdot \oneprediction{}_i}$. 
Similarly, let $n_0^+ =  \sum_{i=1}^{\nexamples}{(1-\hat{\onesensattr{}}_i) \cdot \oneprediction{}_i}$, $n_1^- = \sum_{i=1}^{\nexamples}{\hat{\onesensattr{}}_i \cdot (1-\oneprediction{}_i)}$, and $n_0^- =  \sum_{i=1}^{\nexamples}{(1-\hat{\onesensattr{}}_i) \cdot (1-\oneprediction{}_i)}$. 
The four numbers $n_1^+$, $n_0^+$, $n_1^-$ and $n_0^-$  are the cardinalities of the four groups of examples defining the four possible moves from a fairness perspective. 
For each group, we sort and cumulate the confidence scores associated to its examples and obtain the following arrays: $T_{1^+}$, $T_{0^+}$, $T_{1^-}$ and $T_{0^-}$. 
For instance, $T_{1^+}$ contains the confidence scores associated to the $n_1^+$ training examples positively predicted by the target model and assigned to group $1$ by the initial adversary's guess. 
$T_{1^+}[i]$ is the sum of the $i$ lowest confidence scores among this group. 
Indeed, $T_{1^+}[i]$ is the exact minimal cost of switching the final reconstruction guess from $1$ to $0$ for $i$ examples positively predicted by the target model. 
We use four positive integer decision variables, modeling the number of times each of the four moves is performed to correct the reconstruction.
We now define our efficient model for sensitive attributes reconstruction correction: $\efficientmodel$.

\paragraph{Inputs}
\begin{itemize}
    \item Original guesses
    cardinalities $n_1^+$, $n_0^+$, $n_1^-$ and $n_0^-$.
    \item Arrays of sorted and cumulated adversary's probabilities for each original guess
    : $T_{1^+}$, $T_{0^+}$, $T_{1^-}$ and $T_{0^-}$.
    \item Fairness information: $\classifier$ satisfies fairness constraints for some metric (\emph{e.g.}, SP) and some tolerance $\tol$
     
\end{itemize}

\paragraph{Decision variables} 
\begin{itemize}
    \item $s_{01}^{+} \in [0, n_0^+]$: number of changes of $\hat{\onesensattr{}_i}$ from 0 to 1, for examples such that $\oneprediction{}_i=1$.
    \item $s_{10}^{+} \in [0, n_1^+]$: number of changes of $\hat{\onesensattr{}_i}$ from 1 to 0, for examples such that $\oneprediction{}_i=1$.
    \item $s_{01}^{-} \in [0, n_0^-]$: number of changes of $\hat{\onesensattr{}_i}$ from 0 to 1, for examples such that $\oneprediction{}_i=0$.
    \item $s_{10}^{-} \in [0, n_1^-]$: number of changes of $\hat{\onesensattr{}_i}$ from 1 to 0, for examples such that $\oneprediction{}_i=0$.
\end{itemize}

\paragraph{Model $\efficientmodel$} 
    \begin{align}
        \min~ &T_{0^+}[s_{01}^{+}] + T_{1^+}[s_{10}^{+}] + T_{0^-}[s_{01}^{-}] + T_{1^-}[s_{10}^{-}]\label{line21}\\
        s.t.:~&n_0^+ + n_0^- - s_{01}^{+} - s_{01}^{-} + s_{10}^{+} + s_{10}^{-} > 0\label{line22}\\
        &n_1^+ + n_1^- - s_{10}^{+} - s_{10}^{-} + s_{01}^{+} + s_{01}^{-} > 0\label{line23}\\
        - \tol \leq &\frac{\sum_{i=1}^{\nexamples}{\oneprediction{}_i}}{\nexamples} - \frac{n_1^+ - s_{10}^{+} + s_{01}^{+}}{n_1^+ + n_1^- - s_{10}^{+} - s_{10}^{-} + s_{01}^{+} + s_{01}^{-}}\leq \tol\label{line24}\\
        - \tol \leq &\frac{\sum_{i=1}^{\nexamples}{\oneprediction{}_i}}{\nexamples} - \frac{n_0^+ - s_{01}^{+} + s_{10}^{+}}{n_0^+ + n_0^- - s_{01}^{+} - s_{01}^{-} + s_{10}^{+} + s_{10}^{-}} \leq \tol\label{line25}
    \end{align}
    
Similarly to the general model, the objective~(\ref{line21}) minimizes the confidence-weighted sum of the changes. 
It can be efficiently implemented using \texttt{element} constraints within a Constraint Programming (CP) solver. 
Such constraints are used to access a data array at index given by the value of a variable: $T_{0^+}[s_{01}^{+}] = \texttt{element}(T_{0^+},s_{01}^{+})$. 
Furthermore, when minimizing only the number of changes, one could simply sum the four decision variables. The objective then becomes linear as the whole model which can be solved using off-the-shelf Mixed Integer Linear Programming solvers.

Constraints~(\ref{line22}) and~(\ref{line23}) simply ensure that the reconstruction contains at least one example from each protected group. 
Finally, constraints~(\ref{line24}) and~(\ref{line25}) encode the fairness constraint for the Statistical Parity metric. 
More generally, $\efficientmodel$ could be used to encode any rate constraints on the target model's outputs (using the sensitive attributes), including (but not restricted to) all statistical fairness metrics. 

Once the model is solved, optimal assignments of the four decision variables define the (confidence-weighted) minimal number of moves that must be done to ensure fairness. 
In a post-processing step, the associated moves are performed to the corresponding examples in an increasing order of the confidence scores (so that the overall cost is exactly the objective value~(\ref{line21}) of the solved model). 
This results in the corrected reconstruction vector $\sensattr{}^*$. 
One can notice that $\sensattr{}^*$ is also an optimal solution to the general reconstruction correction model $\generalmodel$. 
Indeed, as stated in Theorem~\ref{th:modeloptsequivalence}, both models share the same set of optimal solutions, even though their encodings of such solutions differ. 
The difference is that some non-optimal solutions to the general model $\generalmodel$ do not correspond to any solution to our efficient model $\efficientmodel$ (\emph{i.e.}, they are simply not part of its search space). 
Such solutions are all the assignments in which the corrector makes one of the four aforementioned moves but does not select the example with the lowest confidence score (which in this context does not make sense).

\newtheorem{theorem}{Theorem}
\begin{theorem}[Equivalence of models]
\label{th:modeloptsequivalence}
In the context of statistical fairness constraints, the general reconstruction correction model $\generalmodel$ and the efficient one 
$\efficientmodel$ share the same set of optimal solutions.
\begin{IEEEproof}
\emph{(a) Any optimal solution to $\generalmodel$ corresponds to a solution to $\efficientmodel$.} 
Let $\sensattr{}^*$ be an optimal solution to $\generalmodel$. 
Then, count the number of performed changes of each type between $\sensattr{}^*$ and $\hat{\sensattr{}}$ (\emph{i.e.}, for an example $i$ with $\oneprediction{}_i=0$ (or $1$), switching $\hat{\onesensattr{}}_i$ from $0$ to $1$ (or the contrary)). 
When performing such changes, the solver must have chosen the examples with the lowest confidence scores, or else another solution also satisfies the fairness constraint and has a better objective function value, which contradicts the optimality hypothesis. 
Afterwards, $\sensattr{}^*$ corresponds to a solution to $\efficientmodel$, represented by the counts for the four moves. 
Indeed, application of the aforementioned post-processing procedure then allows to retrieve $\sensattr{}^*$.

\emph{(b) Any solution to $\efficientmodel$ corresponds to a solution to $\generalmodel$.}
Consider a solution to $\efficientmodel$ and then apply the post-processing step aforementioned. 
The obtained reconstruction vector is a solution to $\generalmodel$.

\emph{(c) The objective function value of any solution of $~\efficientmodel$ is the same in $\efficientmodel$ and $\generalmodel$.}
Consider a solution to $\efficientmodel$ with objective value $o$ and apply the aforementioned post-processing step before plugging the resulting reconstruction vector into $\generalmodel$. 
By construction, the objective value of this solution of $\generalmodel$ will be exactly $o$.

Overall, by~\emph{(a)},~\emph{(b)}, and~\emph{(c)}, each optimal solution to one of the models is also an optimal solution to the other.
\end{IEEEproof}
\end{theorem}

Model $\efficientmodel$ uses four variables whose total sum cannot exceed $\nexamples{}$. 
Its search space is then $O(\nexamples{}^4)$, which is polynomial in the training set cardinality. 
Our resolution method also requires some polynomial $O(\nexamples \cdot log(\nexamples))$ pre-processing and $O(\nexamples)$ post-processing computations, which does not modify the overall solving complexity.
Overall, for statistical fairness constraints, solving our new model is equivalent to solving the general one, but with polynomial search space instead of exponential one. 
In practice, this will lead to running times smaller by several orders of magnitude.


\subsection{Generalizing the Reconstruction Correction}\label{subsec:generalization_metrics}

The proposed models directly encode the Statistical Parity fairness constraints, but can also be used to correct sensitive attributes reconstructions from all the other metrics of Table~\ref{tab:metrics}.
Recall that the Predictive Equality (PE) metric equalizes the False Positive rates (across the protected groups), which is equivalent to satisfying Statistical Parity over the negatively-labelled subset of the training set. 
Then, one can simply use the reconstruction correction model on the negatively-labelled subset of the training set. 
Indeed, PE gives no information on the positively-labelled subset of the training set. 
Similarly, Equal Opportunity equalizes the True Positive rates, and reconstruction can be achieved using the proposed model on the positively-labelled subset of the training set. 
Finally, dealing with the Equalized Odds metric can be done by successively applying Predictive Equality and Equal Opportunity reconstruction corrections. 
Overall, the model proposed for the Statistical Parity metric can actually be used for any of the statistical fairness metrics of Table~\ref{tab:metrics}, by applying the reconstruction correction on the appropriate data slice.

Observe that even though $\generalmodel$ is proposed for the binary sensitive attributes setting, it could easily be generalized by adapting the domains of the $\variables_i$ variables and adding the appropriate cardinalities and fairness constraints for the additional groups.
Extending $\efficientmodel$ 
can also be done by declaring additional variables and constraints. Appendix~\ref{appendix:multi-valued} depicts how both models can be extended to the general case of multi-valued sensitive attributes, along with a discussion regarding the resulting complexity.

\section{Experiments}
\label{sec:expes}

In this section, we present 
our large experimental study regarding the proposed reconstruction framework. 
We consider a wide range of scenarios using two 
fair learning algorithms intervening at different stages of the machine learning pipeline, three datasets of various sizes with diverse sensitive attributes, four fairness metrics and a variety of unfairness tolerances.
First, we describe our baseline adversaries before detailing the experimental setup and the results obtained.

\subsection{Baseline Adversaries Initial Reconstruction}
\label{subsubsection:attackers}
We instantiate the framework described in Fig.~\ref{fig:framework} with two different baseline adversaries, $\attackersix$ and $\attackerseven$. Both have access to an auxiliary \emph{attack set}, $\dataset{}_A = (\unsensattr{}_A, \sensattr{}_A, \groundtruth{}_A)$ which is drawn from the same distribution as the actual training set. 
This attack set models the knowledge of an approximation of the underlying distribution. 
In line with the reconstruction literature~\cite{DBLP:conf/pods/DinurN03,10.1145/1250790.1250804,DBLP:journals/jpc/CohenN20,DBLP:conf/uss/GadottiHRLM19}, we consider that the dataset contains a \emph{``large amount of nonprivate identifying information and a secret bit, one per individual"}~\cite{doi:10.1146/annurev-statistics-060116-054123}. 
Here, the private bit of every individual $i$ is its sensitive attribute $\onesensattr{}_i$. 
Both adversaries hence know the training set non-sensitive attributes vector $\unsensattr{}$ and ground truth labels $\groundtruth{}$ (\emph{i.e.}, all dataset columns except the \emph{secret} one, which is the sensitive attribute in our case).
Adversary $\attackerseven$ also knows the target model's predictions on the training set $\predictions{} = \classifier{}(\unsensattr{})$ and on the attack set $\predictions{}_A = \classifier{}(\unsensattr{}_A)$. 

Adversary $\attackersix$ (respectively $\attackerseven$) relies on the attack set to train an \emph{attack model} to predict $\sensattr{}_A$ from $(\unsensattr{}_A,\groundtruth{}_A)$ (respectively from $(\unsensattr{}_A,\groundtruth{}_A, \predictions{}_A)$). 
Adversary $\attackersix$ (respectively $\attackerseven$) then uses his trained \emph{attack model} to predict $\sensattr{}$ from $(\unsensattr{},\groundtruth{})$ (respectively from $(\unsensattr{},\groundtruth{}, \predictions{})$).
On the one hand, adversary $\attackersix$ can be used to estimate to what extent general knowledge about the distribution can be used to reconstruct the sensitive attributes of the training set. 
Indeed, $\attackersix$ does not have any knowledge about the sensitive attributes singularities of the training set, as such sensitive attributes were not used directly or indirectly for any of its inputs. 
On the other hand, $\attackerseven$ has access to all information that our reconstruction correction will later use, 
which constitutes the strongest baseline possible to compare against our reconstruction correction.
The use of an \emph{attack set} to train an \emph{attack model} is in line with the literature,
and $\attackerseven$ corresponds to the adversary proposed in~\cite{aalmoes2022dikaios}.

The attack model performs binary classification, hence its confidence scores lie between $0.5$ and $1.0$. 
Using these scores directly to weight our reconstruction correction problem would imply that modifying a prediction with confidence $1.0$ (the attacker was certain about it) is better than modifying two predictions with confidence $0.51$ (the attacker was unsure). 
To encourage the reconstruction correction 
to 
target the predictions with the lowest scores, we normalize all confidence scores and exponentiate them in order to enlarge their differences. 
In practice, 
all the normalized scores are set to the power of $k$, in which $k$ is chosen to maximize reconstruction correction accuracy on part of the attacker's data used as a validation set. 
However, other confidence scores processing techniques are possible and may improve the reconstruction correction step.
For instance, an adversary could learn how to best discriminate the confidence scores between correct and incorrect predictions on his attack set.
Overall, each adversary outputs a guess $\hat{\sensattr{}} = \{ \hat{\onesensattr{}}_{i \in \{1\ldots\nexamples\}}\}$ for the sensitive attributes vector, along with a confidence vector $\confidencevector= \{ \confidence_{i \in \{1\ldots\nexamples\}}\}$.

\subsection{Setup}\label{sec:expes_setup}

\subsubsection{Datasets}
\begin{table*}[hbt]
\caption{Summary of the datasets used in our experiments}
\label{tab:datasets}
\centering
\begin{tabular}{@{}cccccc@{}}
\hline
\textbf{Ref.}                    & \textbf{Dataset}                                                                & \textbf{Binary Prediction Task}                                                  & \textbf{\#Datapoints} & \textbf{\#Non-Sensitive Features}                                    & \textbf{Sensitive Feature}                                         \\ \hline
\cite{Dua:2019}                 & UCI Adult Income                                                                & Income above \$50K                                                               & 45,222                & \begin{tabular}[c]{@{}c@{}}7 categorical, 6 numerical\end{tabular}  & Gender (Male/Female)                                               \\
\cite{DBLP:conf/nips/DingHMS21} & \begin{tabular}[c]{@{}c@{}}ACSPublicCoverage\textsuperscript{*}\end{tabular} & \begin{tabular}[c]{@{}c@{}}Coverage from public health insurance\end{tabular} & 98,928                & \begin{tabular}[c]{@{}c@{}}17 categorical, 1 numerical\end{tabular} & Age (First Quartile/Others)                                         \\
\cite{DBLP:conf/nips/DingHMS21} & \begin{tabular}[c]{@{}c@{}}ACSIncome\textsuperscript{*}\end{tabular}        & Income above \$50K                                                               & 135,924               & \begin{tabular}[c]{@{}c@{}}7 categorical, 2 numerical\end{tabular}  & \begin{tabular}[c]{@{}c@{}}Race Code (White/Other)\end{tabular} \\ \hline
\multicolumn{6}{l}{\textsuperscript{*}\footnotesize{(Texas State, 2018)}}
\end{tabular}
\end{table*}
To obtain sufficiently diverse scenarios, we consider three datasets of the fairness literature with different sizes, each with a different binary sensitive attribute. 
The first one is the UCI Adult Income dataset~\cite{Dua:2019}, which gathers records about the 1994 US Census database, with the classification task being to predict whether individuals earn more than \$$50,000$ per year. 
The considered sensitive attribute is gender (female/male). 
We also consider two datasets built from the American Community Survey (ACS) Public Use Microdata Sample (PUMS) of the US Census Bureau. 
More precisely, the datasets are built from data collected in the Texas state in 2018. 
The second dataset, ACSPublicCoverage~\cite{DBLP:conf/nips/DingHMS21}, contains data about individuals under the age of $65$, with an income of less than \$$30,000$, with the classification task being to predict whether they are covered by public health insurance. 
Here, age is used as the sensitive attribute (younger quartile/others). 
The third dataset, ACSIncome~\cite{DBLP:conf/nips/DingHMS21}, gathers records about individuals above the age of $16$, who reported usual working hours of at least 1 hour per week in the past year, and an income of at least \$$100$. 
Similar to the original UCI Adult Income dataset, the classification task is to predict whether individuals earn more than \$$50,000$ per year. 
We rely on the binarized race (white/others) as the sensitive attribute.

Table~\ref{tab:datasets} 
summarizes the
datasets used in our experiments.
For all experiments, each dataset is split between a training set $\left(\frac{1}{3}\right)$, a test set $\left(\frac{1}{3}\right)$ and an attack set $\left(\frac{1}{3}\right)$. 
The test set is only used to ensure that the fair target model is trained appropriately (in particular, to show that it does not overfit). 
The attack set is known by the baseline adversary (see Section~\ref{subsubsection:attackers}).

\subsubsection{Target Fair Models}
To validate our approach, we have tested two off-the-shelf fair learning methods implemented in the \texttt{Fairlearn} library~\cite{bird2020fairlearn}: one in-processing method, ExponentiatedGradient~\cite{DBLP:conf/icml/AgarwalBD0W18}, as well as a post-processing method, ThresholdOptimizer~\cite{hardt2016equality}.
In a nutshell, ExponentiatedGradient~\cite{DBLP:conf/icml/AgarwalBD0W18} formulates the fair classification problem as a sequence of cost-sensitive classification problems. 
Given a cost-sensitive base learner, it follows a two-player game structure in which one player trains the base learner while the other adapts the training examples weights. 
ThresholdOptimizer~\cite{hardt2016equality} takes as input a trained (possibly unfair) classifier and computes group-specific thresholds on the outputs of the classifier to \emph{adjust} its predictions. 
The thresholds are optimized to enforce some fairness constraints while having minimal impact on classification accuracy.
By using two fair learning techniques intervening at different steps of the machine learning pipeline, we want to emphasize that our method is completely agnostic to the type of fairness intervention. 
Indeed, the only information used by our reconstruction correction strategy is the final fairness information, along with the predictions of the model.
For both methods, we use \texttt{scikit-learn}~\cite{scikit-learn} Decision Tree classifiers as base learners with the maximum depth being set to 8 and all other parameters left to their default values.

\subsubsection{Fairness Metrics}
We run experiments for the four fairness metrics presented in Table~\ref{tab:metrics}.
Experiments using the ExponentiatedGradient method use 49 different values of the unfairness tolerance $\tol$, ranging non-linearly from $0.0$ (exact fairness) to $0.20$ (loose constraint).
The ThresholdOptimizer method modifies the initial model's predictions to approximate 0.0 unfairness, so we cannot vary the unfairness tolerance here.

\subsubsection{Attack Models}

The attack models used by our baseline adversaries are \texttt{scikit-learn}~\cite{scikit-learn} Random Forest classifiers, which are known to be resistant to overfitting and generalize well in many situations.
This hypothesis class was chosen based on thorough preliminary experiments. 
To handle sensitive attributes imbalance~\cite{aalmoes2022dikaios}, we use a class-balanced loss. 
The Random Forest hyperparameters are optimized using the \texttt{HyperOpt-Sklearn} framework~\cite{komer2014hyperopt}, with a maximum of 100 evaluations for its Tree of Parzen Estimators search algorithm. 
This setup ensures that the baseline adversary 
implements a strong baseline and is in line with the literature. 

\subsubsection{Reconstruction Correction}
Our efficient reconstruction correction model $\efficientmodel$ (depicted in Section~\ref{sec:polynomialmodel}) is implemented and solved using the \texttt{IBM ILOG CP Optimizer Version 12.10}\footnote{{https://www.ibm.com/analytics/cplex-cp-optimizer}} via the \texttt{DOcplex}\footnote{{http://ibmdecisionoptimization.github.io/docplex-doc/}} Python Modeling API (version 2.21.207) and its default configuration. 
The number of threads used in CP Optimizer is set to 1 and the optimality tolerance (absolute and relative) is set to 0.0. 
Indeed, due to the probabilities exponentiation process presented in Section~\ref{subsubsection:attackers}, some values can be very small and would lie below the solver's default optimality tolerance. 
Our reconstruction correction method is implemented as a Python module and will be 
released publicly upon acceptance.

\def\figwidth{0.42}
 \begin{figure*}[h!]
    \begin{center}
    \includegraphics[width=\figwidth\textwidth]{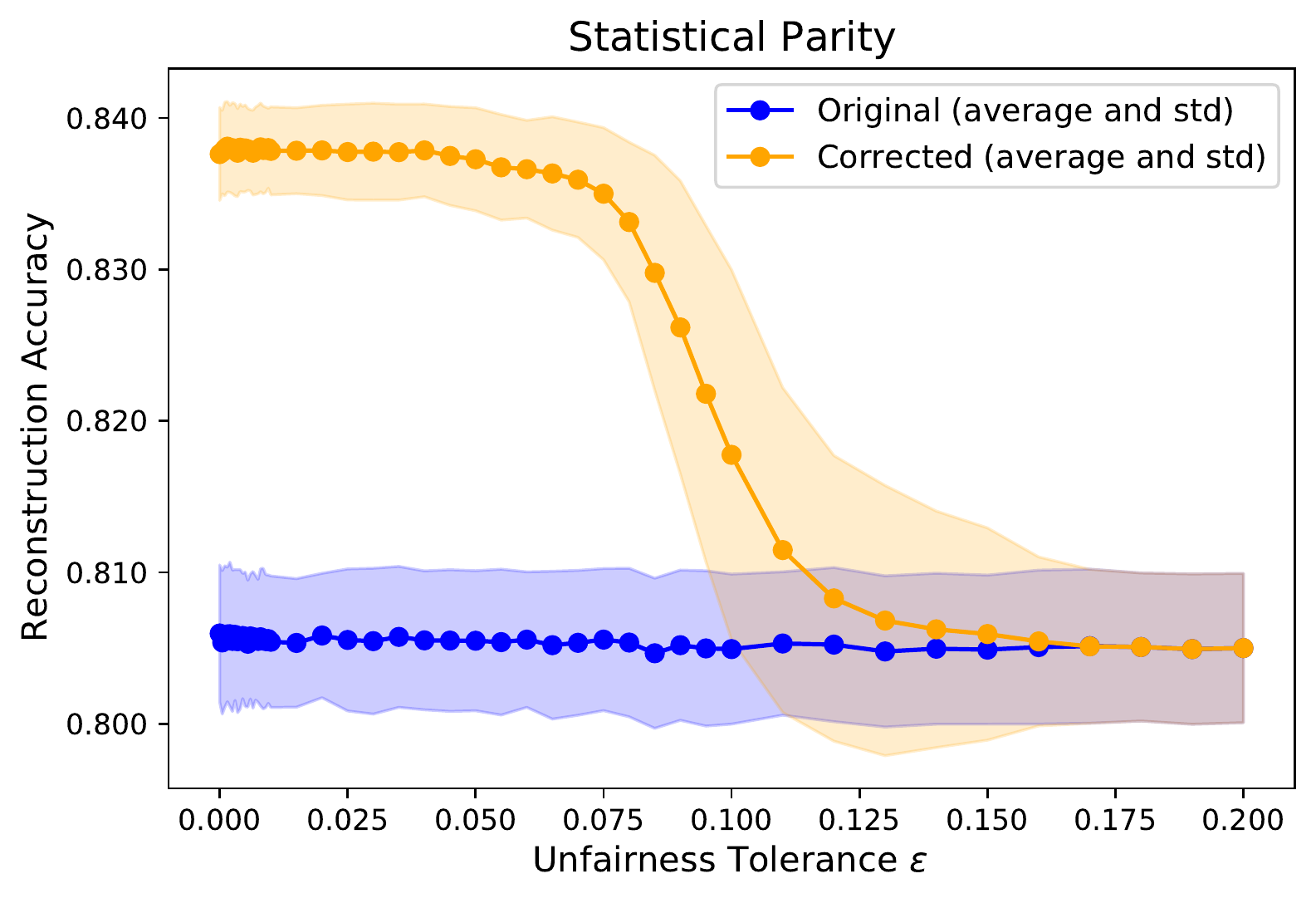} 
   \includegraphics[width=\figwidth\textwidth]{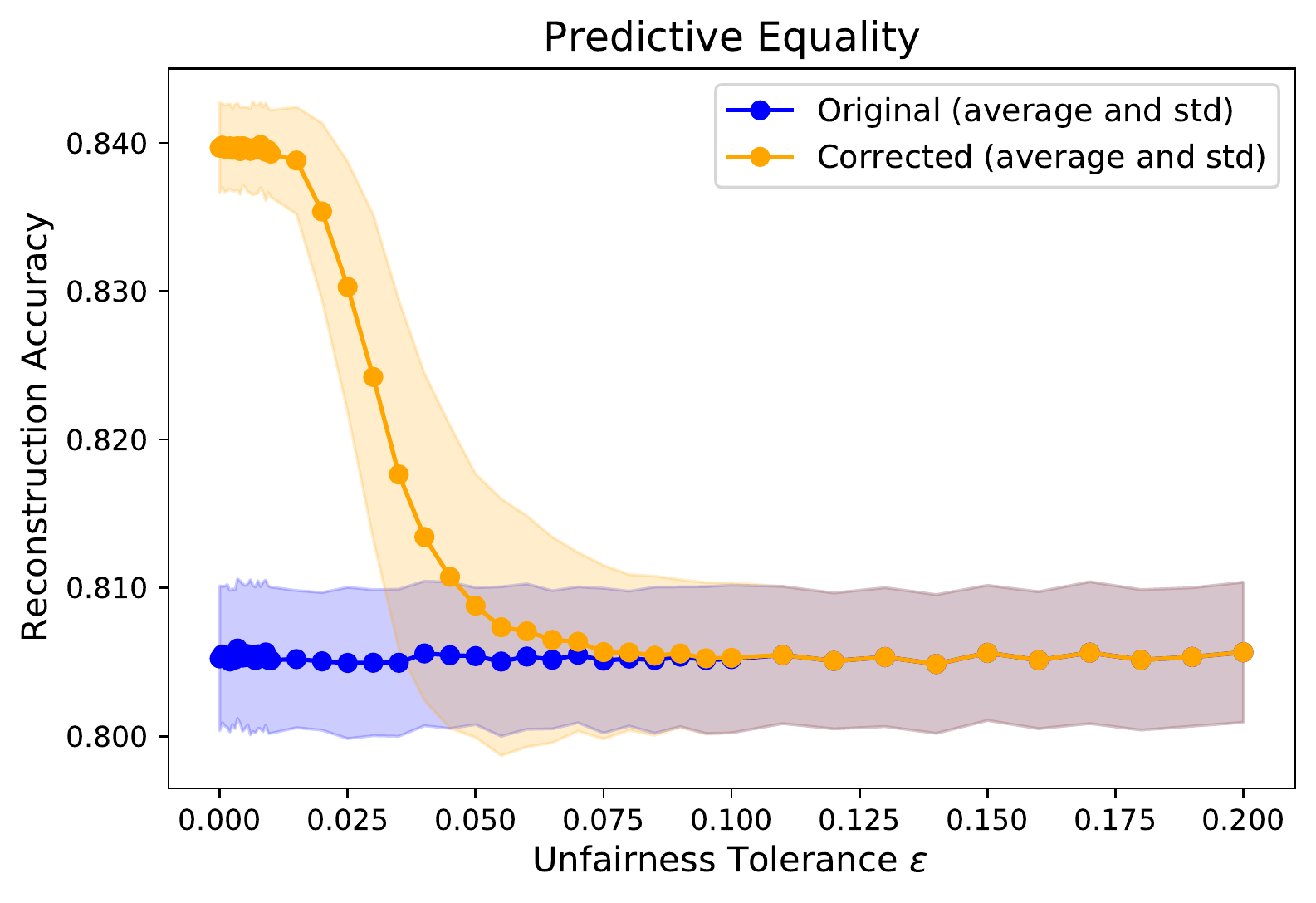}
    \includegraphics[width=\figwidth\textwidth]{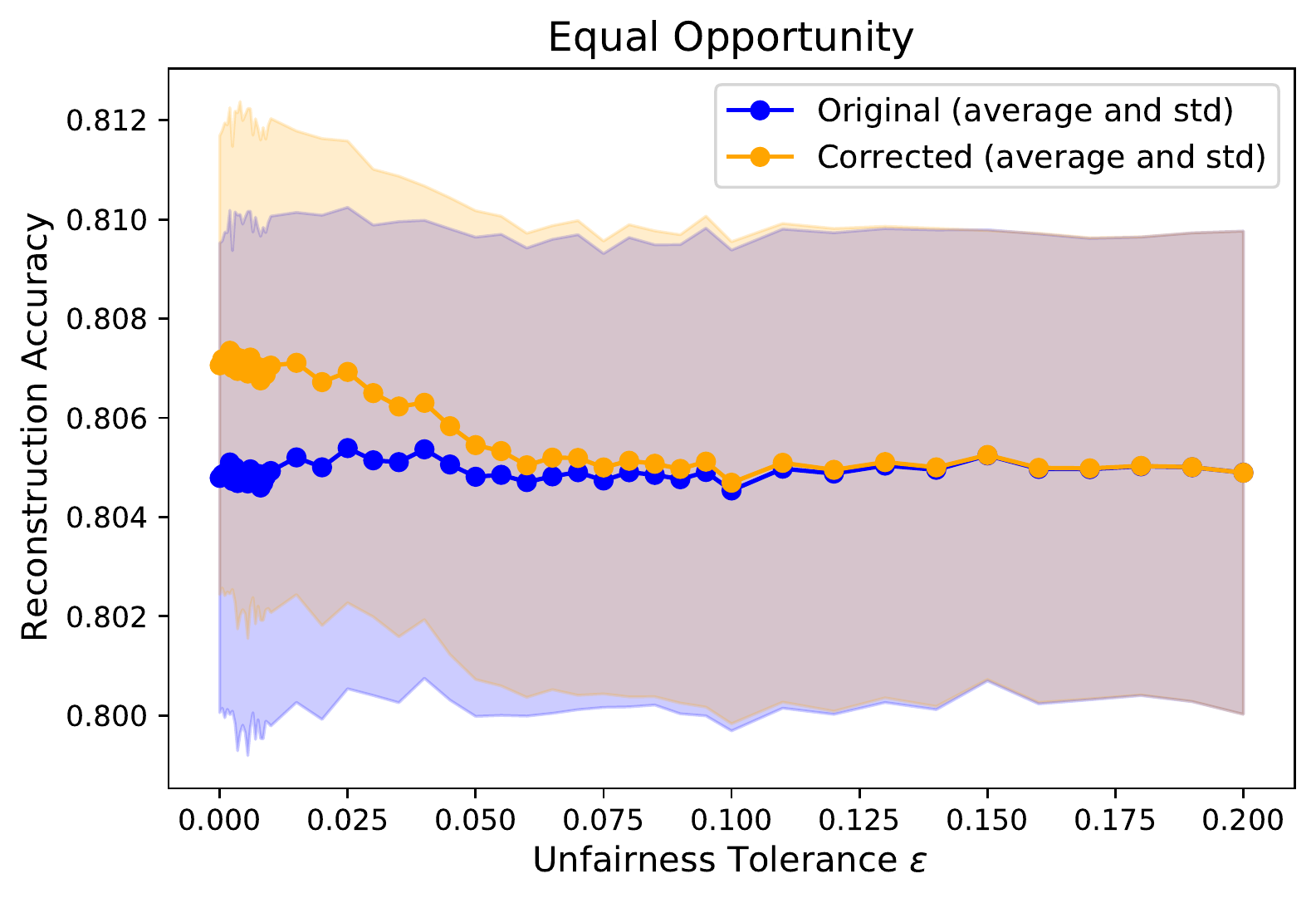}
    \includegraphics[width=\figwidth\textwidth]{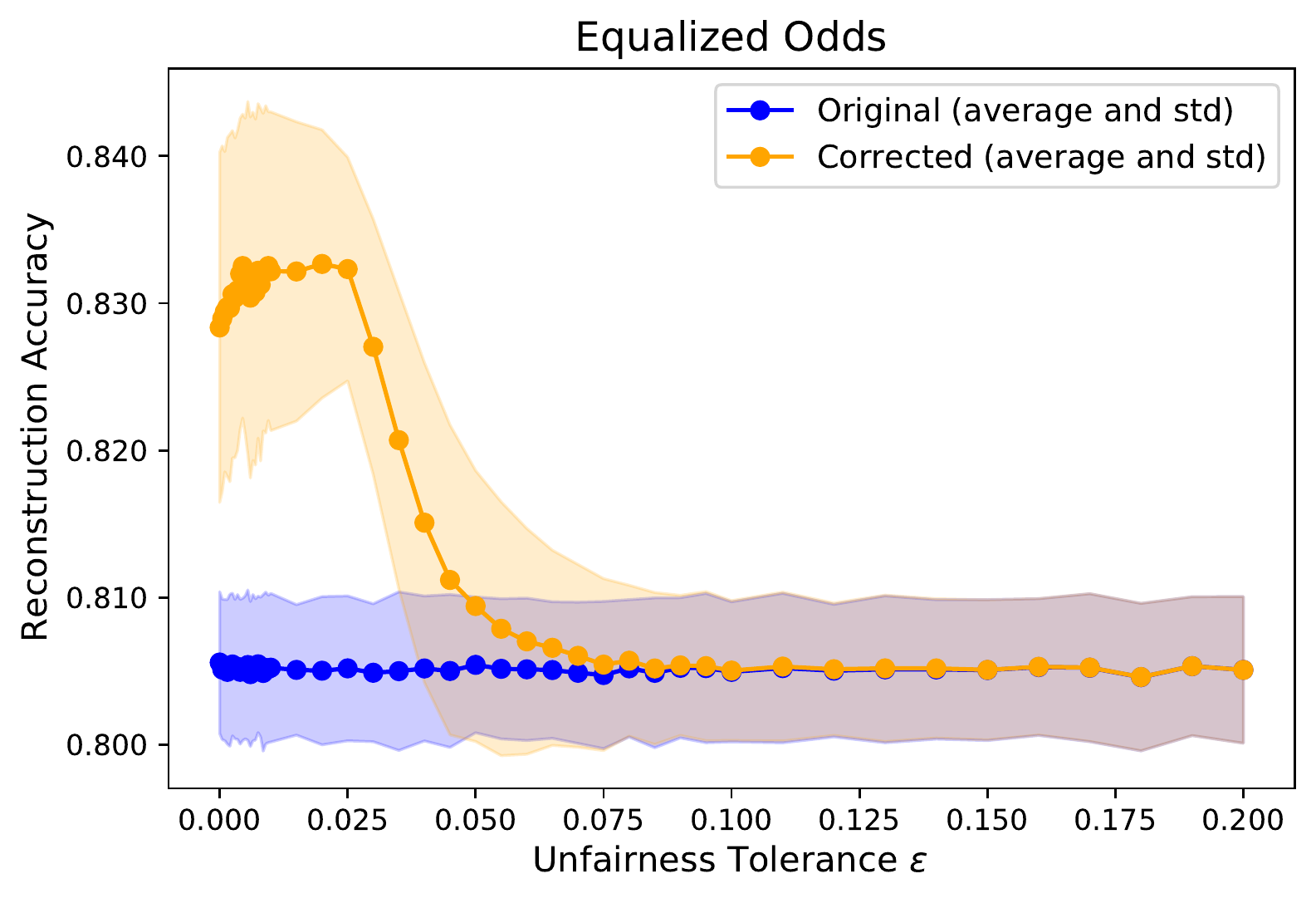}
    \end{center}
    \caption{Corrected and original (adversary $\attackerseven$) reconstruction quality, for our experiments using the UCI Adult Income dataset.}
\label{fig:adult_results_inproc}
\end{figure*}

 \begin{figure*}[h!]
    \begin{center}
    \includegraphics[width=\figwidth\textwidth]{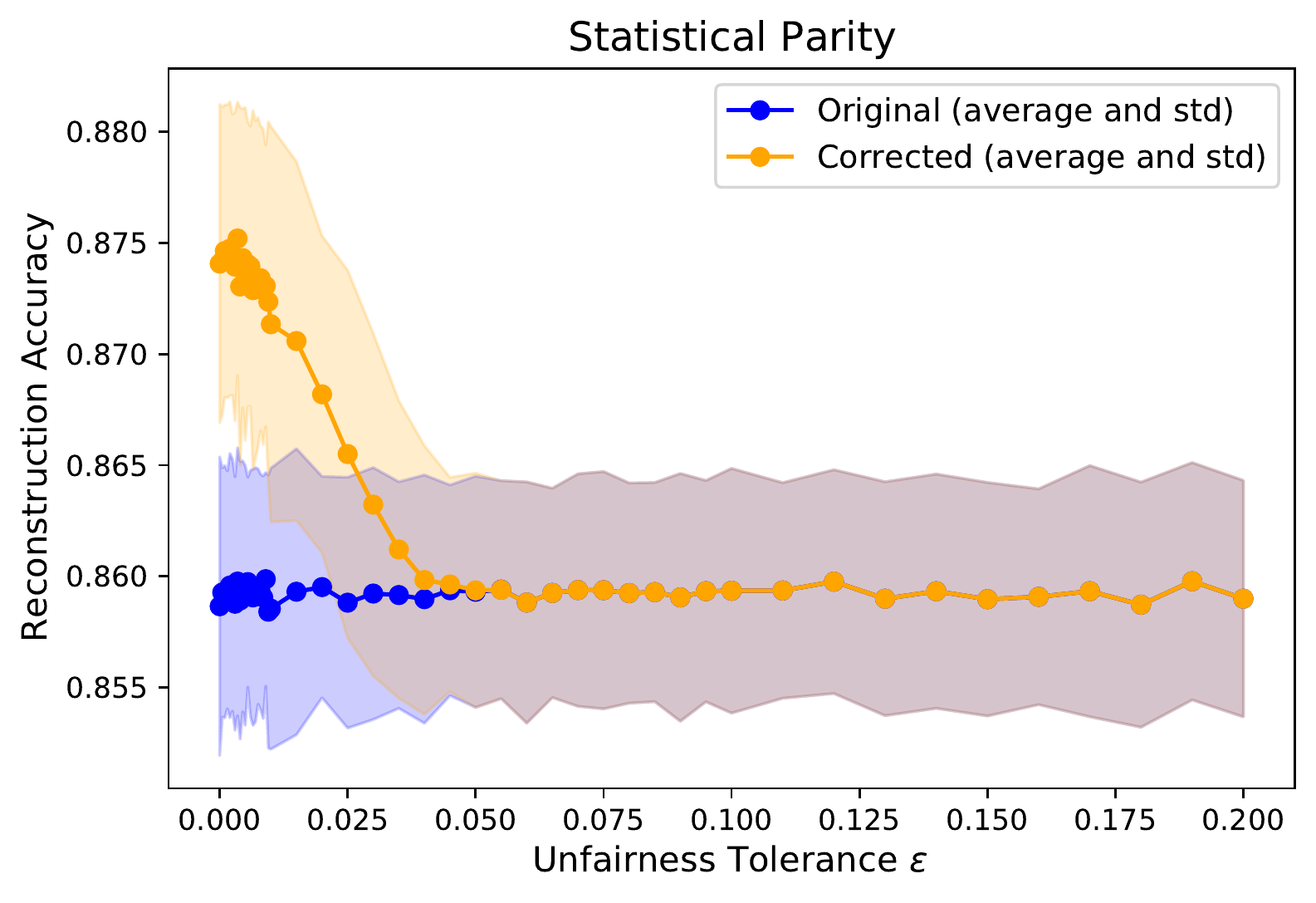} 
   \includegraphics[width=\figwidth\textwidth]{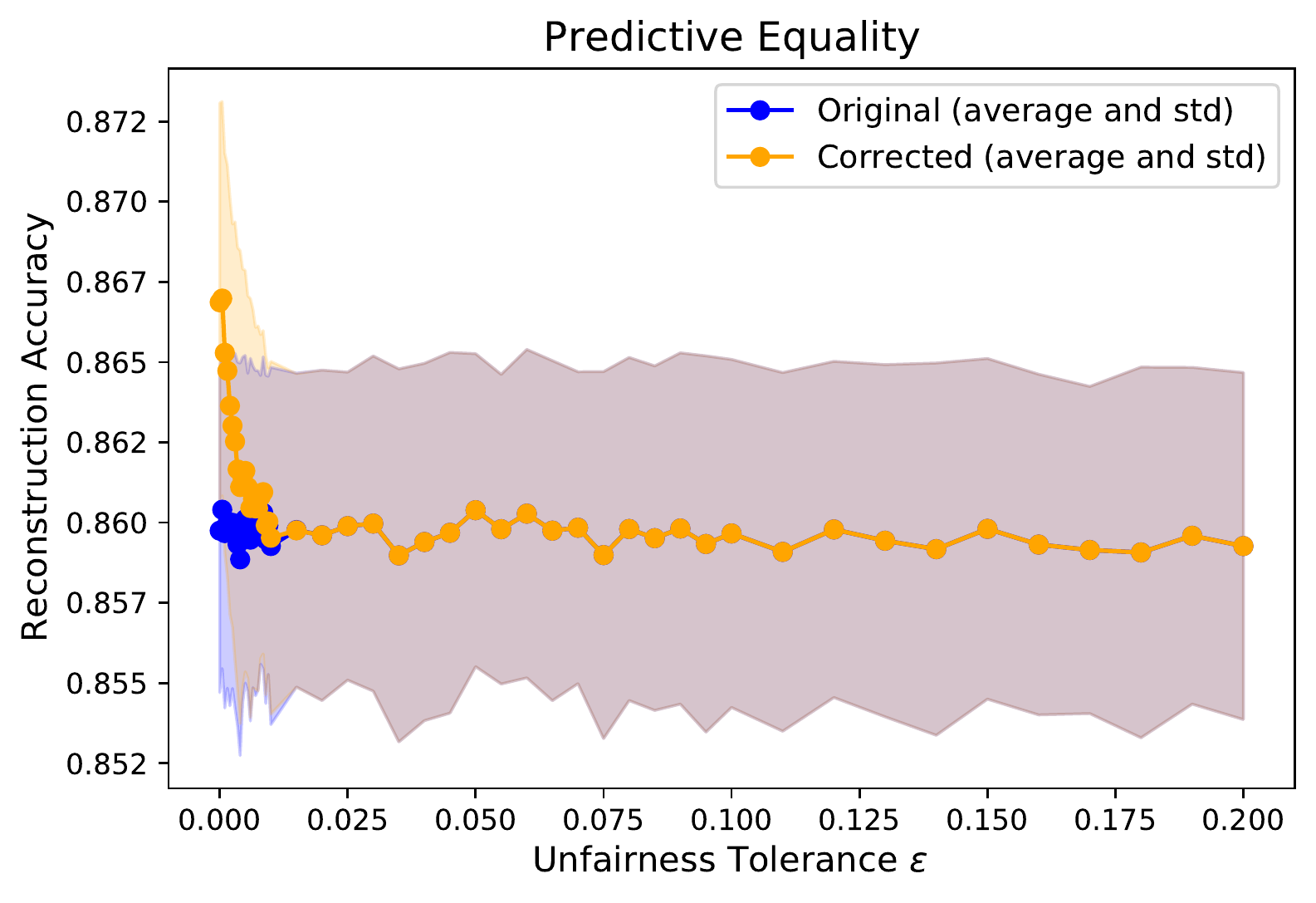}
    \includegraphics[width=\figwidth\textwidth]{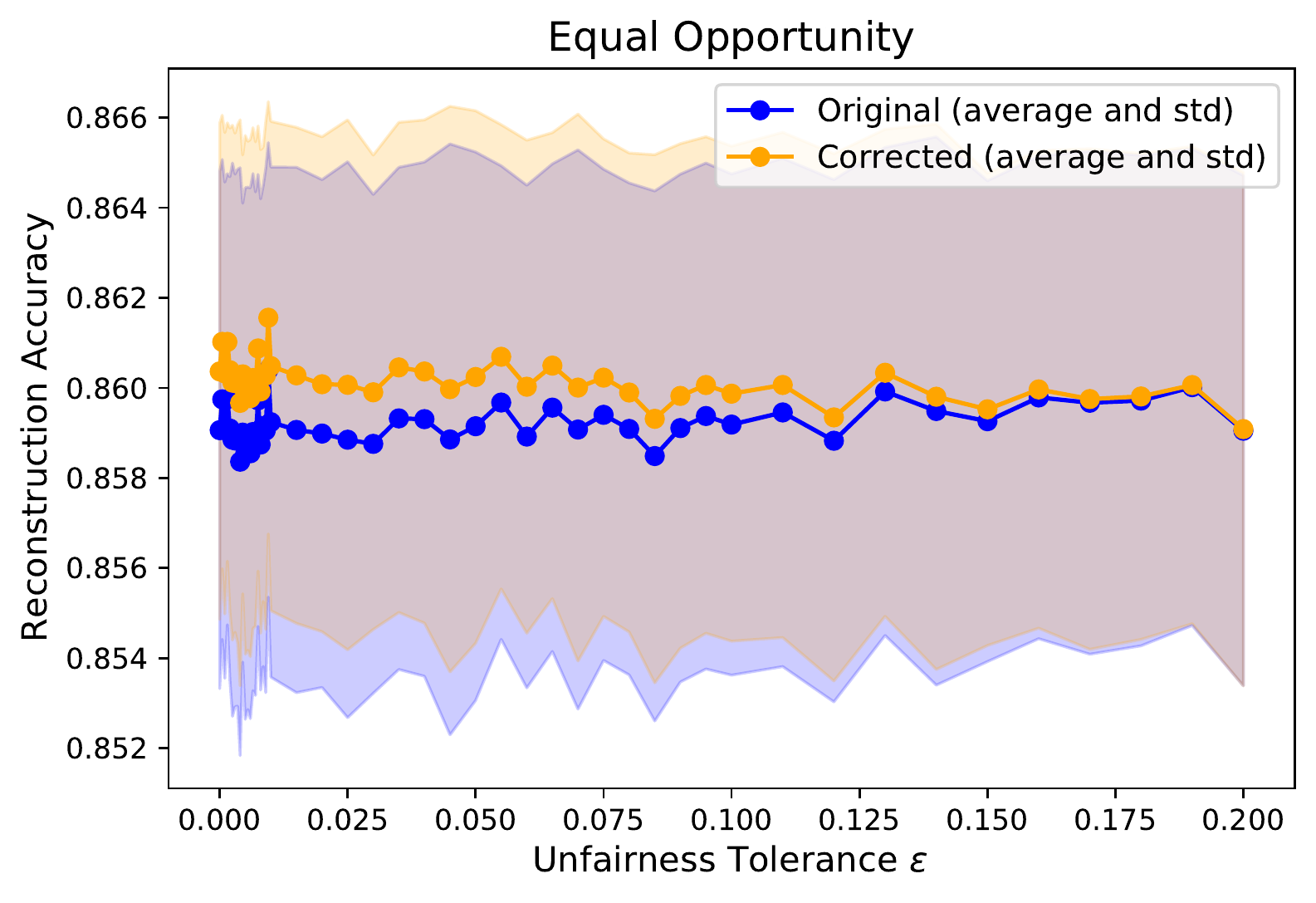}
    \includegraphics[width=\figwidth\textwidth]{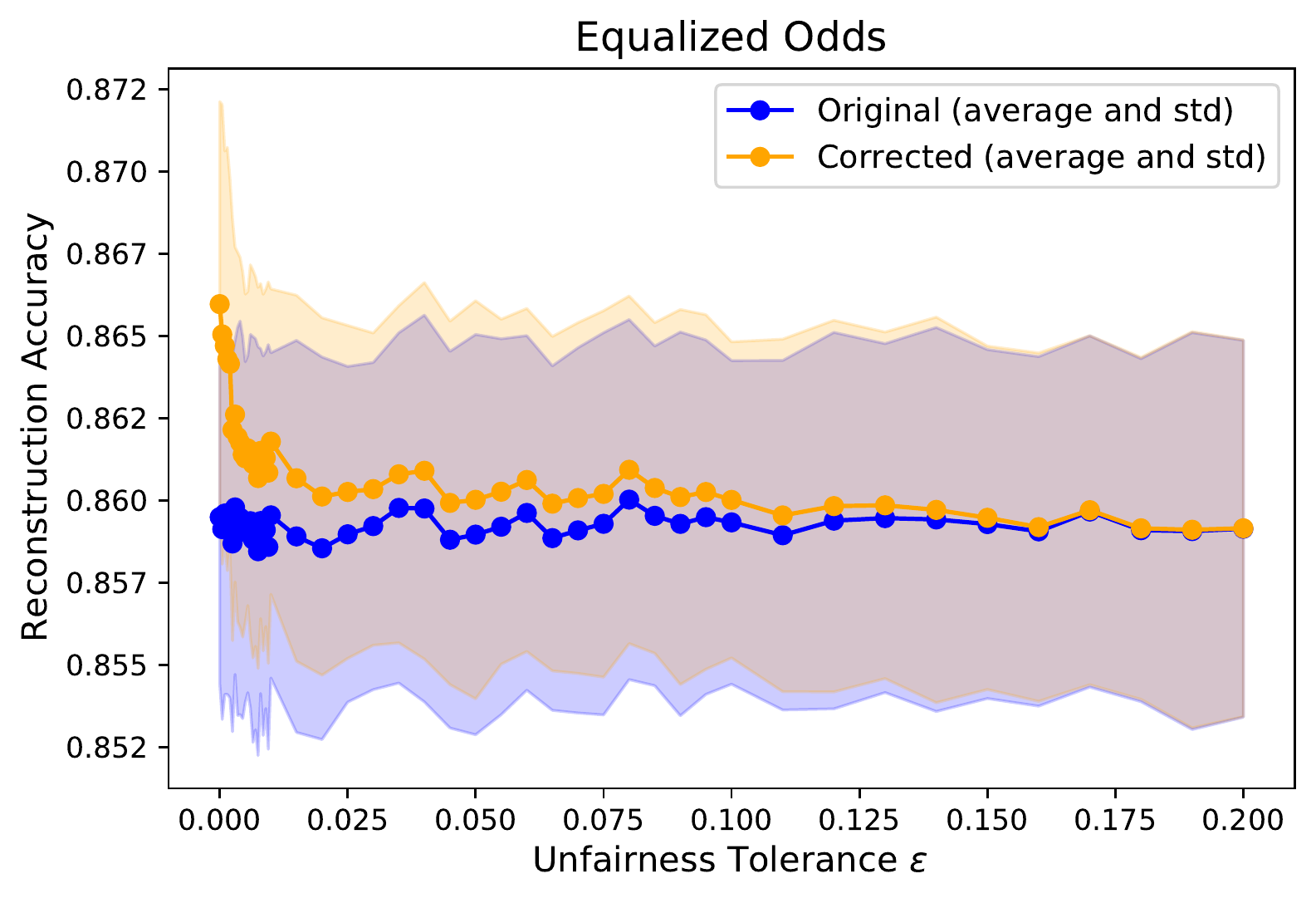}
    \end{center}
    \caption{Corrected and original (adversary $\attackerseven$) reconstruction quality, for our experiments using the ACSPublicCoverage dataset.}
\label{fig:acspubliccoverage_results_inproc}
\end{figure*}


 \begin{figure*}[h!]
    \begin{center}
    \includegraphics[width=\figwidth\textwidth]{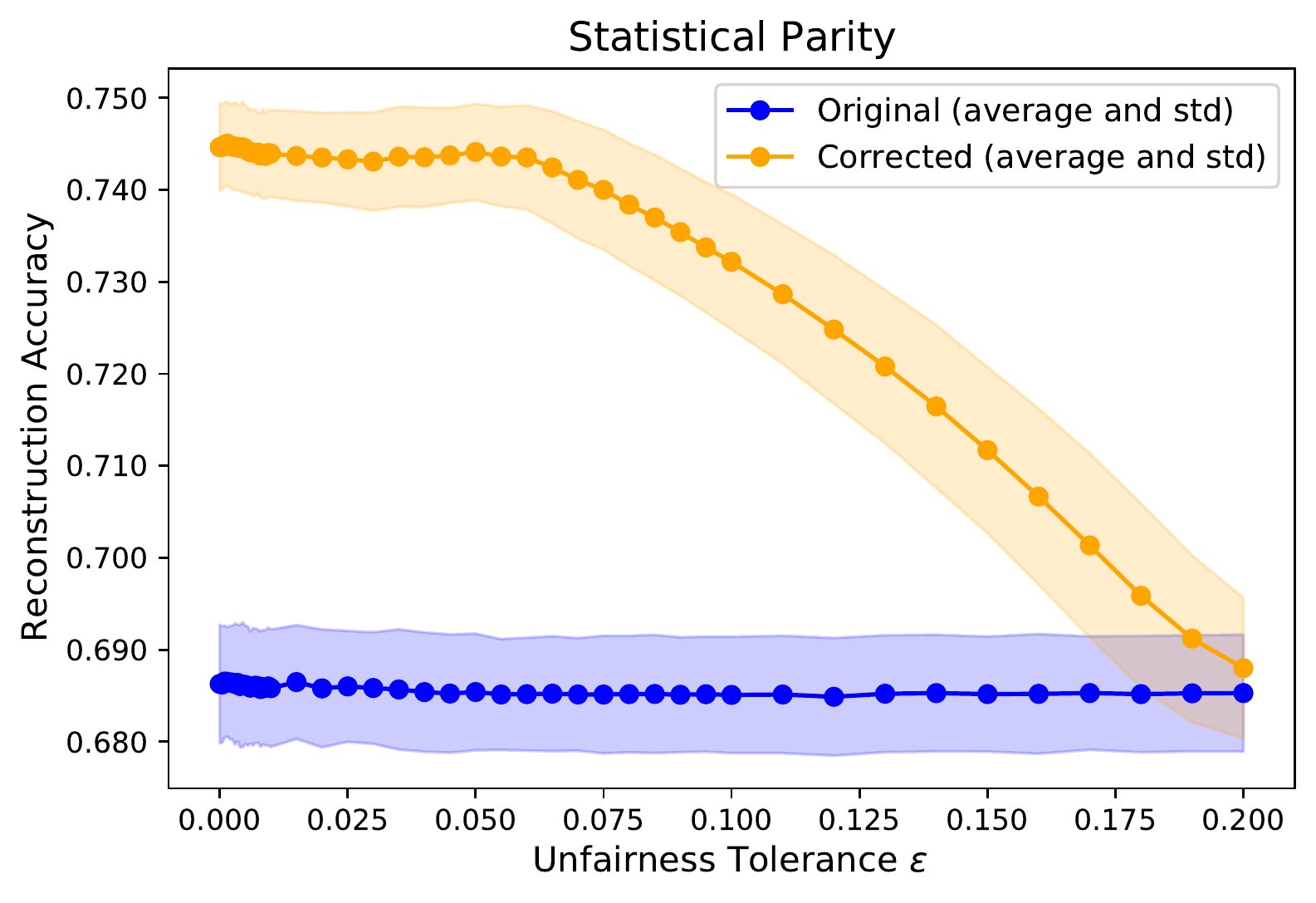} 
   \includegraphics[width=\figwidth\textwidth]{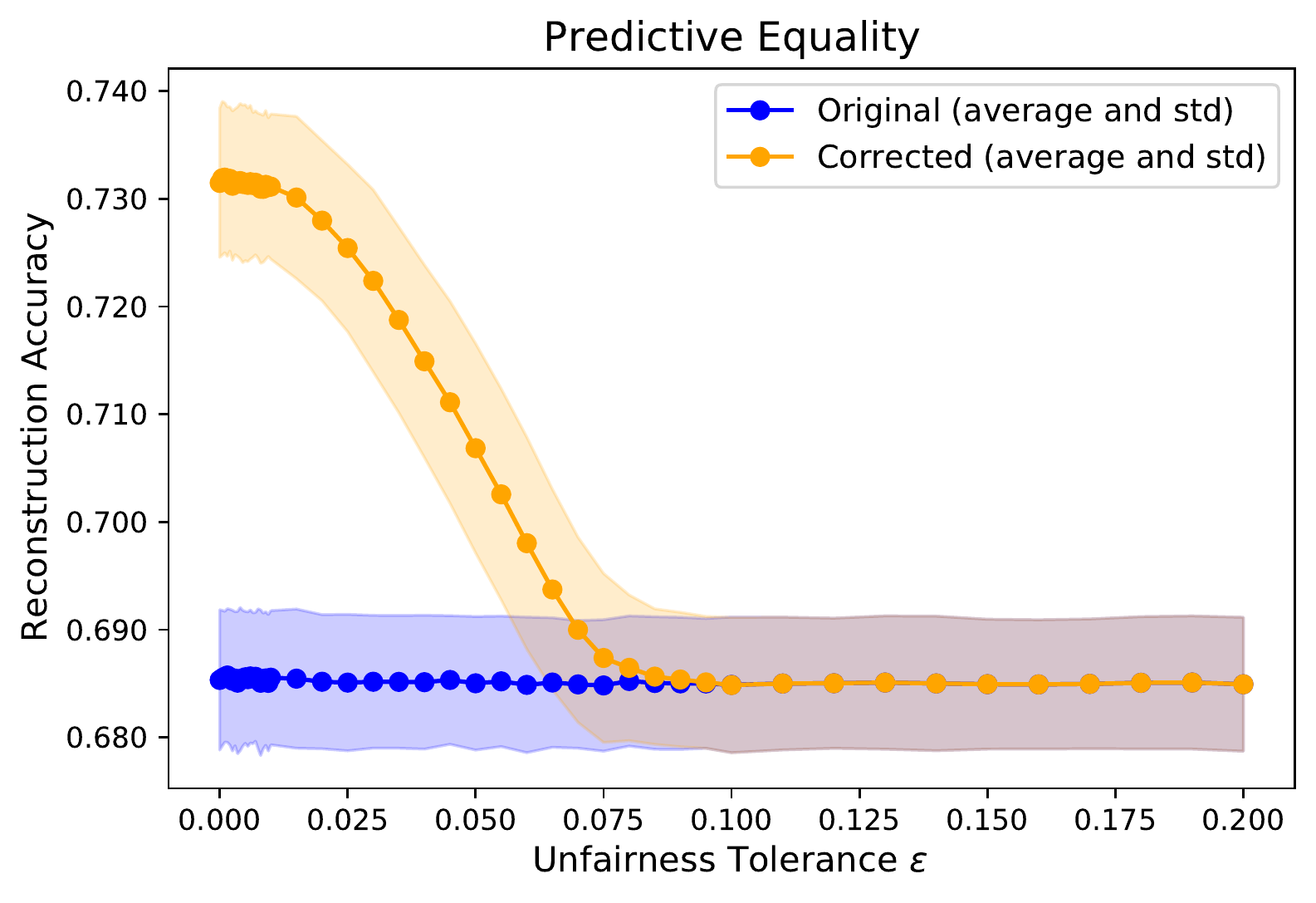}
    \includegraphics[width=\figwidth\textwidth]{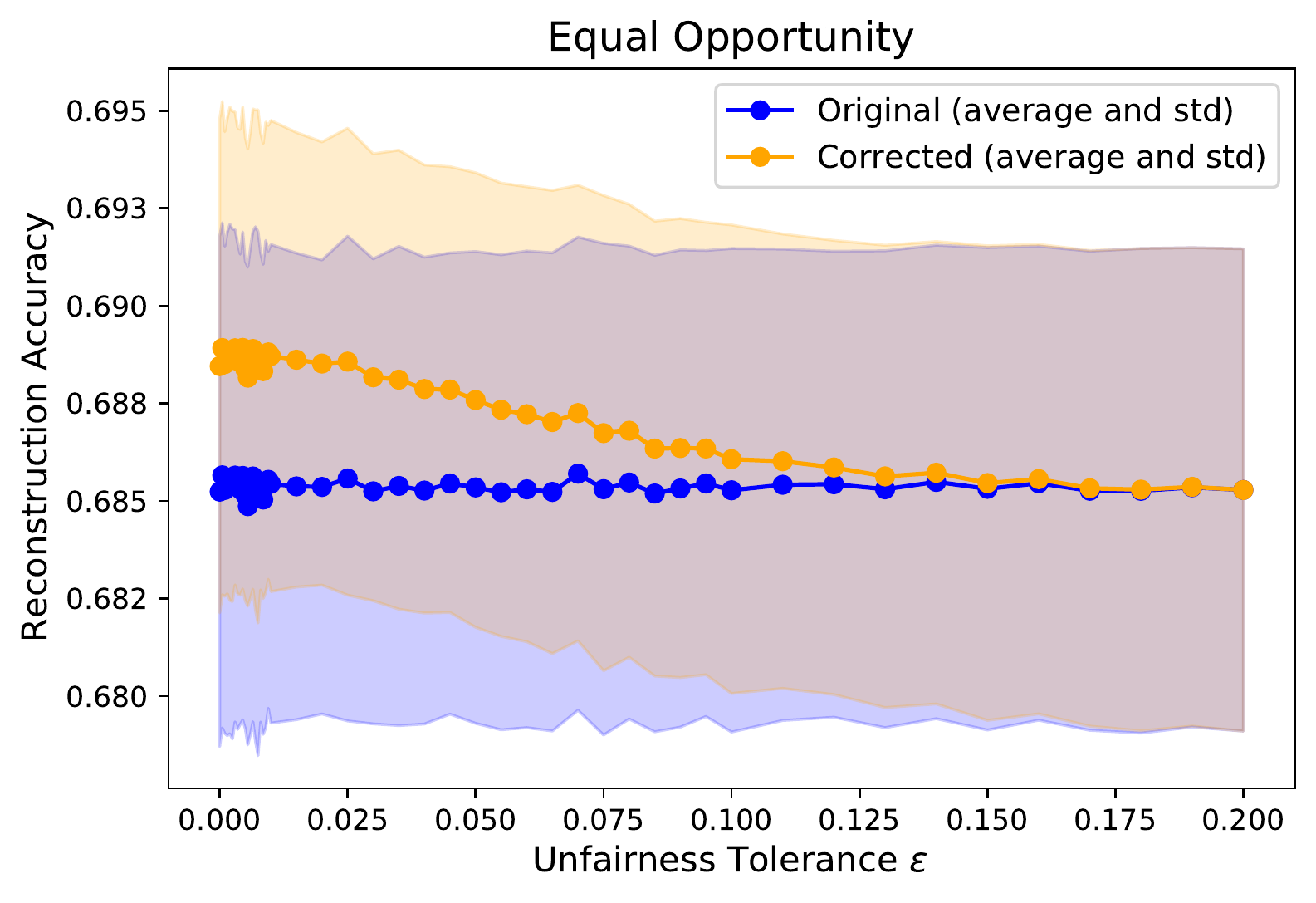}
    \includegraphics[width=\figwidth\textwidth]{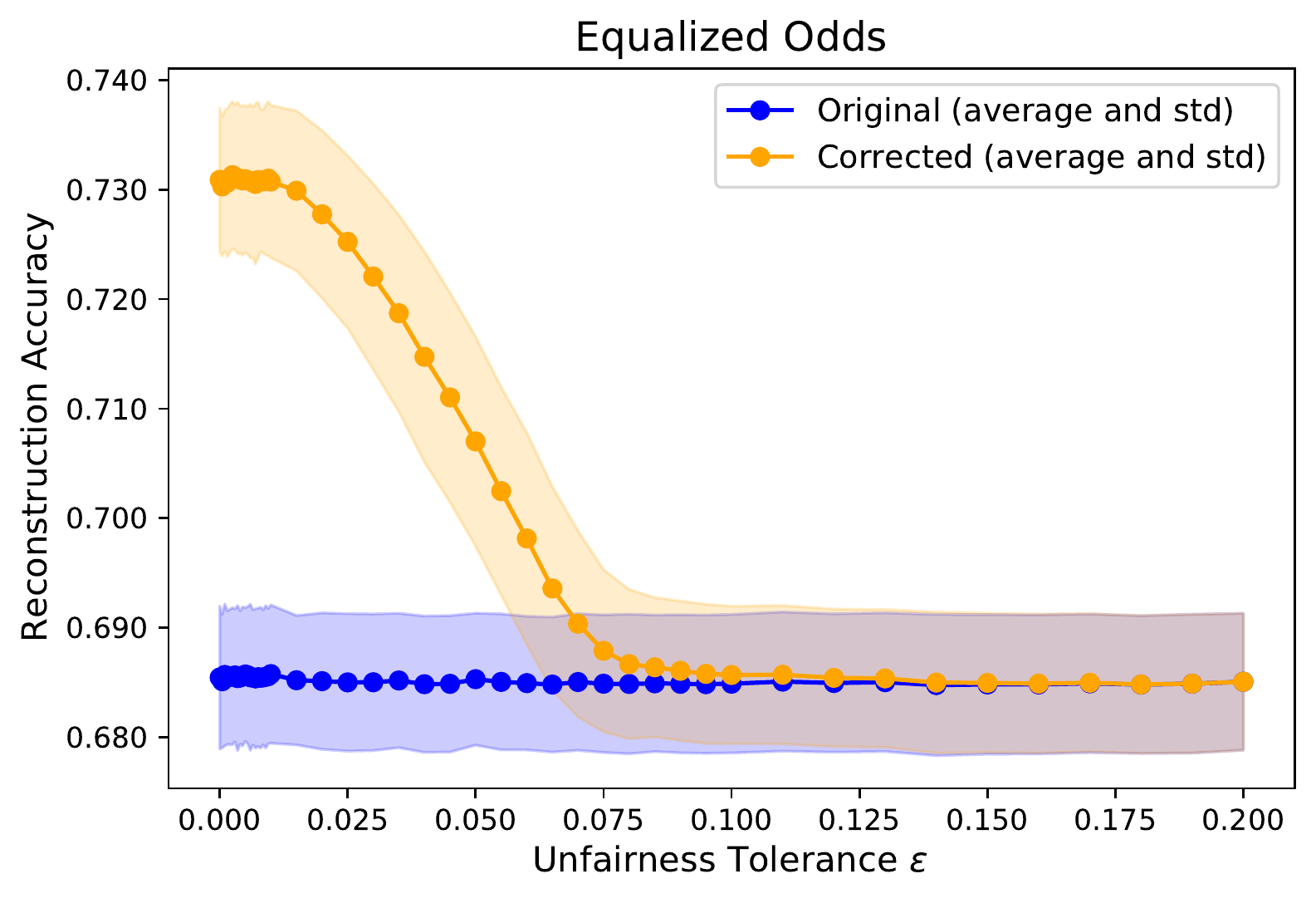}
    \end{center}
    \caption{Corrected and original (adversary $\attackerseven$) reconstruction quality, for our experiments using the ACSIncome dataset}
\label{fig:acsincome_results_inproc}
\end{figure*}

\subsubsection{Experimental Parameters}
We set a one minute timeout for the reconstruction correction step (model creation and solving).
It was never reached in practice, and all models were solved to optimality in less than a few seconds (less than one second in average).
Each experiment is repeated 100 times, with different seeds for the data split process and the random state of the algorithms. 
The results are averaged over the 100 runs and the standard deviation is reported. 
All experiments are run on a computing cluster over a set of homogeneous nodes using Intel Xeon E5-2683 v4 Broadwell @ 2.1GHz CPU.

\subsection{Results}

\subsubsection{Experiments using the ExponentiatedGradient technique}\label{expes:in_proc}

Results of our experiments using the ExponentiatedGradient~\cite{DBLP:conf/icml/AgarwalBD0W18} method are displayed for the different datasets in Fig.~\ref{fig:adult_results_inproc}, \ref{fig:acspubliccoverage_results_inproc} and \ref{fig:acsincome_results_inproc}.
The training and test performances of the 
target fair models are shown in Appendix~\ref{appendix:target_models_perfs}, and show that they do not overfit.
As expected, training accuracy and unfairness both increase when the fairness constraint is relaxed (\emph{i.e.}, $\tol$ increases). 
Due to the models' 
good generalization, test accuracy and unfairness follow the same trends.

The reconstruction accuracy results 
displayed in Fig.~\ref{fig:adult_results_inproc}, \ref{fig:acspubliccoverage_results_inproc} and \ref{fig:acsincome_results_inproc} for the three considered datasets and the four fairness metrics demonstrate the effectiveness of the proposed approach. 
In this section, we report the results for adversary $\attackerseven$. 
Results for adversary $\attackersix$, which are provided in Appendix~\ref{appendix:expes_attacker_6}, are almost perfectly identical and follow the same trends. 
As the adversary $\attackerseven$ exploits all the information that our reconstruction correction uses, any further improvement in the reconstruction accuracy can only be explained by the semantics of the fairness constraint integrated in our Reconstruction Corrector model. 
Recall that the reconstruction accuracy is the proportion of training examples $\example_i$ for which the sensitive attribute $\onesensattr{}_i \in \sensattr{}$ was correctly reconstructed
(in the baseline attacker original guess $\hat{\onesensattr{}}_i \in \hat{\sensattr{}}$ or in the corrected one $\onesensattr{}_i^* \in \sensattr{}^*$).

One can observe that the corrected reconstruction is always more accurate than the adversary's original guess, which means that the changes made by the reconstruction correction model 
are correct most of the time.
Furthermore, the corrected reconstruction accuracy gets better as the fairness constraint becomes tighter (\emph{i.e.}, lower values of the unfairness tolerance $\tol$). 
Indeed, the 
reconstruction accuracy improvement is related to the amount of bias mitigated by the fair learning technique, which in turn depends on the considered fairness metric, the unfairness tolerance and the original data bias. 
For tight fairness constraints, 
we observe reconstruction accuracy absolute improvements 
up to $0.06$, as in the experiments using the Statistical Parity metric on the ACSIncome dataset (Fig.~\ref{fig:acsincome_results_inproc}, top left).
Such improvements are due to the fairness information, which is the only constraint of our correction models.

Recall that the Predictive Equality (respectively Equal Opportunity) metric only applies to the negatively-labelled (respectively positively-labelled) training examples. 
This means that such metrics can only help in partially correcting the adversary's guess (as described in Section~\ref{subsec:generalization_metrics}). 
Because the datasets used are imbalanced, with the majority of training examples belonging to the negative class, the Equal Opportunity metric relates only to a minority of training examples.
As a result, the reconstruction accuracy improvement is more modest than for the remaining metrics. 
Indeed, even with a close rate of correct modifications, the number of corrections applied (and thus the overall improvement) is smaller.

\begin{table*}[htb!]
\caption{Summary of the results of our experiments using a post-processing method for fairness}
\label{tab:postproc_results}
\centering 
\begin{tabular}{ccccccccc}
\hline
\multicolumn{1}{c|}{\multirow{2}{*}{\textbf{Metric}}} & \multicolumn{4}{c|}{\textbf{Target model (under attack)}}                                                                                                                                                   & \multicolumn{2}{c|}{\textbf{\begin{tabular}[c]{@{}c@{}}Baseline Reconstructions\end{tabular}}}      & \multicolumn{2}{c}{\textbf{\begin{tabular}[c]{@{}c@{}}Corrected Reconstructions\end{tabular}}}                                                                                             \\ \cline{2-9} 
\multicolumn{1}{c|}{}                                 & \multicolumn{1}{c|}{\textit{\textbf{Train Acc.}}} & \multicolumn{1}{c|}{\textit{\textbf{Test Acc.}}} & \multicolumn{1}{c|}{\textit{\textbf{Train Unf.}}} & \multicolumn{1}{c|}{\textit{\textbf{Test Unf.}}} & \multicolumn{1}{c|}{\textit{\textbf{$\attackersix$}}} & \multicolumn{1}{c|}{\textit{\textbf{$\attackerseven$}}} & \multicolumn{1}{c|}{\textit{\textbf{\begin{tabular}[c]{@{}c@{}}$\attackersix$\end{tabular}}}} & \textit{\textbf{\begin{tabular}[c]{@{}c@{}}$\attackerseven$\end{tabular}}} \\ \hline
\multicolumn{9}{c}{\textbf{UCI Adult Income dataset}}                                                                                                                                                                                                                                                                                                                                                                                                                                                                                                                      \\ \hline
SP                                                    & $0.820 \pm 0.008$                                 & $0.808 \pm 0.009$                                & $0.003 \pm 0.002$                                 & $0.005 \pm 0.003$                                & $0.808 \pm 0.005$                                 & $0.814 \pm 0.006$                                 & $0.851 \pm 0.003$                                                                                       & $\mathbf{0.858 \pm 0.005}$                                                         \\
PE                                                    & $0.849 \pm 0.005$                                 & $0.836 \pm 0.006$                                & $0.002 \pm 0.001$                                 & $0.003 \pm 0.003$                                & $0.808 \pm 0.005$                                 & $0.807 \pm 0.005$                                 & $0.843 \pm 0.003$                                                                                       & $\mathbf{0.844 \pm 0.004}$                                                         \\
EO                                                    & $0.857 \pm 0.005$                                 & $0.845 \pm 0.005$                                & $0.005 \pm 0.005$                                 & $0.041 \pm 0.023$                                & $0.808 \pm 0.005$                                 & $0.805 \pm 0.005$                                 & $\mathbf{0.810 \pm 0.005}$                                                                              & $0.807 \pm 0.005$                                                                  \\
EOdds                                                 & $0.846 \pm 0.006$                                 & $0.834 \pm 0.007$                                & $0.007 \pm 0.006$                                 & $0.037 \pm 0.021$                                & $0.808 \pm 0.005$                                 & $0.807 \pm 0.004$                                 & $0.839 \pm 0.008$                                                                                       & $\mathbf{0.840 \pm 0.009}$                                                         \\ \hline
\multicolumn{9}{c}{\textbf{ACSPublicCoverage dataset}}                                                                                                                                                                                                                                                                                                                                                                                                                                                                                                                     \\ \hline
SP                                                    & $0.861 \pm 0.003$                                 & $0.851 \pm 0.003$                                & $0.001 \pm 0.001$                                 & $0.003 \pm 0.002$                                & $0.861 \pm 0.005$                                 & $0.860 \pm 0.006$                                 & $0.874 \pm 0.005$                                                                                       & $\mathbf{0.875 \pm 0.007}$                                                         \\
PE                                                    & $0.861 \pm 0.002$                                 & $0.853 \pm 0.002$                                & $0.001 \pm 0.000$                                 & $0.003 \pm 0.002$                                & $0.861 \pm 0.005$                                 & $0.860 \pm 0.005$                                 & $0.864 \pm 0.005$                                                                                       & $\mathbf{0.870 \pm 0.007}$                                                         \\
EO                                                    & $0.851 \pm 0.005$                                 & $0.843 \pm 0.004$                                & $0.002 \pm 0.002$                                 & $0.022 \pm 0.011$                                & $0.861 \pm 0.005$                                 & $0.859 \pm 0.006$                                 & $\mathbf{0.862 \pm 0.004}$                                                                              & $0.861 \pm 0.006$                                                                  \\
EOdds                                                 & $0.841 \pm 0.004$                                 & $0.833 \pm 0.004$                                & $0.003 \pm 0.002$                                 & $0.023 \pm 0.011$                                & $0.861 \pm 0.005$                                 & $0.860 \pm 0.005$                                 & $\mathbf{0.862 \pm 0.004}$                                                                              & $0.861 \pm 0.005$                                                                  \\ \hline
\multicolumn{9}{c}{\textbf{ACSIncome dataset}}                                                                                                                                                                                                                                                                                                                                                                                                                                                                                                                             \\ \hline
SP                                                    & $0.788 \pm 0.003$                                 & $0.776 \pm 0.003$                                & $0.002 \pm 0.001$                                 & $0.005 \pm 0.004$                                & $0.690 \pm 0.007$                                 & $0.715 \pm 0.010$                                 & $0.756 \pm 0.005$                                                                                       & $\mathbf{0.764 \pm 0.006}$                                                         \\
PE                                                    & $0.797 \pm 0.002$                                 & $0.785 \pm 0.002$                                & $0.001 \pm 0.001$                                 & $0.004 \pm 0.003$                                & $0.690 \pm 0.007$                                 & $0.688 \pm 0.007$                                 & $\mathbf{0.736 \pm 0.007}$                                                                              & $0.735 \pm 0.006$                                                                  \\
EO                                                    & $0.796 \pm 0.003$                                 & $0.784 \pm 0.003$                                & $0.001 \pm 0.001$                                 & $0.010 \pm 0.007$                                & $0.690 \pm 0.007$                                 & $0.685 \pm 0.006$                                 & $\mathbf{0.693 \pm 0.007}$                                                                              & $0.689 \pm 0.006$                                                                  \\
EOdds                                                 & $0.795 \pm 0.003$                                 & $0.783 \pm 0.003$                                & $0.002 \pm 0.001$                                 & $0.010 \pm 0.006$                                & $0.690 \pm 0.007$                                 & $0.688 \pm 0.007$                                 & $\mathbf{0.737 \pm 0.007}$                                                                              & $0.735 \pm 0.006$                                                                  \\ \hline
\end{tabular}
\end{table*}

When varying the unfairness tolerance $\tol$, the only input of the 
reconstruction methods that is modified is the fair model's predictions $\predictions{}$ (and the fairness information).
The fact that the reconstruction accuracy of the baseline adversary $\attackerseven$ is rather constant across variations of $\tol$ shows that the fair model's predictions $\predictions{}$ are not used a lot by the learnt attack models. 
In contrast, as our method knows exactly how to interpret the fairness information with respect to $\predictions{}$, it is able to exploit it
to significantly improve the final reconstruction accuracy.

Finally, the empirical results show that our 
reconstruction correction method is able to considerably improve the reconstruction accuracy of the training set sensitive attributes, even when the original adversary is as informed as our method. 
\subsubsection{Experiments using the ThresholdOptimizer technique}

Results of our experiments using the ThresholdOptimizer~\cite{hardt2016equality} fair post-processing method are displayed in Table~\ref{tab:postproc_results}. 
The observed trends are similar to that of the previous subsection, which demonstrates that the type of fairness intervention does not impact our framework.
One can observe that the performances of both baseline adversaries are very close. 
As he possesses more information than $\attackersix$, $\attackerseven$ always performs better on the attack set (used to train the attack models). 
However, his generalization is sometimes poorer, resulting in worse reconstruction performances when used on the target model training set. 
This may be due to the distribution of the target fair model's predictions on its own training set $\predictions{}$ being different from that on the adversary's attack set $\predictions{}_A$.

Importantly, we observe that the reconstruction correction step always improves the reconstruction accuracy. 
Indeed, the improvement obtained depends on the considered fairness metric and on the original bias of the reconstruction (which is related to the inherent bias of the original training set). 
The reconstruction accuracy improvements over the two baseline adversaries are of the same magnitude than with the ExponentiatedGradient method. 
Again, reconstruction correction using the Equal Opportunity metric offers modest improvements due to the fact that it applies to a minority of training examples.

\section{Discussion on Countermeasures}
\label{sec:countermeasures}

We have seen that the proposed reconstruction correction
is able to exploit the fairness information to significantly improve the reconstruction accuracy, even with an informed adversary. 
In this section, we discuss possible countermeasures to limit the effectiveness of the reconstruction correction step.

\subsection{Differential Privacy}

Differential Privacy (DP)~\cite{10.1007/978-3-540-79228-4_1,10.1561/0400000042} is considered to be one of the state-of-the-art methods for preventing inference attacks against machine learning models.
While it may affect the performances of a baseline adversary, DP cannot be an effective countermeasure to our proposed reconstruction correction step. 
Indeed, it is designed to ensure that the output of a mechanism does not rely too much on any single example, but rather on general patterns. 
However, statistical fairness metrics are measured over an entire dataset and do not specifically rely on individual examples. 
Thus, as our reconstruction correction method only relies on group-level statistics, DP cannot effectively affect its performances~\cite{DBLP:journals/corr/abs-1011-2511}. 

Additionally, 
DP 
is
incompatible with the strict respect of any statistical fairness measure~\cite{10.1145/3314183.3323847,1548832}. 
Indeed, releasing a model along with information regarding its strict respect of any statistical fairness constraint is intrinsically non-DP compliant. 

\subsection{Hiding the Fairness Information}\label{subsec:hiding_fairness}

Intuitive countermeasures consist in perturbing the fairness information (type of fairness metric used or unfairness tolerance parameter $\tol$).
Note that this may not be possible when a particular fairness requirement is also a legal requirement, as for the ``$80$ percent rule'' for Statistical Parity~\cite{DBLP:conf/kdd/FeldmanFMSV15} stated by the US Equal Employment Opportunity Commission (EEOC)~\cite{uniformguidelinesemployeeselection}.
When possible, releasing noisy or empty fairness information may be a reasonable defense mechanism. 
However, adversaries may still use diverse strategies to infer both the fairness metric that was optimized and the unfairness tolerance parameter. 
Depending on the adversarial knowledge, such property inference attacks~\cite{DBLP:journals/corr/abs-2005-08679} might give a good estimation to the adversary, which we can expect would still allow reasonable reconstruction correction performances from our approach. 
Indeed, recall that our proposed method only needs information regarding the model's predictions fairness and the reconstruction correction still works even if the set fairness constraint is not the one that was used for training.

Using our baseline adversaries $\attackersix$ or $\attackerseven$, a simple strategy would be to quantify the target model unfairness on the attack set $\dataset{}_A$ for the different considered metrics. Then, one can select the metric with the smallest measured unfairness, and consider that the model is fair for this metric with unfairness tolerance $\tol$ equal to the measured unfairness. 
To assess its effectiveness, we implemented this fairness information estimation strategy and performed our experiments again. 

\begin{table}[bt]
\caption{Summary of the results of our experiments using a post-processing method for fairness, for the simple countermeasure of not revealing the fairness information. {\upshape \emph{Reconstruction results have to be compared with those of Table~\ref{tab:postproc_results}}}}
\label{tab:tentative_countermeasure_postprocessing}
\centering
\begin{tabular}{ccccc}
\hline
\multicolumn{1}{c|}{\multirow{2}{*}{\textbf{Metric}}} & \multicolumn{2}{c|}{\textbf{Estimated Constraint}}                                                                                                                                                    & \multicolumn{2}{c}{\textbf{\begin{tabular}[c]{@{}c@{}}Corrected Reconstr.\\ (Estimated Constraint)\end{tabular}}} \\ \cline{2-5} 
\multicolumn{1}{c|}{}                                 & \multicolumn{1}{c|}{\textit{\textbf{\begin{tabular}[c]{@{}c@{}}Metric\\ Detect.\end{tabular}}}} & \multicolumn{1}{c|}{\textit{\textbf{\begin{tabular}[c]{@{}c@{}}Average \\ Tolerance\end{tabular}}}} & \multicolumn{1}{c|}{\textit{\textbf{$\attackersix$}}}             & \textit{\textbf{$\attackerseven$}}            \\ \hline
\multicolumn{5}{c}{\textbf{UCI Adult Income dataset}}                                                                                                                                                                                                                                                                                                                             \\ \hline
SP                                                    & $0.95$                                                                                          & $0.004 \pm 0.003$                                                                                   & $0.848 \pm 0.009$                                                 & $0.856 \pm 0.011$                             \\
PE                                                    & $0.97$                                                                                          & $0.003 \pm 0.002$                                                                                   & $0.841 \pm 0.006$                                                 & $0.843 \pm 0.007$                             \\
EO                                                    & $0.26$                                                                                          & $0.018 \pm 0.010$                                                                                   & $0.829 \pm 0.012$                                                 & $0.828 \pm 0.013$                             \\
EOdds                                                 & $0.00$                                                                                          & $0.005 \pm 0.005$                                                                                   & $0.841 \pm 0.006$                                                 & $0.843 \pm 0.007$                             \\ \hline
\multicolumn{5}{c}{\textbf{ACSPublicCoverage dataset}}                                                                                                                                                                                                                                                                                                                            \\ \hline
SP                                                    & $1.00$                                                                                          & $0.002 \pm 0.002$                                                                                   & $0.873 \pm 0.005$                                                 & $0.873 \pm 0.009$                             \\
PE                                                    & $1.00$                                                                                          & $0.003 \pm 0.002$                                                                                   & $0.863 \pm 0.005$                                                 & $0.865 \pm 0.007$                             \\
EO                                                    & $0.28$                                                                                          & $0.008 \pm 0.005$                                                                                   & $0.862 \pm 0.005$                                                 & $0.862 \pm 0.005$                             \\
EOdds                                                 & $0.00$                                                                                          & $0.002 \pm 0.002$                                                                                   & $0.868 \pm 0.006$                                                 & $0.869 \pm 0.007$                             \\ \hline
\multicolumn{5}{c}{\textbf{ACSIncome dataset}}                                                                                                                                                                                                                                                                                                                                    \\ \hline
SP                                                    & $0.80$                                                                                          & $0.003 \pm 0.003$                                                                                   & $0.743 \pm 0.026$                                                 & $0.754 \pm 0.020$                             \\
PE                                                    & $0.86$                                                                                          & $0.003 \pm 0.003$                                                                                   & $0.729 \pm 0.016$                                                 & $0.728 \pm 0.016$                             \\
EO                                                    & $0.73$                                                                                          & $0.008 \pm 0.006$                                                                                   & $0.704 \pm 0.019$                                                 & $0.700 \pm 0.020$                             \\
EOdds                                                 & $0.00$                                                                                          & $0.002 \pm 0.002$                                                                                   & $0.723 \pm 0.021$                                                 & $0.721 \pm 0.022$                             \\ \hline
\end{tabular}
\end{table}

Results for the experiments using the
ThresholdOptimizer~\cite{hardt2016equality} method are reported in Table~\ref{tab:tentative_countermeasure_postprocessing}. 
More precisely, we report the performances of the fairness constraint estimation process, namely the rate of correct metric identification, and the average unfairness tolerance inferred. 
Due to the simple estimation process, the Equalized Odds metric can never be identified as its violation is the maximum of the Predictive Equality and Equal Opportunity violations (hence it can never be the smallest value). 
However, for the other metrics we observe that even this simple estimation process is often able to correctly identify the optimized metric.

Several trends can be noted when comparing the reconstruction results with those of Table~\ref{tab:postproc_results}, in which the reconstruction correction is done using the actual fairness constraint. 
A first situation occurs when the fairness constraint is correctly inferred, which is the case in most experiments using the Statistical Parity or Predictive Equality metrics. 
For instance, when using the ACSIncome dataset, the Statistical Parity metric was correctly identified 
in all our experiments. 
In this scenario, the reconstruction correction still brings important improvement - slightly
weakened by the fact that the estimated tolerance is usually not as tight as the actual one.
A second interesting situation is when the fairness metric is not correctly identified, which is the case for all experiments using the Equalized Odds metric. 
Nonetheless, the fairness information estimation process can still come with a valid fairness constraint (even if it is not the one that was optimized during training), which can effectively be leveraged by the reconstruction correction step. 
When the fairness estimation proposes a metric more informative (in terms of number of involved examples) than the actual one (\emph{e.g.}, for some experiments with the Equal Opportunity metric), the reconstruction improvement can sometimes be better than with the original constraint.
For instance, consider the experiment using the UCI Adult Income dataset with the Equal Opportunity metric. 
In $74\%$ of the runs, the fairness constraint estimation process came up with a Predictive Equality constraint. 
Even though this is not the actual constraint that was optimized during training, this constraint is approximately valid and the corresponding metric relates to a greater number of examples.
As a consequence and somewhat counter-intuitively, the final reconstruction is better than with the actual constraint (see Table~\ref{tab:postproc_results}). 
Finally, one important drawback of the fairness estimation process is that the performances of the reconstruction correction step are more variable as shown by greater standard deviation values. 

\begin{figure}[tb]
    \centering
    \includegraphics[width=0.41\textwidth]{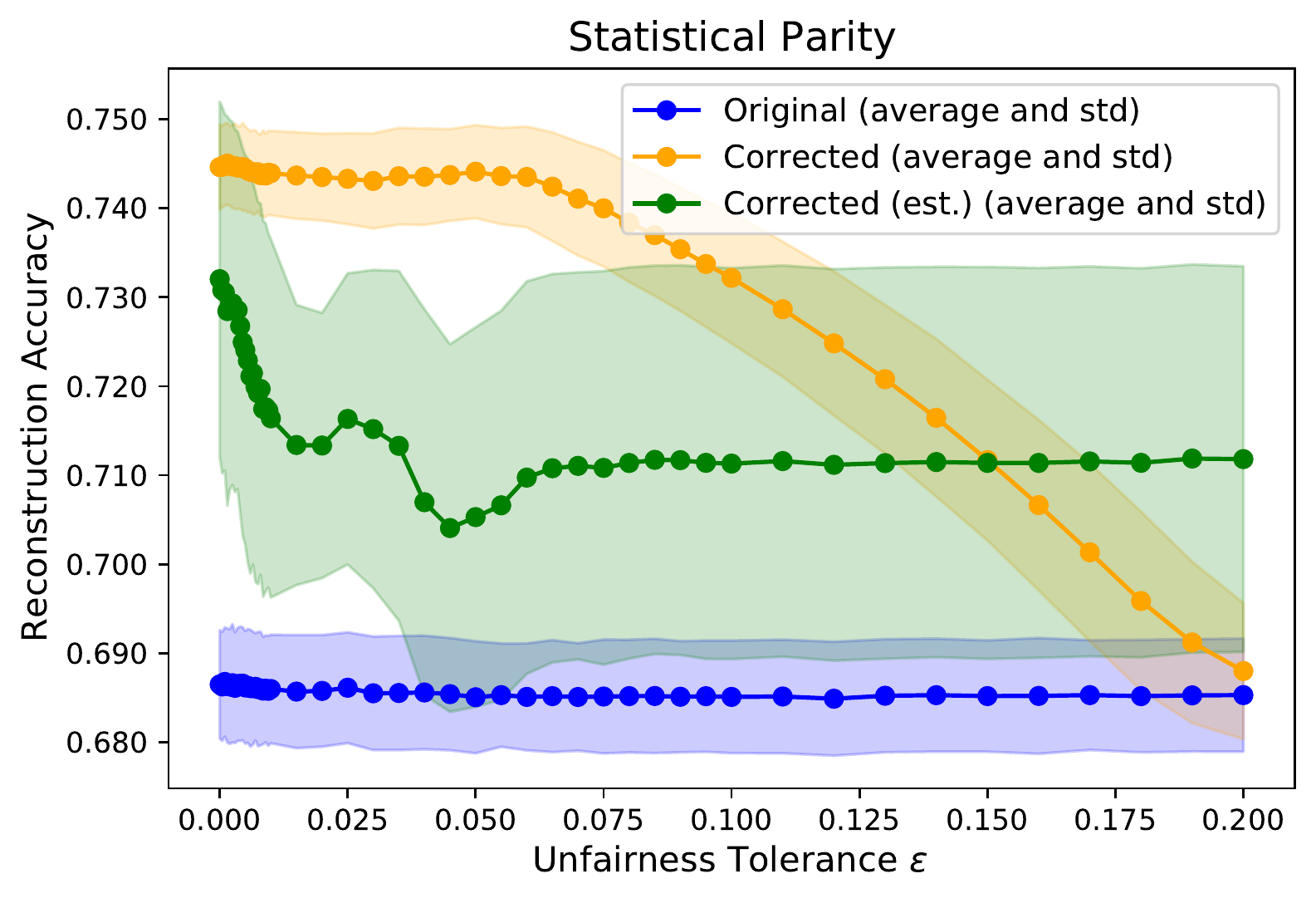}
    \caption{Original (adversary $\attackerseven$), corrected (from actual fairness constraint, and from estimated one (\texttt{est.})) reconstruction quality, for experiments using the ACSIncome dataset}
    \label{fig:reconstruction_countermeasure_inproc_example}
\end{figure}

Results using the ExponentiatedGradient method~\cite{DBLP:conf/icml/AgarwalBD0W18} are provided in Fig.~\ref{fig:reconstruction_countermeasure_inproc_example} for the experiment using the ACSIncome dataset with the Statistical Parity metric and baseline adversary $\attackerseven$, and in Appendix~\ref{appendix:countermeasures_other_expes} for the remaining ones. 
They show similar trends as those using the ThresholdOptimizer method: estimating the fairness constraint still allows for good reconstruction correction performances but leads to a greater variability in the final reconstruction accuracy. 
Here also, inferring a fairness constraint different from the actual one can improve reconstruction correction, especially when the original tolerance is larger than the actual bias contained in the data (\emph{i.e.}, large values of $\tol$). 
In such cases, the adversary's baseline reconstruction already meets the actual fairness requirement and the reconstruction correction process cannot improve it. 
In contrast, the fairness constraint estimation process can 
infer a tighter value, allowing some reconstruction improvement.

Overall, we see that the knowledge of the actual fairness constraint is not necessary as estimations can provide comparable-quality reconstruction correction performances. 
Using the proposed fairness constraint estimation process, we provide in Appendix~\ref{appendix:preproc_expes} additional reconstruction experiments using a pre-processing method for enhancing fairness. 
Results demonstrate the effectiveness of the proposed reconstruction correction approach, even when fairness metrics are not directly optimized and no fairness information is available.

\section{Conclusion}
\label{sec:conclusion}

In this work, we have proposed a novel approach using declarative programming to improve the reconstruction performances of any baseline adversary by incorporating user-defined constraints. 
While the general problem may be computationally challenging, we have demonstrated that in the case of statistical fairness metrics (and, more generally, group-level constraints), it can be reformulated and solved efficiently.
In addition, our thorough experimental study shows that due to the use of the sensitive attribute information to ensure fairness of the built model, fairness-enhancing learning techniques inherently leak information about it. 
Even if such information is at the group level, it can be leveraged by an adversary to improve baseline reconstructions of the sensitive attributes. 
Furthermore, the tighter the fairness requirement, the more significant the reconstruction improvement.

We additionally observed that, even if the fairness information is not available, an adversary can still try to infer it, 
and obtain good (and sometimes, even better) reconstruction correction performances.
While the fairness information is simply an input of our proposed reconstruction correction component, this finding demonstrates the applicability of our approach. 
It also illustrates the fact that due to their use of the sensitive attributes information, statistical fairness metrics intrinsically conflict with protecting the privacy of such attributes.

Future work includes combining our reconstruction correction attack with different baseline adversaries, optimizing the adversary confidence vector $\confidencevector$ processing as well as applying our framework in the wider context of non-binary sensitive attributes.
One of the key points of our framework is the declarative nature of the reconstruction correction step, which allows considering a wide range of constraints. 
Extending our proposed pipeline to improve baseline reconstruction attacks by enforcing other constraints (\emph{e.g.}, proportion constraints, rate constraints, \ldots) is also an interesting research direction.

\bibliographystyle{IEEEtran}
\bibliography{IEEEabrv,references}

\appendices

\section{Reconstruction Correction Models for Multi-Valued Sensitive Attributes} \label{appendix:multi-valued}
\def\sensattrcard{\lvert \sensattrdistribution \rvert}

In this appendix section, we discuss the most general setting in which the sensitive attribute is multi-valued and takes one of $\sensattrcard$ values (hence effectively defining $\sensattrcard$ protected groups). 
We explain how both models can be extended to handle multi-valued sensitive attributes reconstruction and discuss the complexity cost induced by this extension.
We begin with the general reconstruction correction model, which is suitable to encode any constraint on the protected attributes. 
We then treat the efficient model, which can be used to encode any rate constraints on the protected attributes (such as, but not restricted to, statistical fairness constraints). 

\subsection{General Reconstruction Correction Model}

The general reconstruction correction model $\generalmodel$ uses exactly one decision variable to encode each training example's sensitive attribute. 
Extension to the general multi-valued sensitive attributes case hence requires modifying the domains of such variables to match that of the sensitive attributes (with $\sensattrcard$ different possible values). 
The $\nexamples$ decision variables now have domain of cardinality $\sensattrcard$. 
The objective function sums the (weighted) changes in the adversary's sensitive attributes guess, as was done in the binary case in~(\ref{line1}).
$\sensattrcard{}$~constraints ensure that there is at least one example from each protected group (as was done with~(\ref{line2}) and~(\ref{line3}) for the binary sensitive attribute setting).
Finally, one fairness constraint is declared for each protected group (sensitive attribute value), ensuring that its positive prediction rate is no further than $\tol$ from that of the entire dataset (as was done with~(\ref{line4}) and~(\ref{line5}) for the binary sensitive attribute setting).

Overall, the size of the search space of $\generalmodel$ is $O(\sensattrcard{}^\nexamples{})$, which generalizes the binary sensitive attribute case for which it was $O(2^\nexamples{})$. 

\subsection{Efficient Model for Statistical Fairness}

The efficient reconstruction correction model $\efficientmodel$ uses one decision variable to count the number of changes from one sensitive attribute value to another, for each pair of sensitive attributes values. 
Extension to the general multi-valued sensitive attributes case hence requires declaring $O(\sensattrcard{}^2)$ variables. 
To ensure that each example is counted only once, $O(\sensattrcard{})$ constraints must be declared.
Furthermore, to quantify the total cost of the performed changes, $O(\sensattrcard{}^2)$ \texttt{element} constraints have to be summed in the objective function, as was performed in~(\ref{line21}) in the binary sensitive attributes case.
$\sensattrcard{}$~constraints ensure that there is at least one example from each protected group (as was done with~(\ref{line22}) and~(\ref{line23}) for the binary sensitive attribute setting).
Finally, one fairness constraint is declared for each protected group (sensitive attribute value), ensuring that its positive prediction rate is no further than $\tol$ from that of the entire dataset (as was done with~(\ref{line24}) and~(\ref{line25}) for the binary sensitive attribute setting).

Overall, the size of the search space of $\efficientmodel$ is $O(\nexamples{}^{\sensattrcard{}^2})$, which generalizes the binary sensitive attribute case for which it was $O(\nexamples{}^4)$. 

\section{Detailed Results: Target Models Performances}

\label{appendix:target_models_perfs}

\begin{figure*}[ht]
    \begin{center}

    \includegraphics[width=\figwidth\textwidth]{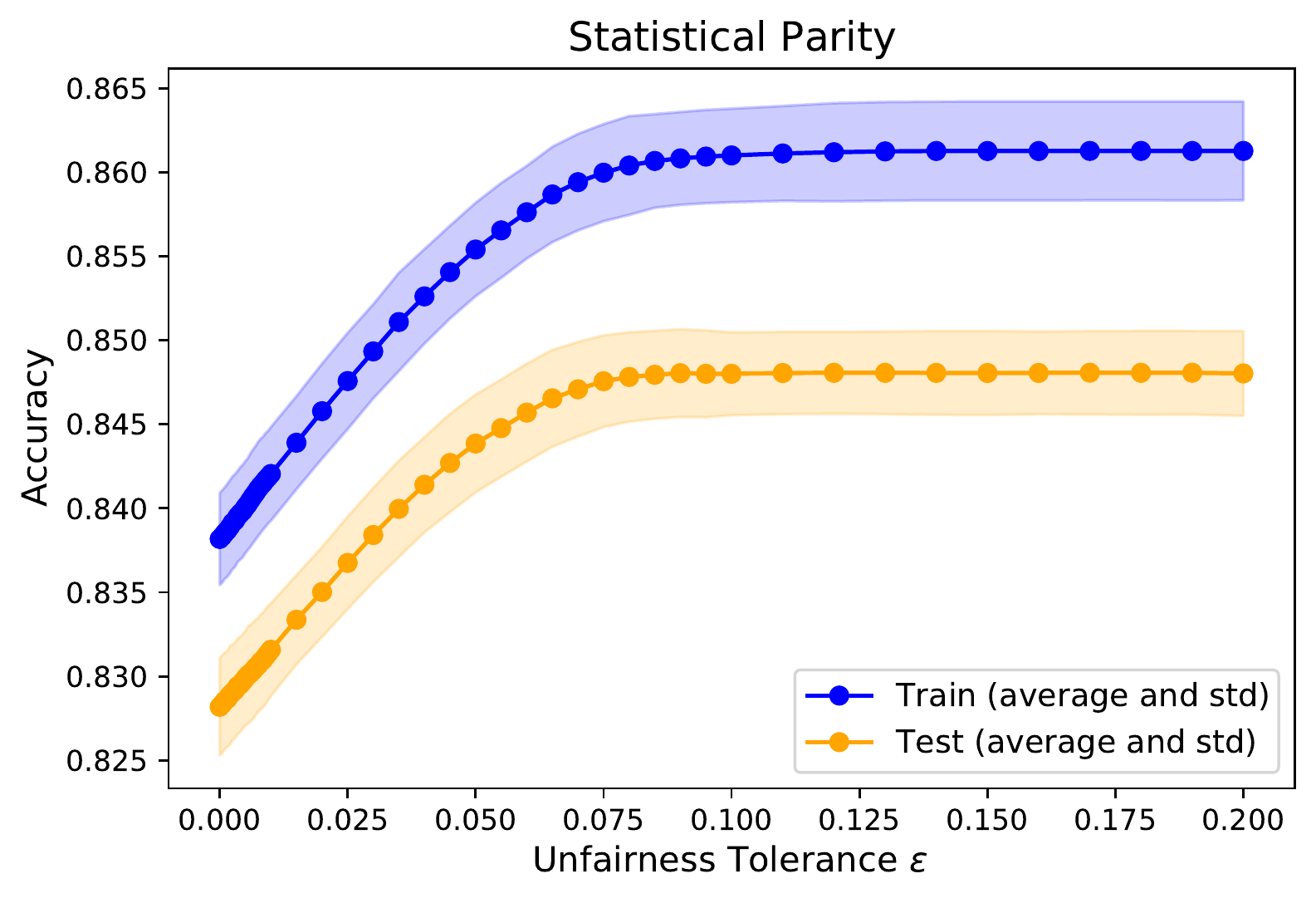}
   \includegraphics[width=\figwidth\textwidth]{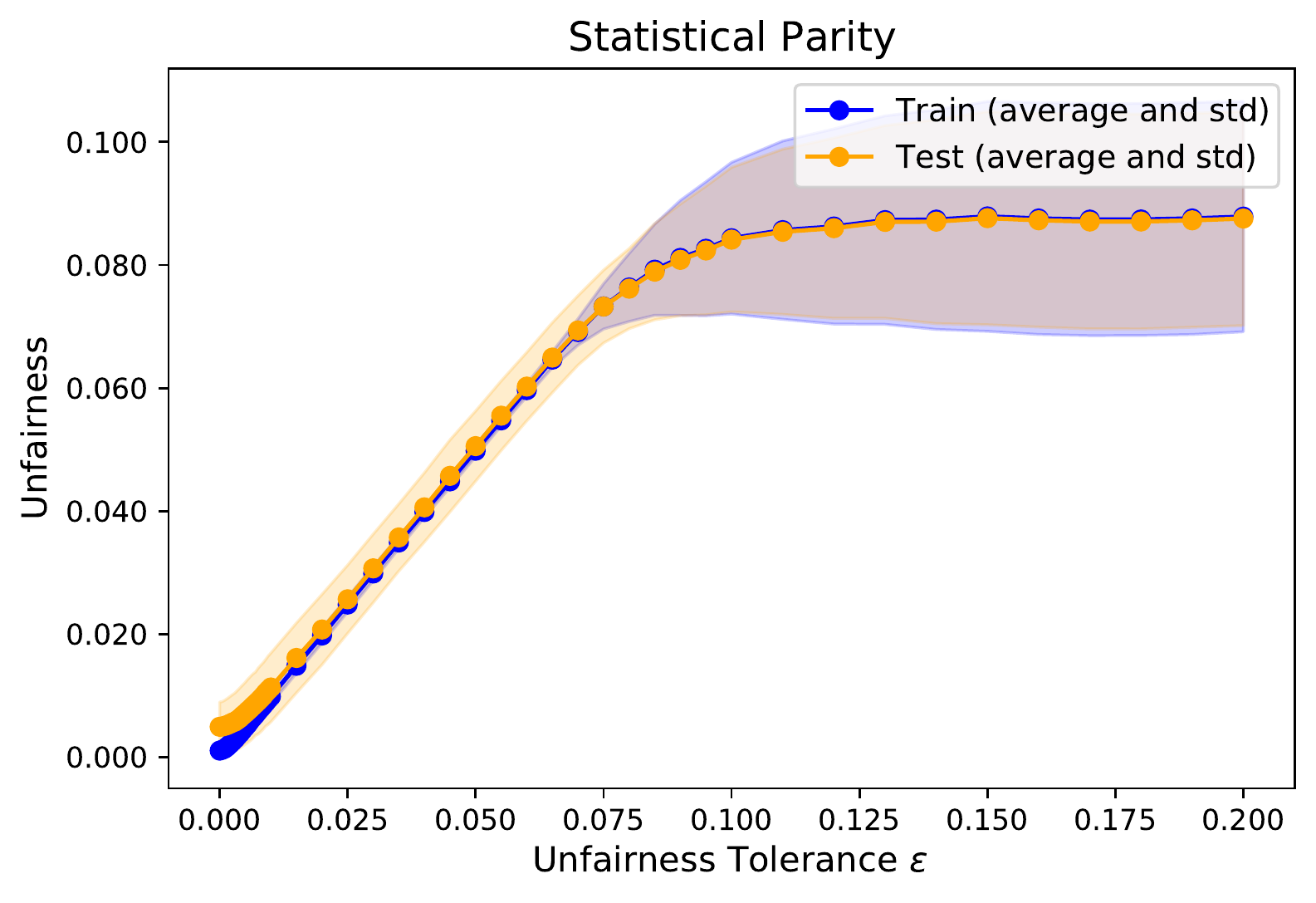}
   \vskip 0.0in
    \includegraphics[width=\figwidth\textwidth]{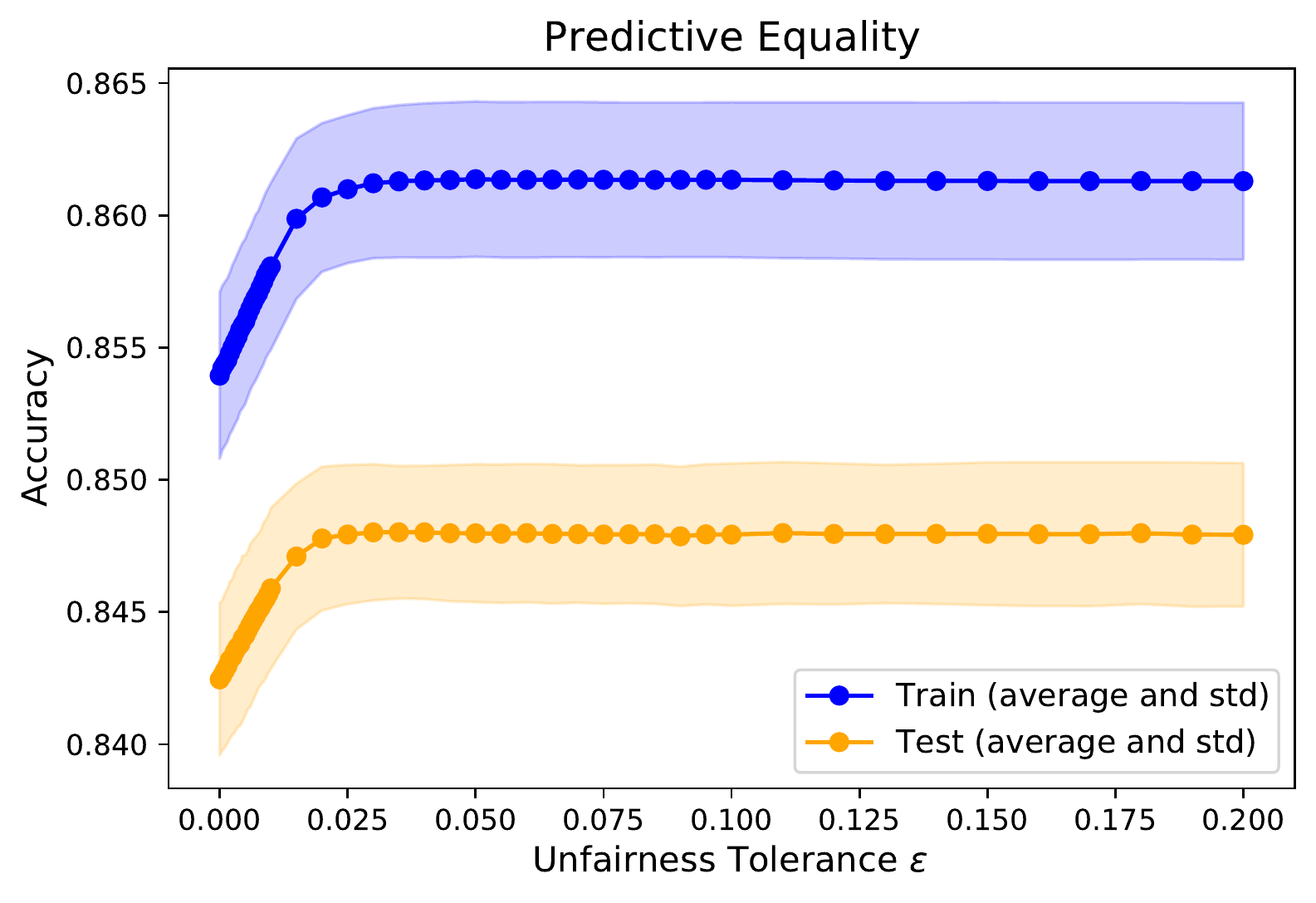}
   \includegraphics[width=\figwidth\textwidth]{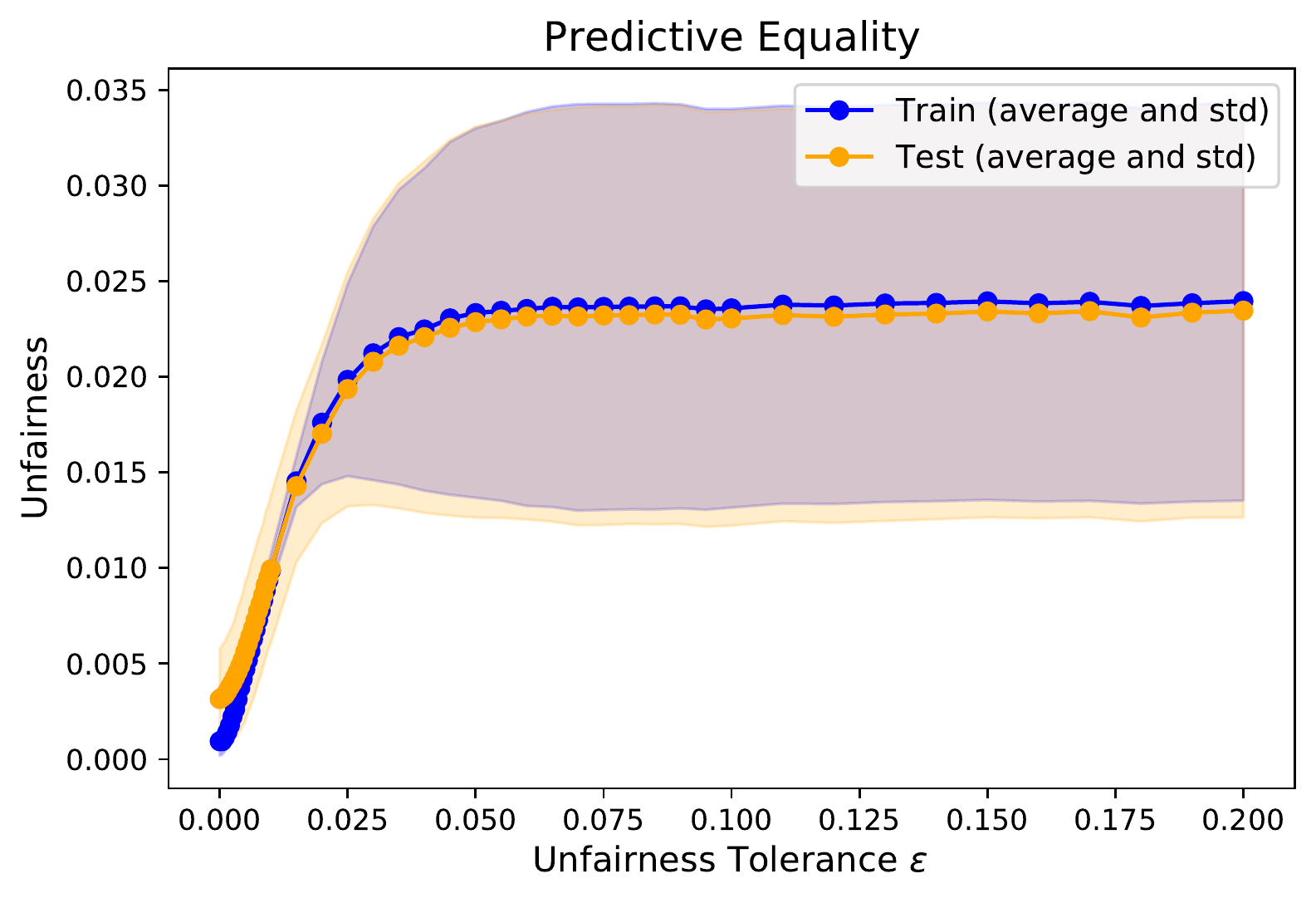}
    \vskip 0.0in
    \includegraphics[width=\figwidth\textwidth]{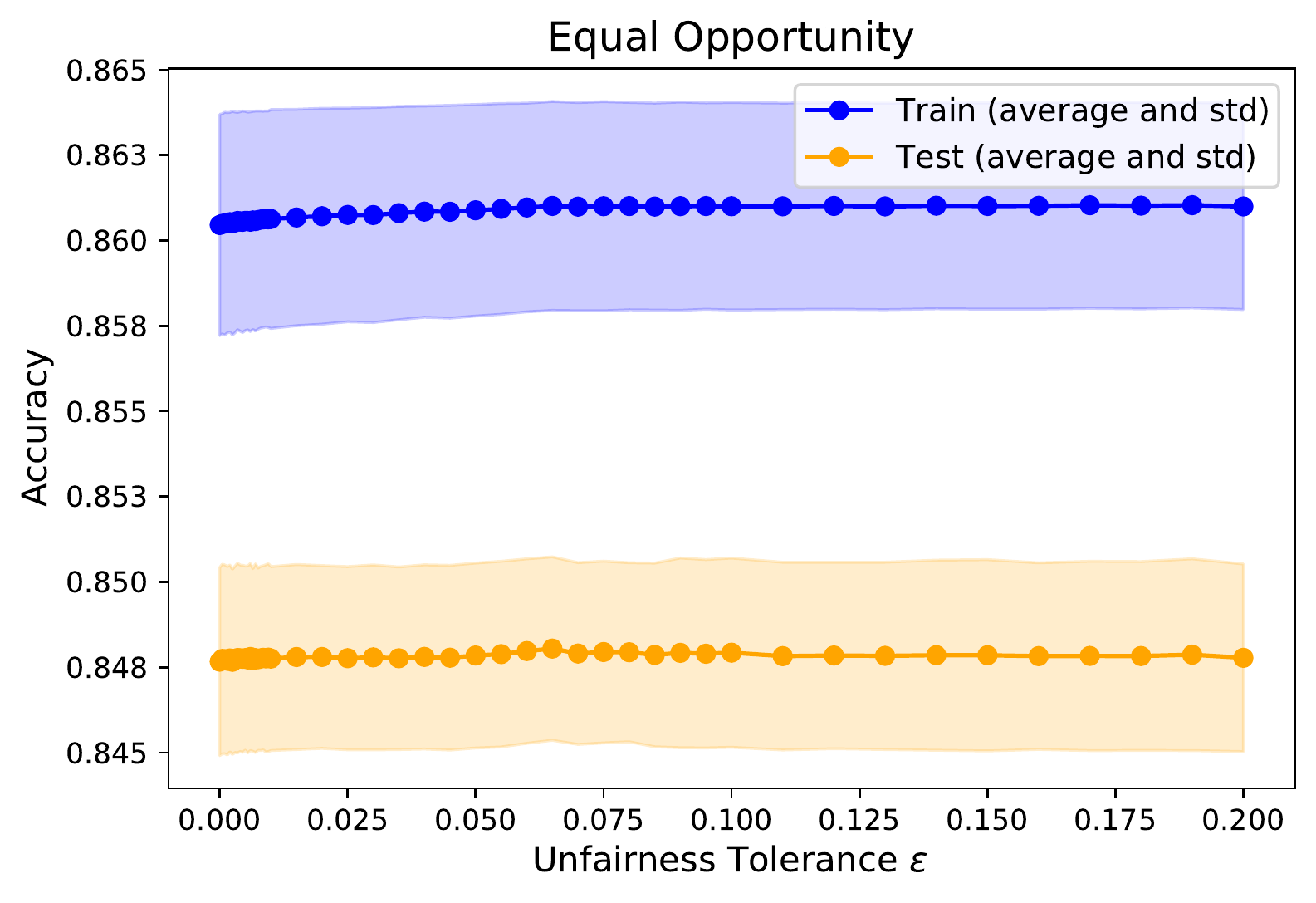}
   \includegraphics[width=\figwidth\textwidth]{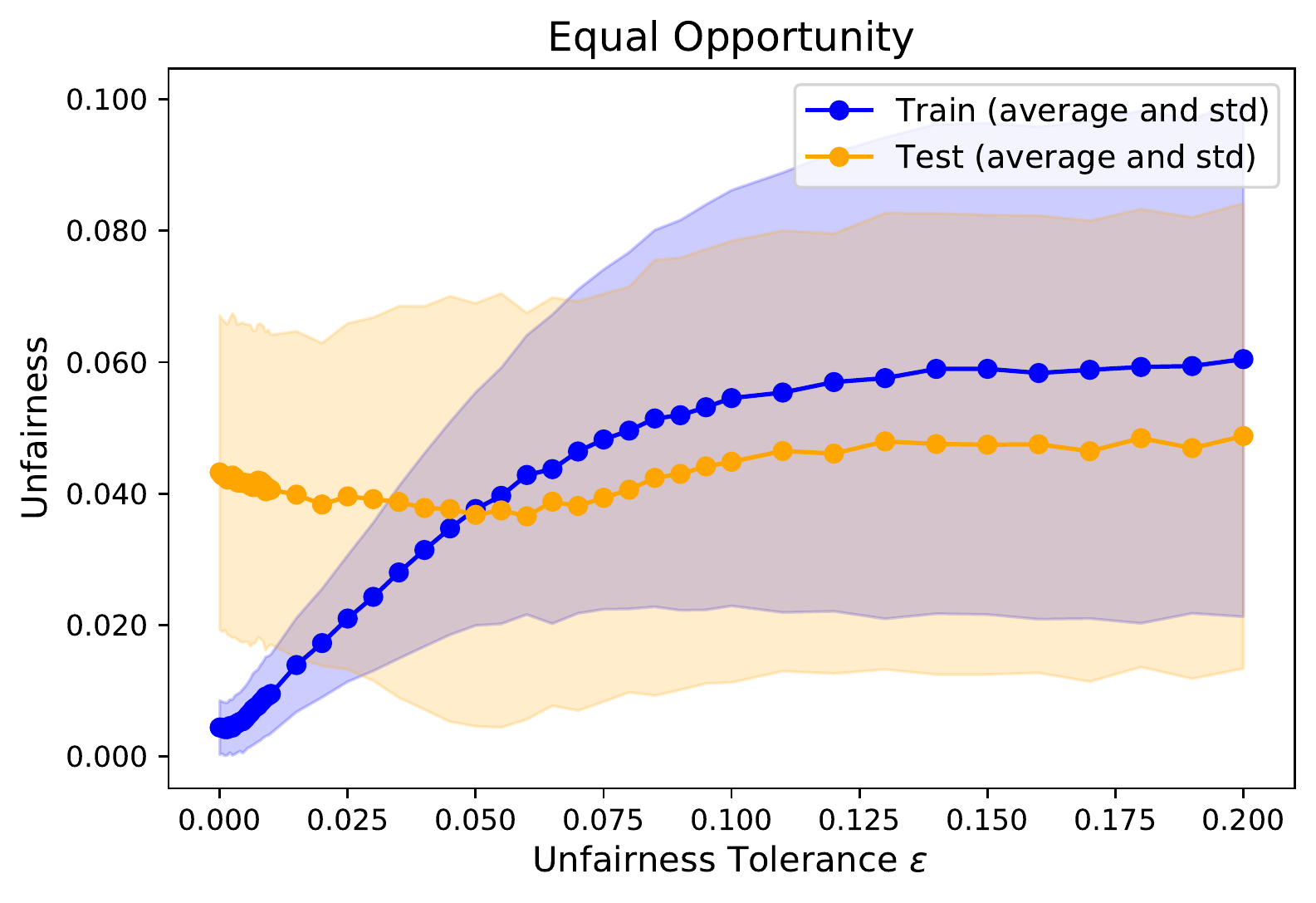}
   \vskip 0.0in
    \includegraphics[width=\figwidth\textwidth]{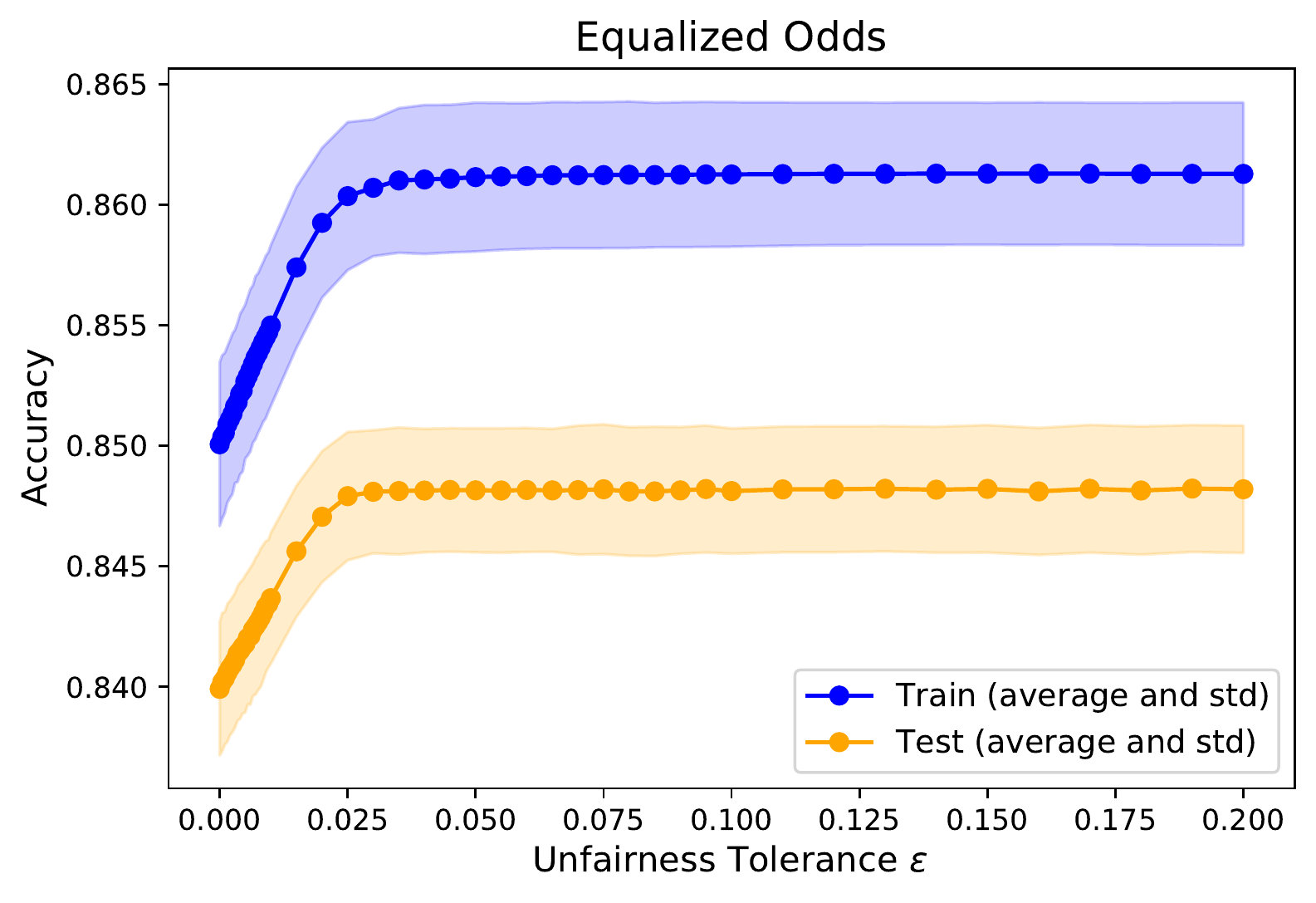}
   \includegraphics[width=\figwidth\textwidth]{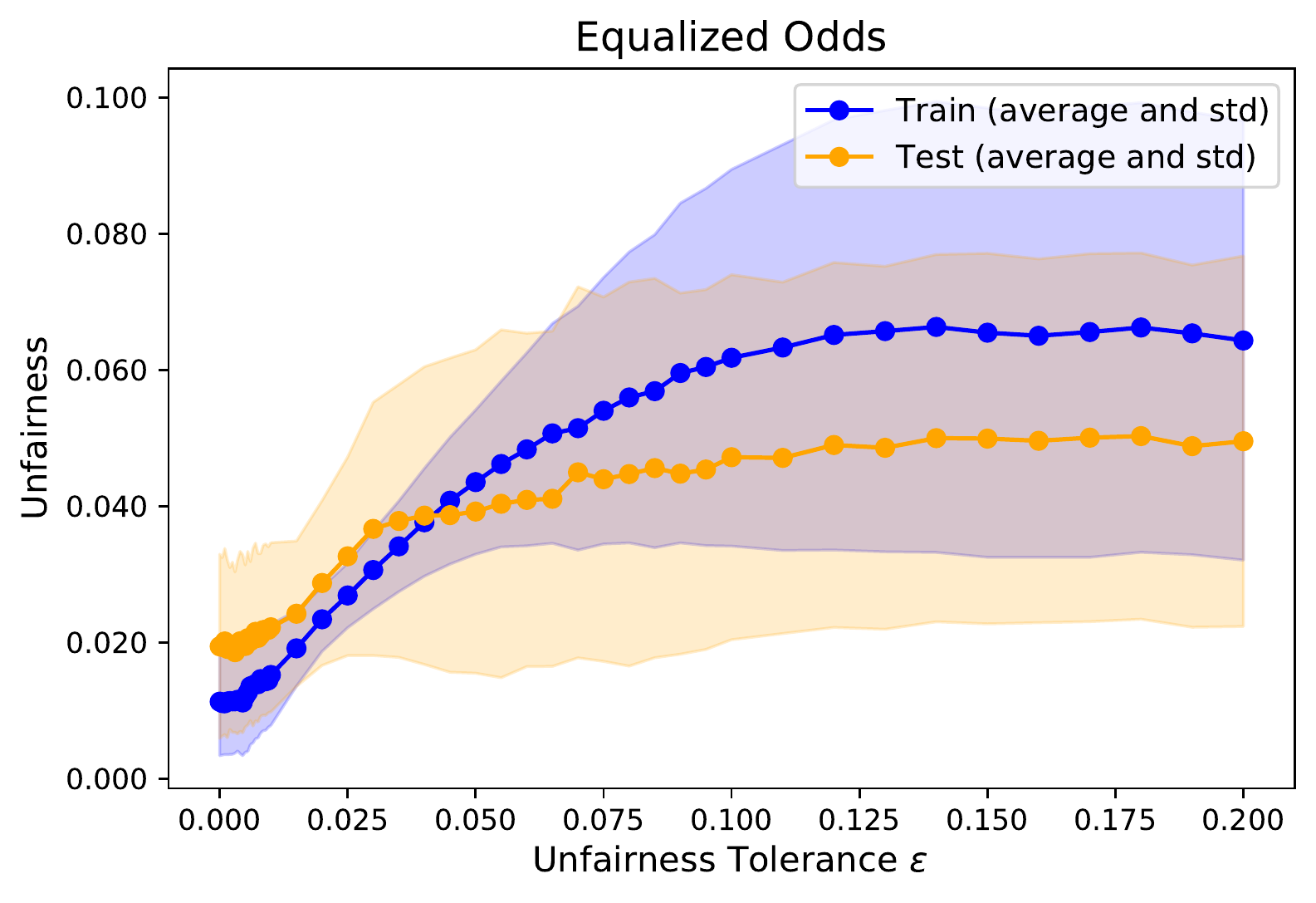}

    \end{center}
    \caption{Target models performances for our experiments using the UCI Adult Income dataset.}
\label{fig:target_perfs_adult}
\end{figure*}

\begin{figure*}[ht]
    \begin{center}

    \includegraphics[width=\figwidth\textwidth]{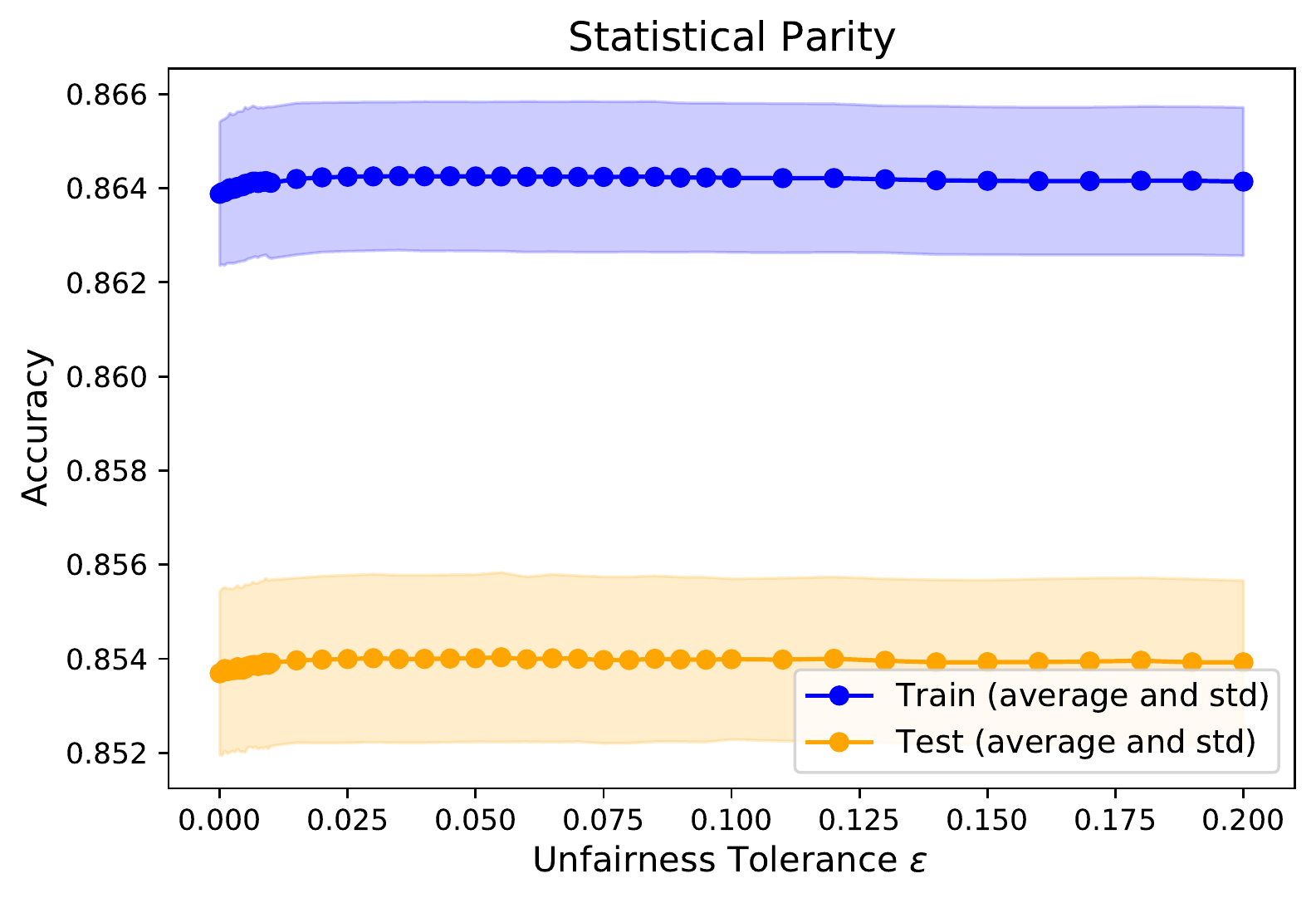}
   \includegraphics[width=\figwidth\textwidth]{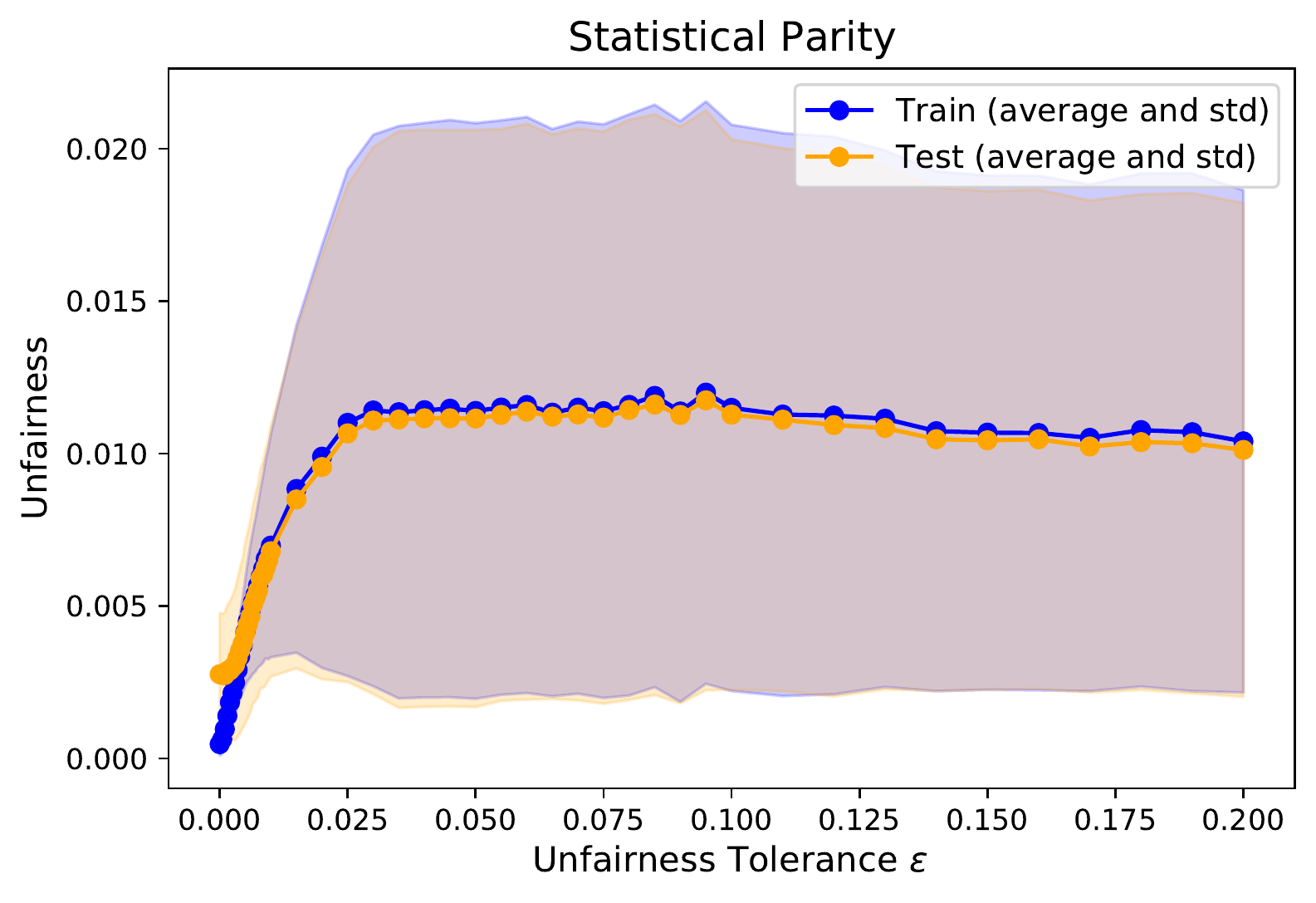}
   \vskip 0.0in
    \includegraphics[width=\figwidth\textwidth]{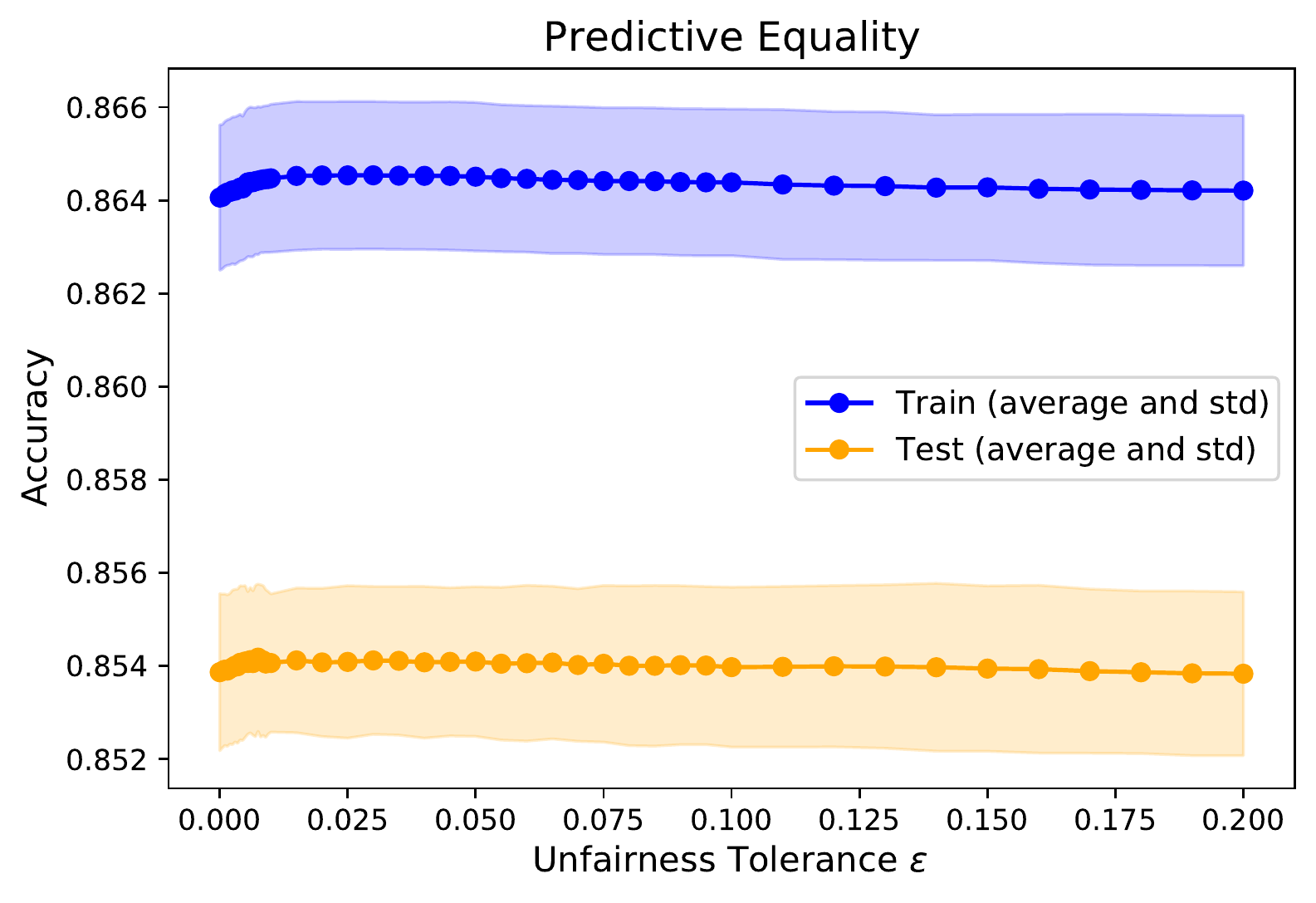}
   \includegraphics[width=\figwidth\textwidth]{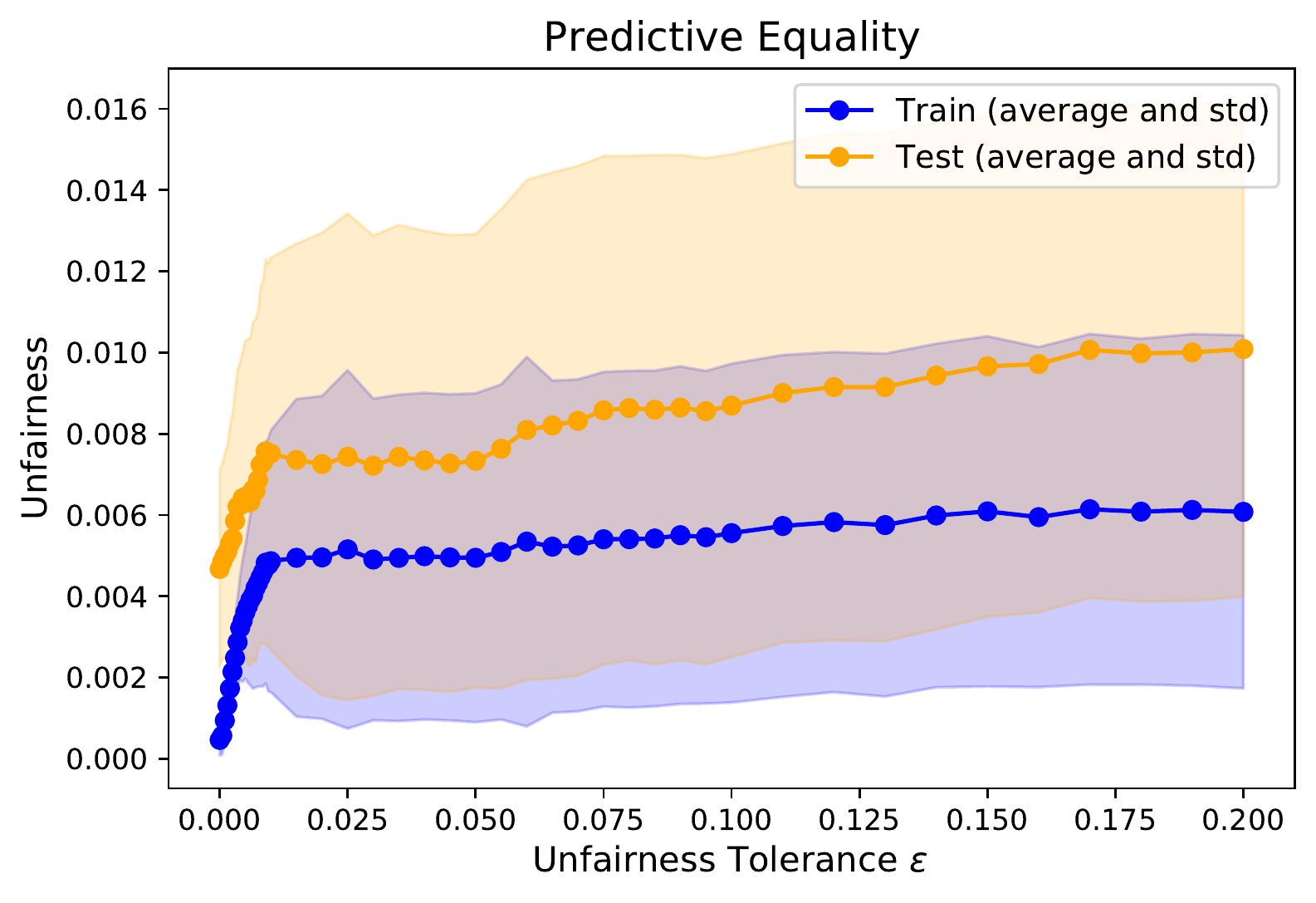}
    \vskip 0.0in
    \includegraphics[width=\figwidth\textwidth]{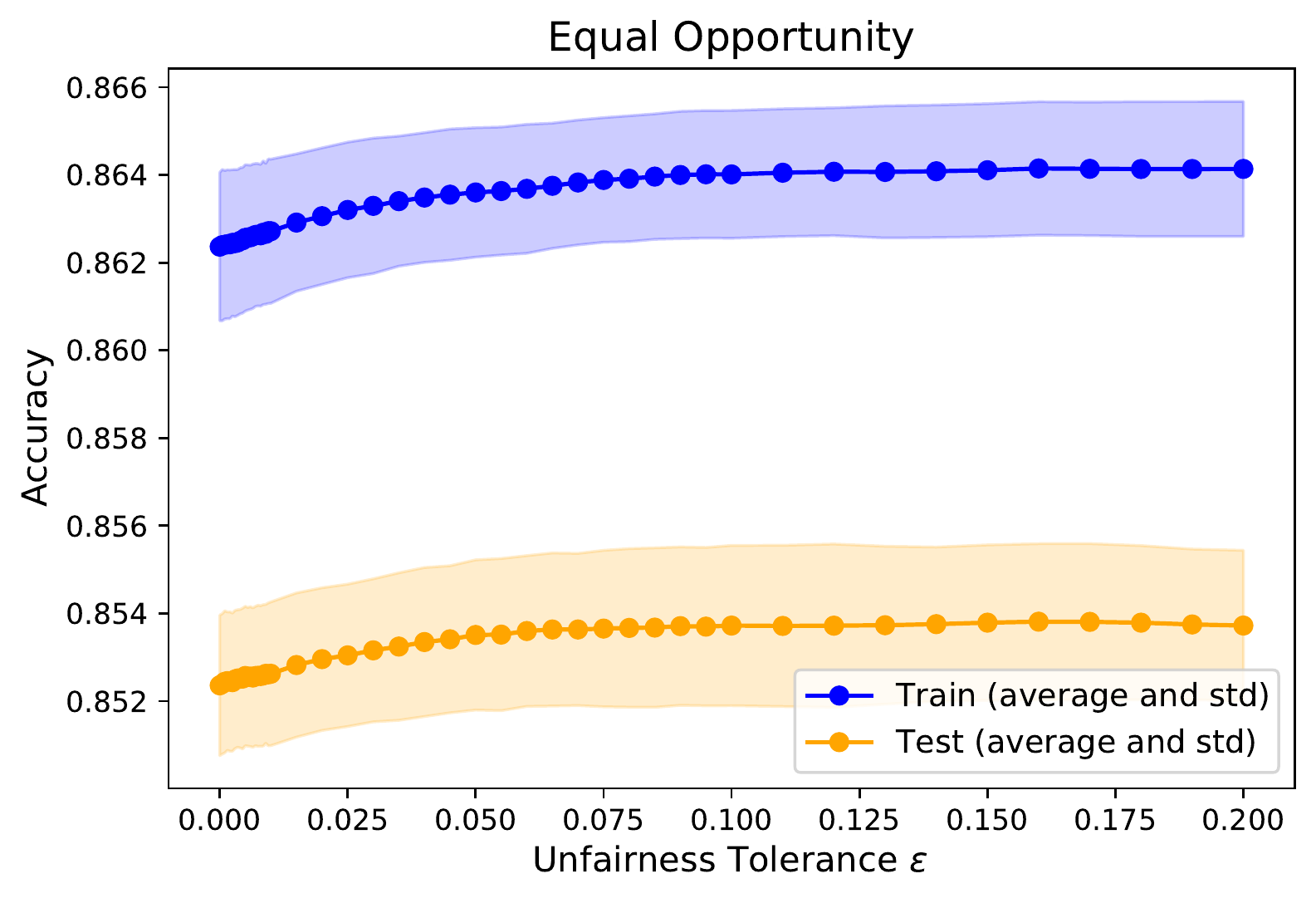}
   \includegraphics[width=\figwidth\textwidth]{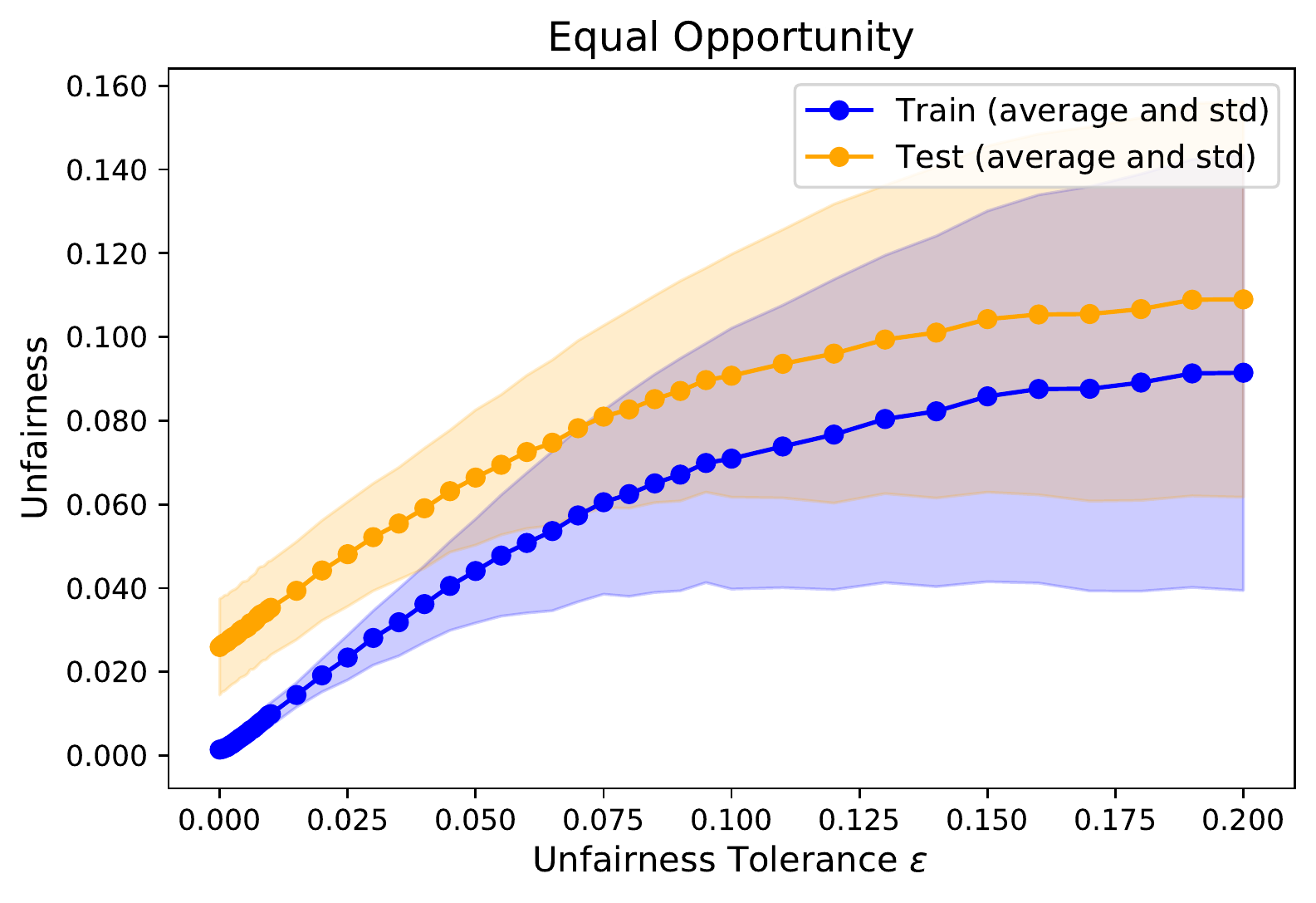}
   \vskip 0.0in
    \includegraphics[width=\figwidth\textwidth]{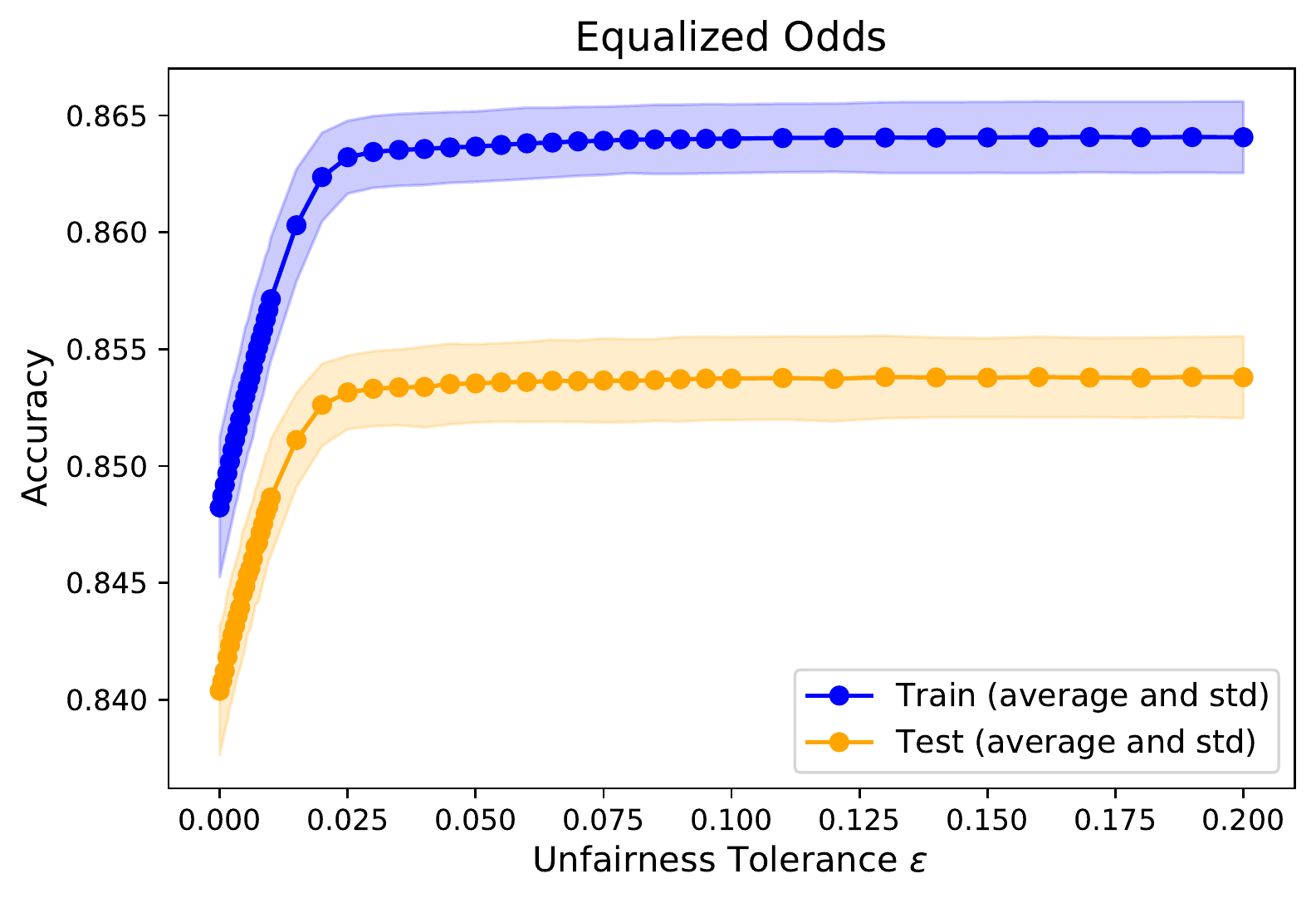}
   \includegraphics[width=\figwidth\textwidth]{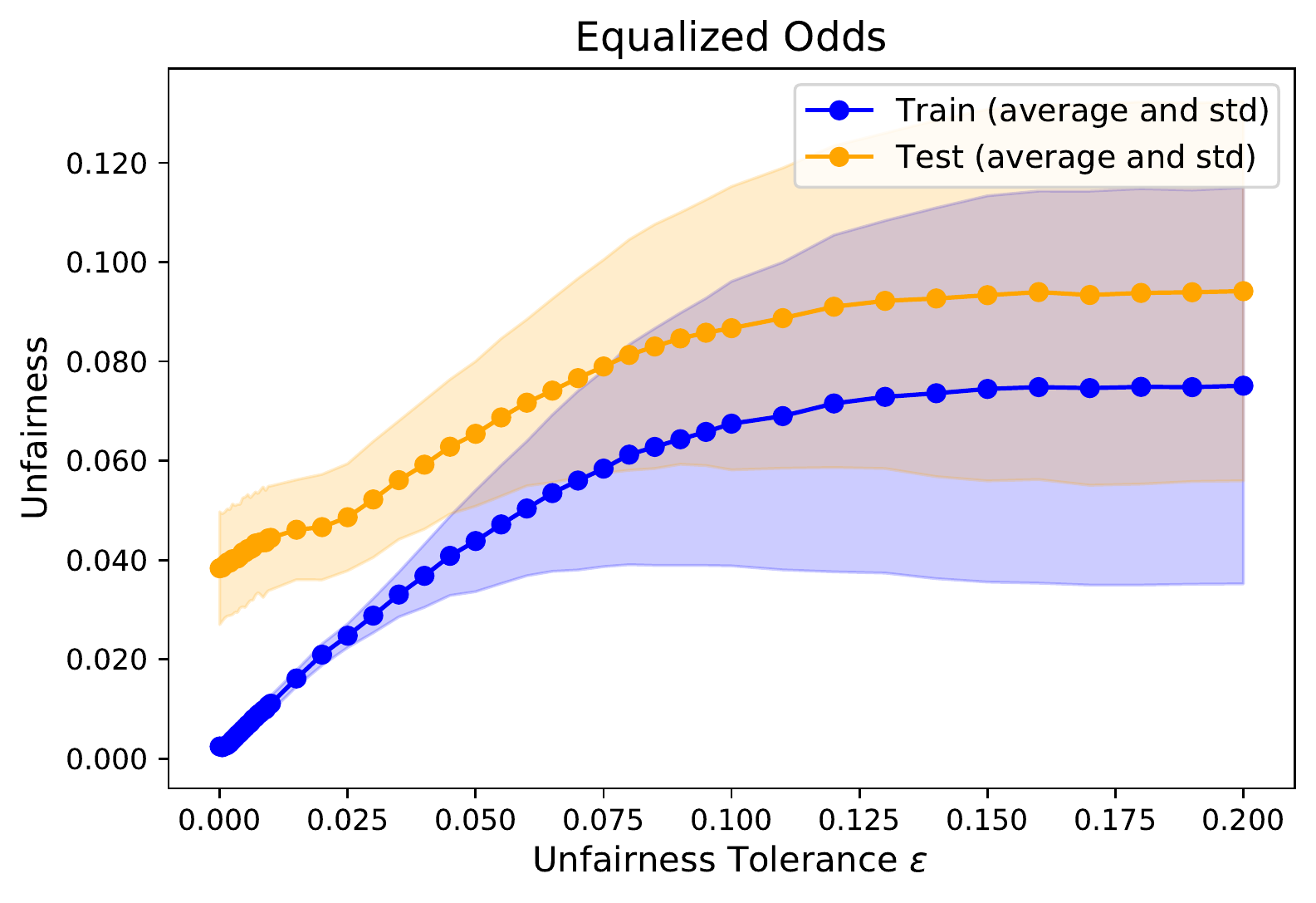}

    \end{center}
    \caption{Target models performances for our experiments using the ACSPublicCoverage dataset.}
\label{fig:target_perfs_acspubliccoverage}
\end{figure*}

\begin{figure*}[ht]
    \begin{center}

    \includegraphics[width=\figwidth\textwidth]{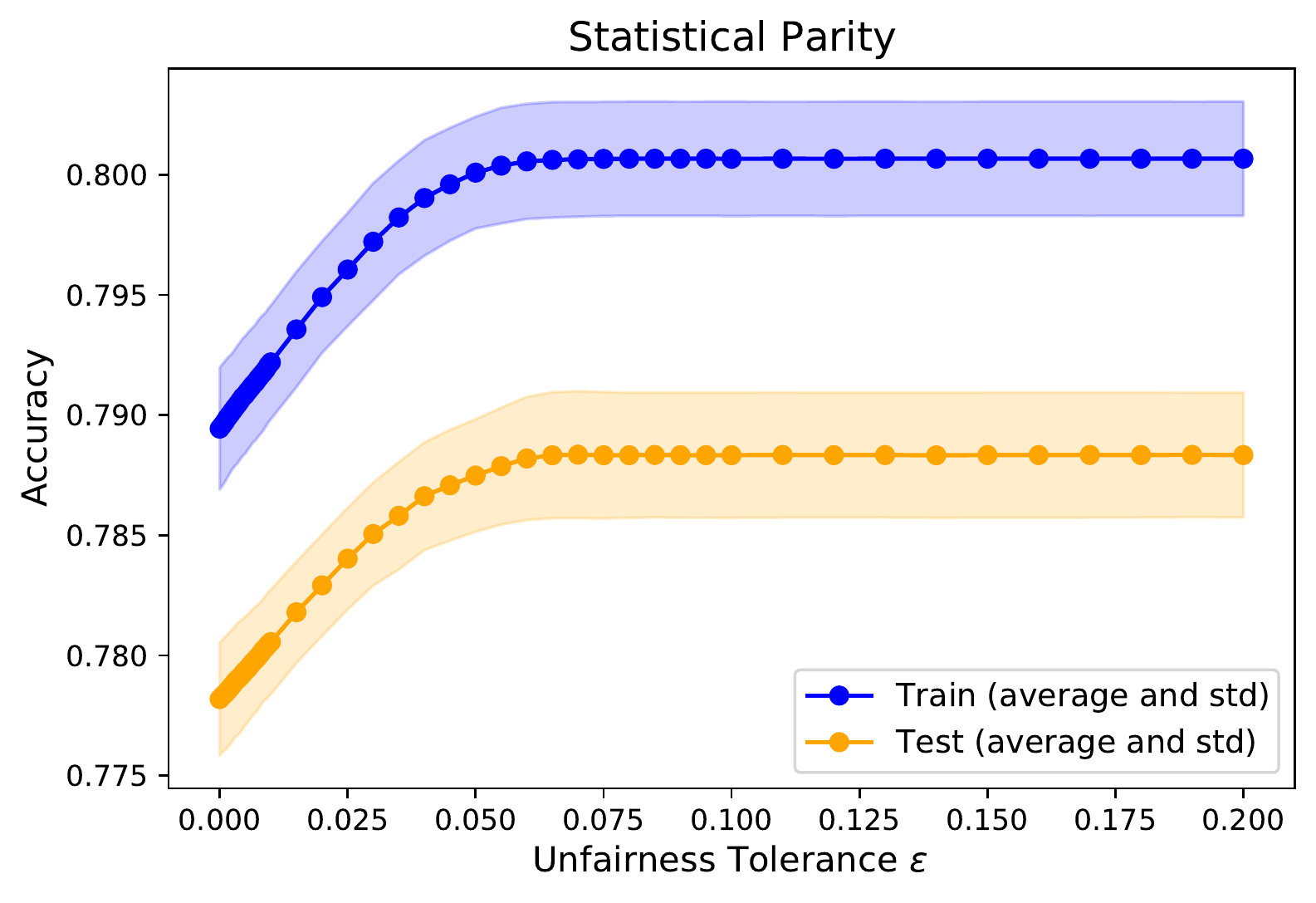}
   \includegraphics[width=\figwidth\textwidth]{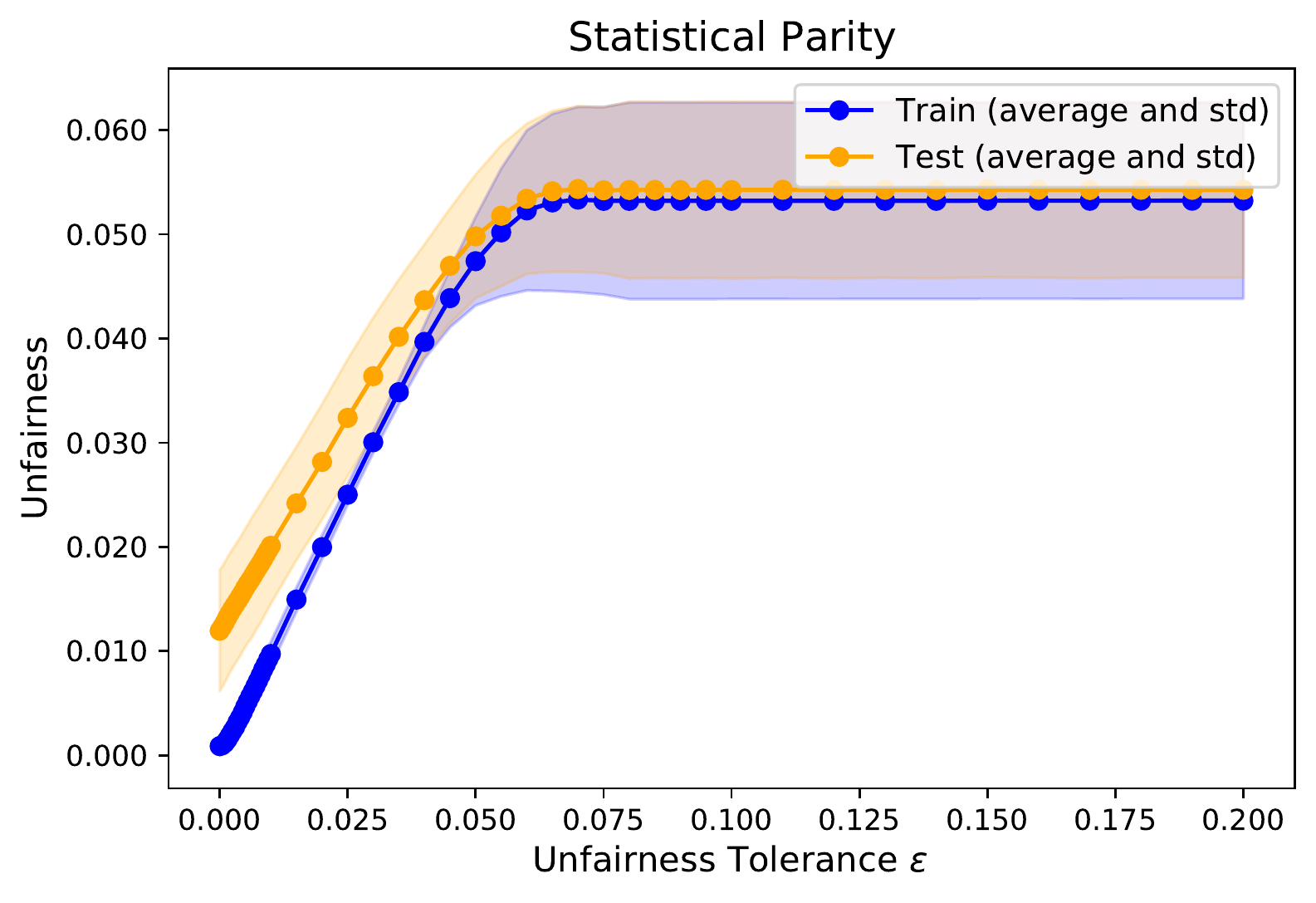}
   \vskip 0.0in
    \includegraphics[width=\figwidth\textwidth]{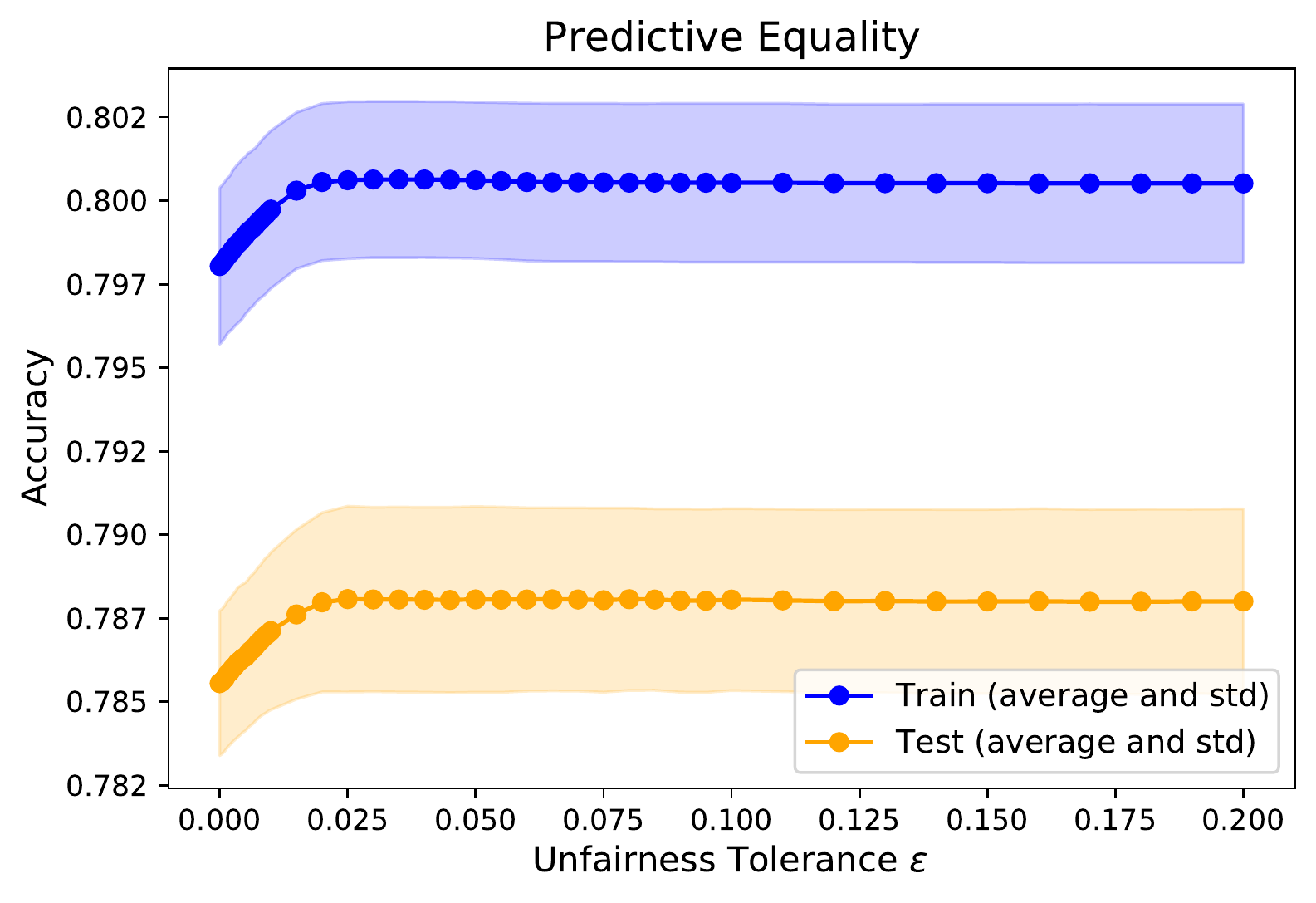}
   \includegraphics[width=\figwidth\textwidth]{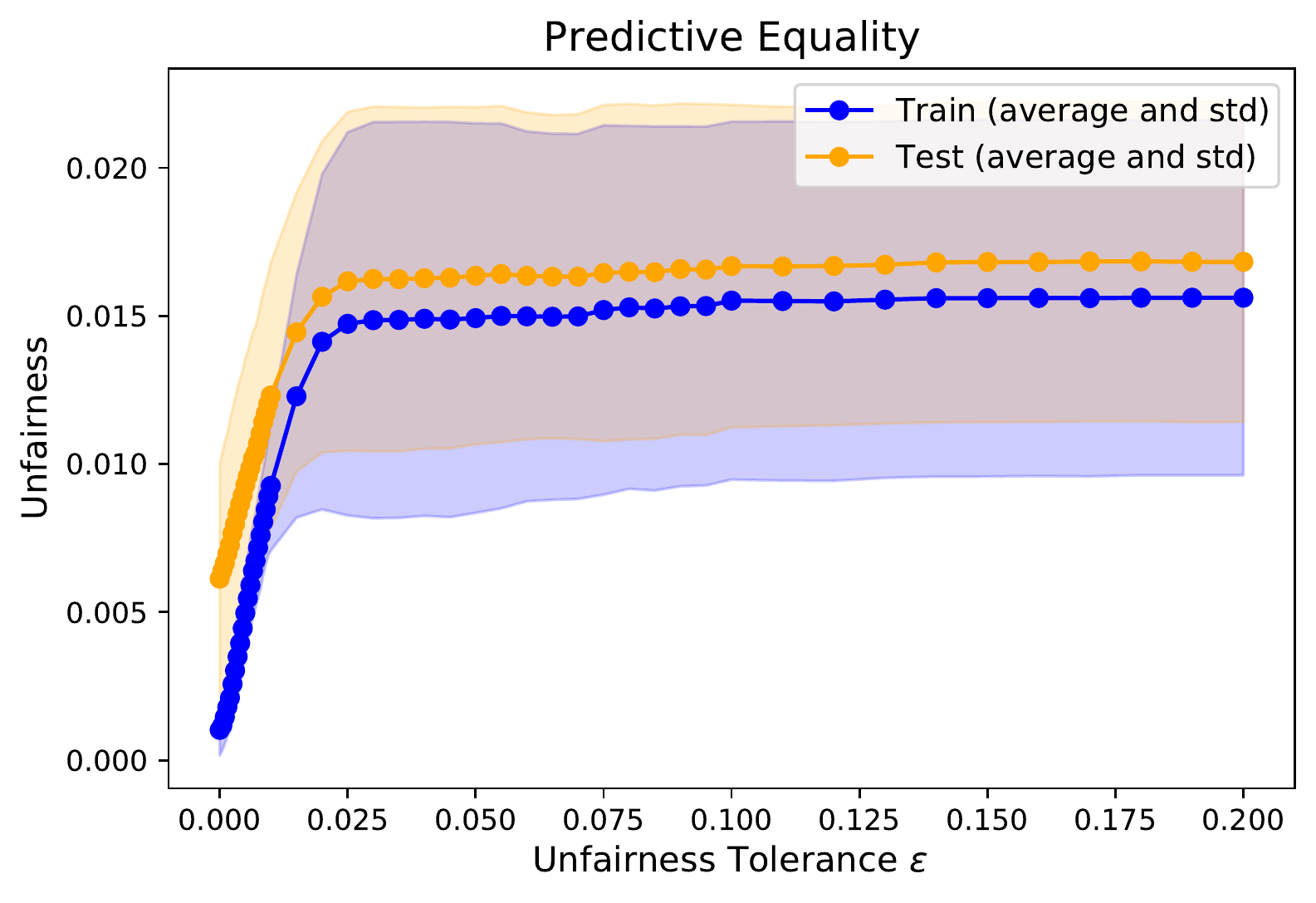}
    \vskip 0.0in
    \includegraphics[width=\figwidth\textwidth]{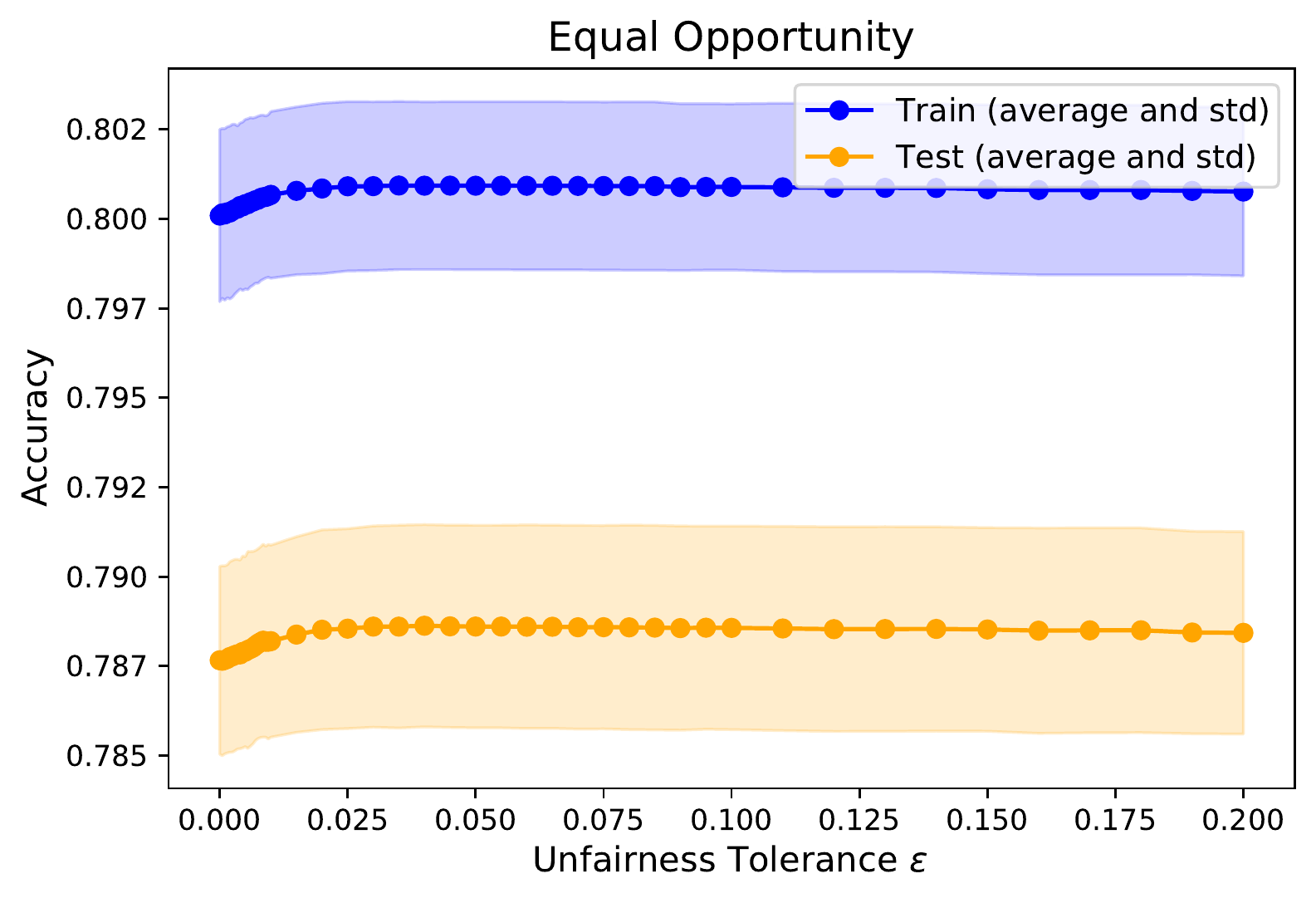}
   \includegraphics[width=\figwidth\textwidth]{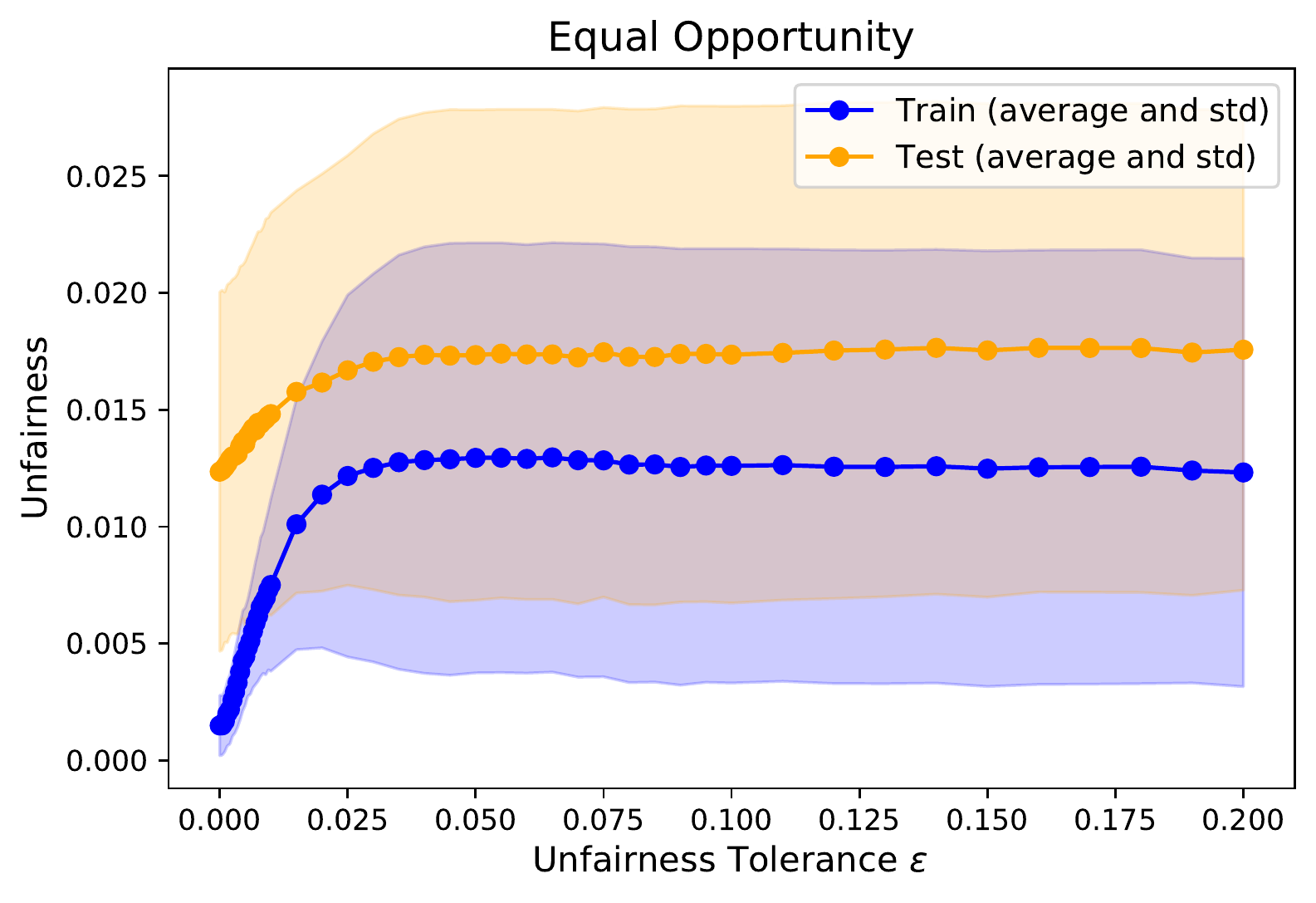}
   \vskip 0.0in
    \includegraphics[width=\figwidth\textwidth]{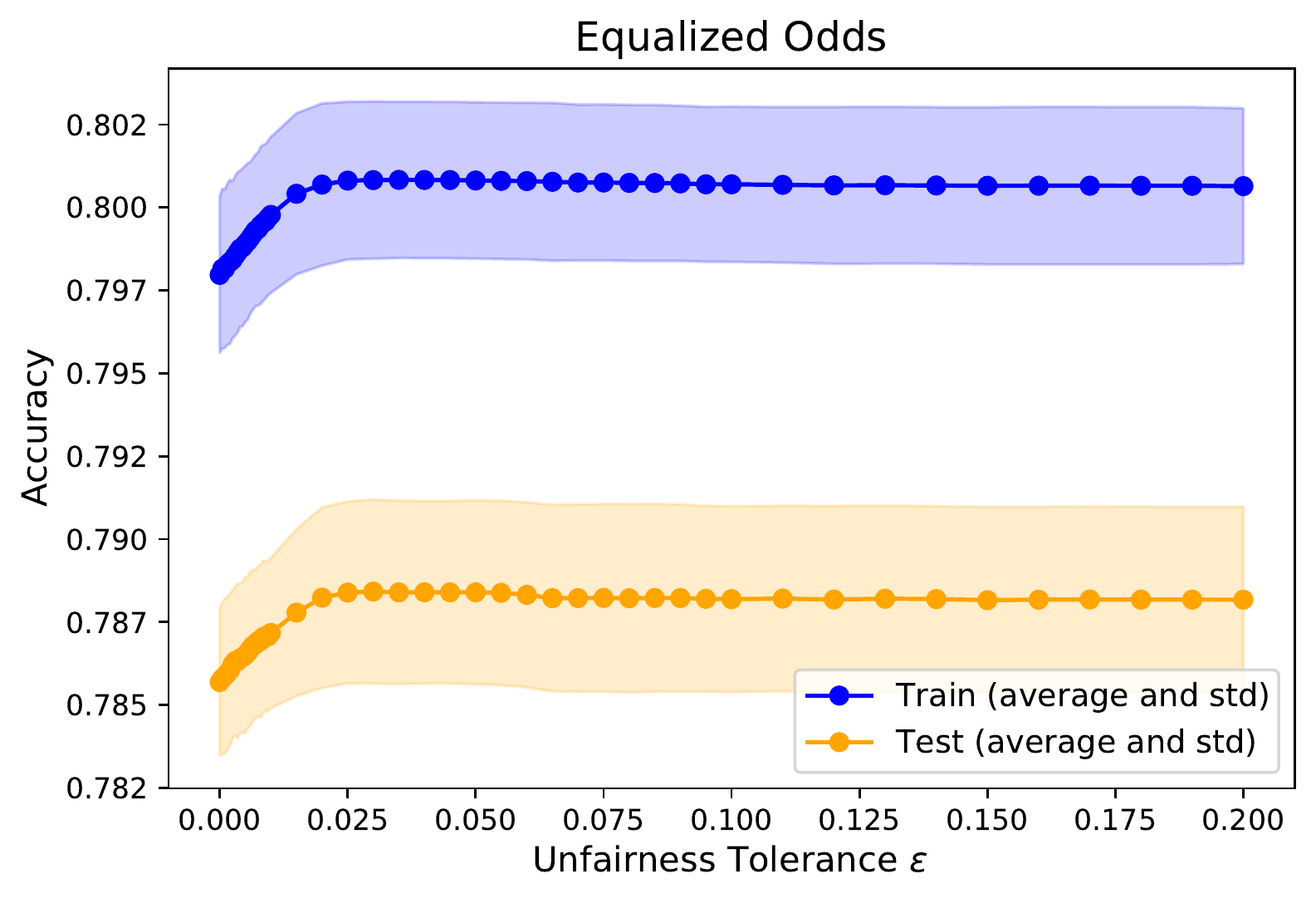}
   \includegraphics[width=\figwidth\textwidth]{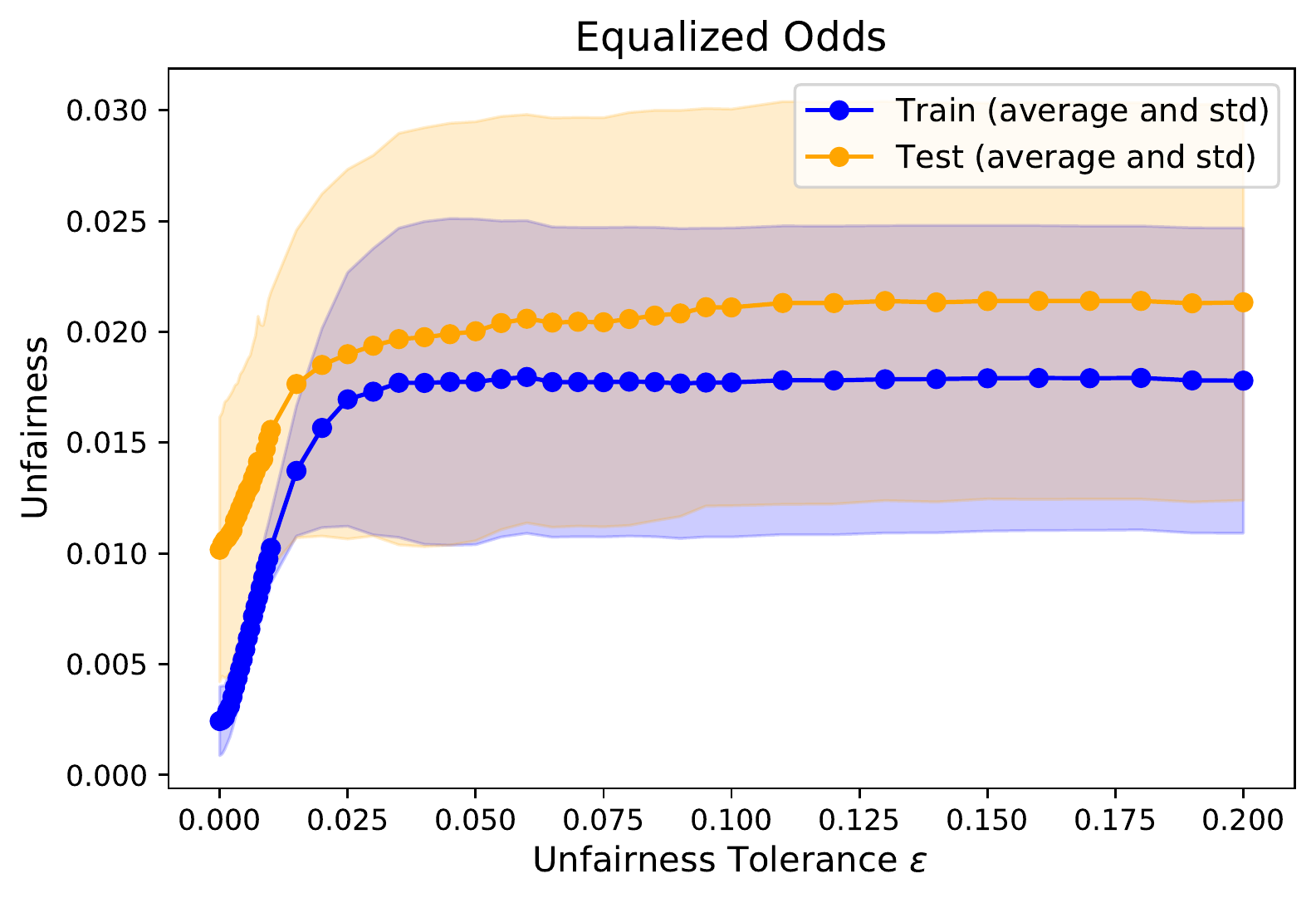}

    \end{center}
    \caption{Target models performances for our experiments using the ACSIncome dataset.}
\label{fig:target_perfs_acsincome}
\end{figure*}

In this section, we provide the training and test performances (accuracy and unfairness) of the (target) trained fair models for our experiments using the ExponentiatedGradient~\cite{DBLP:conf/icml/AgarwalBD0W18} method (Section~\ref{expes:in_proc}). 
Fig.~\ref{fig:target_perfs_adult}, \ref{fig:target_perfs_acspubliccoverage} and \ref{fig:target_perfs_acsincome} display these results for our experiments on the three datasets, for the four fairness metrics. 

As expected, we observe that both accuracy and unfairness decrease when the fairness constraint is tightened ($\tol$ diminishes). 
The trained fair models generalize rather well and so similar trends are observed on the test sets.

\section{Detailed Results: Reconstruction Accuracy using Baseline Adversary $\attackersix$}
\label{appendix:expes_attacker_6}

In this section, we provide the experimental results (reconstruction accuracy) for the experiments using the baseline adversary $\attackersix$, for a target model trained using the ExponentiatedGradient~\cite{DBLP:conf/icml/AgarwalBD0W18} method. 

\begin{figure*}[htb]
    \begin{center}
    \includegraphics[width=\figwidth\textwidth]{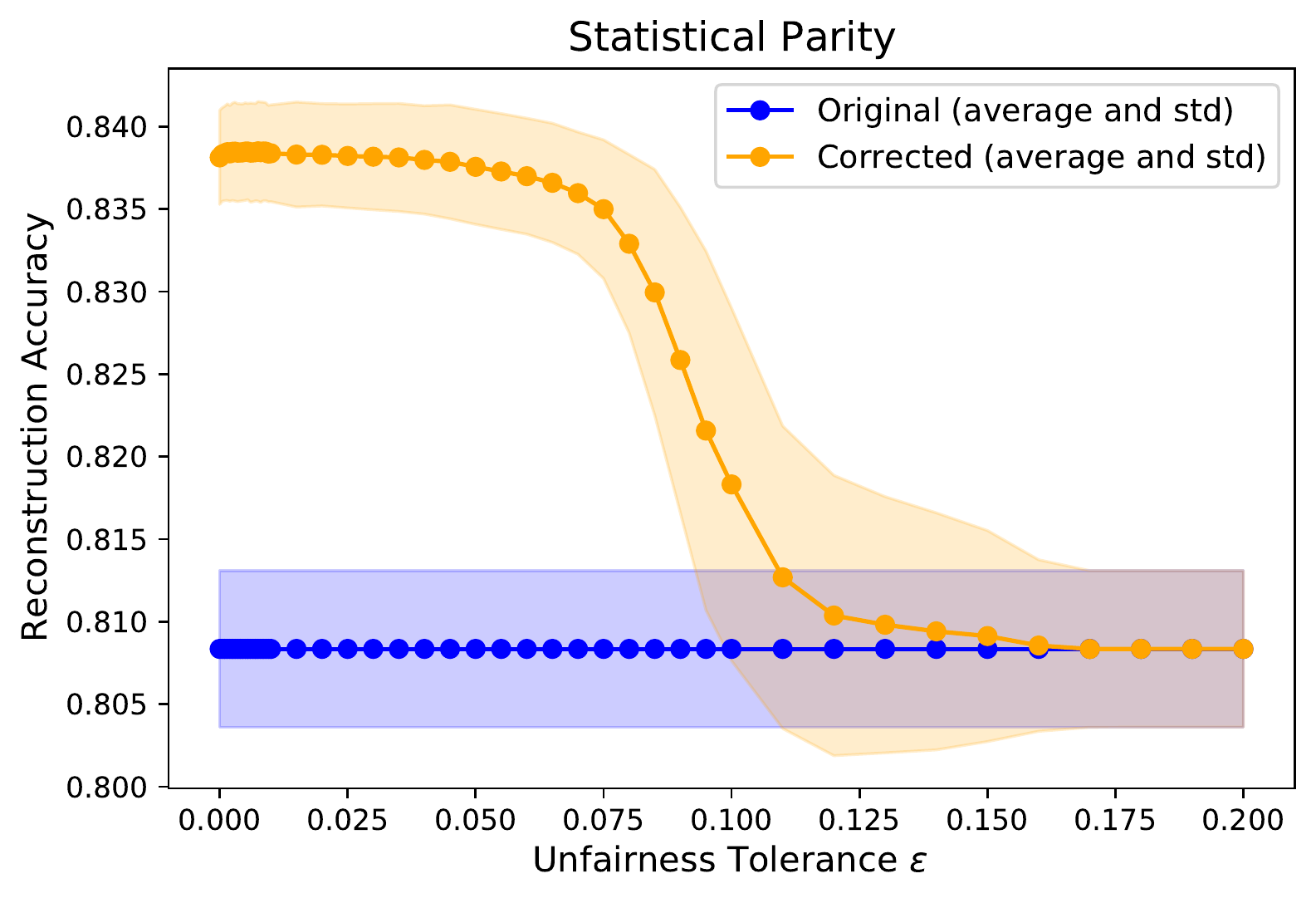} 
   \includegraphics[width=\figwidth\textwidth]{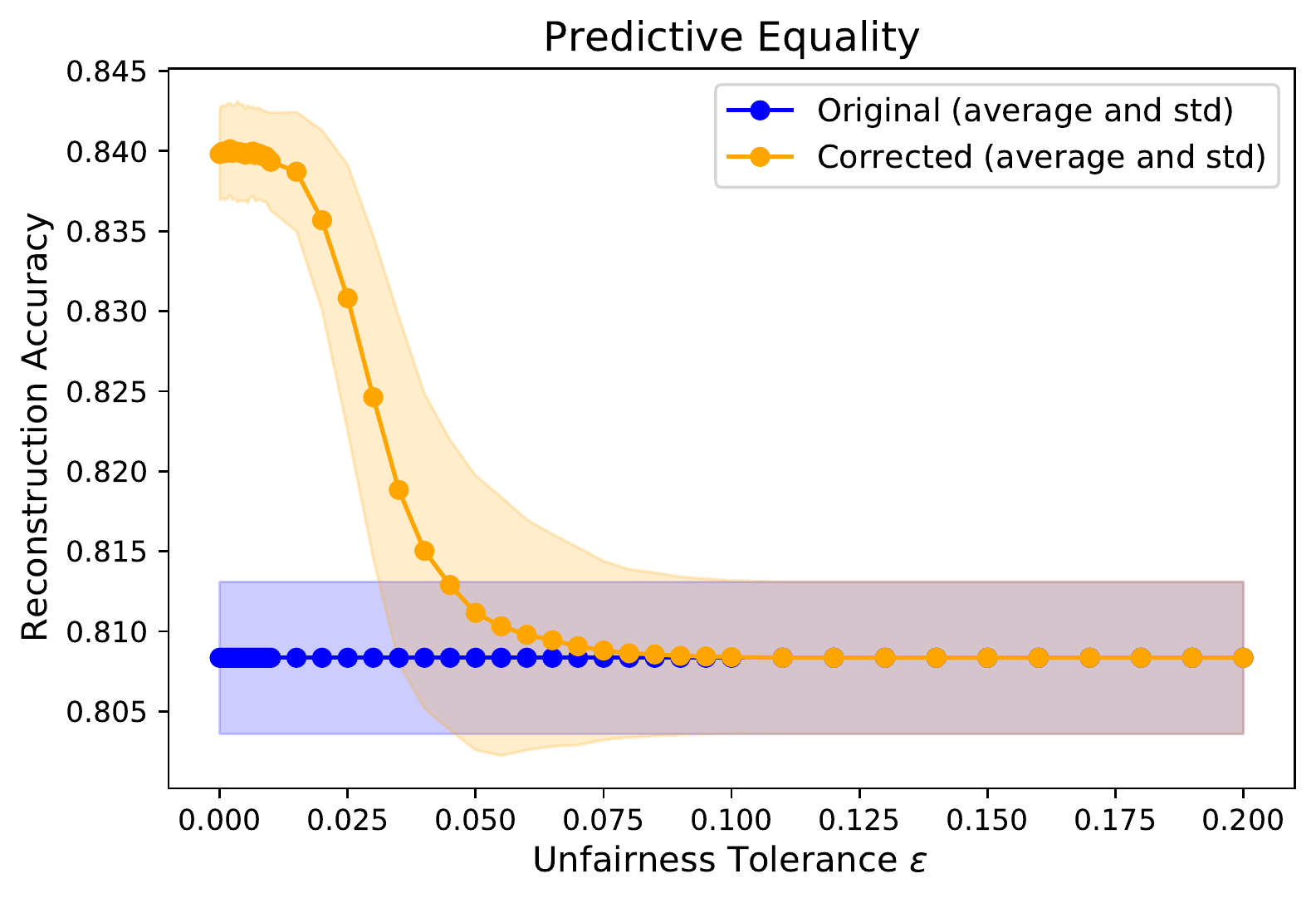}
    \includegraphics[width=\figwidth\textwidth]{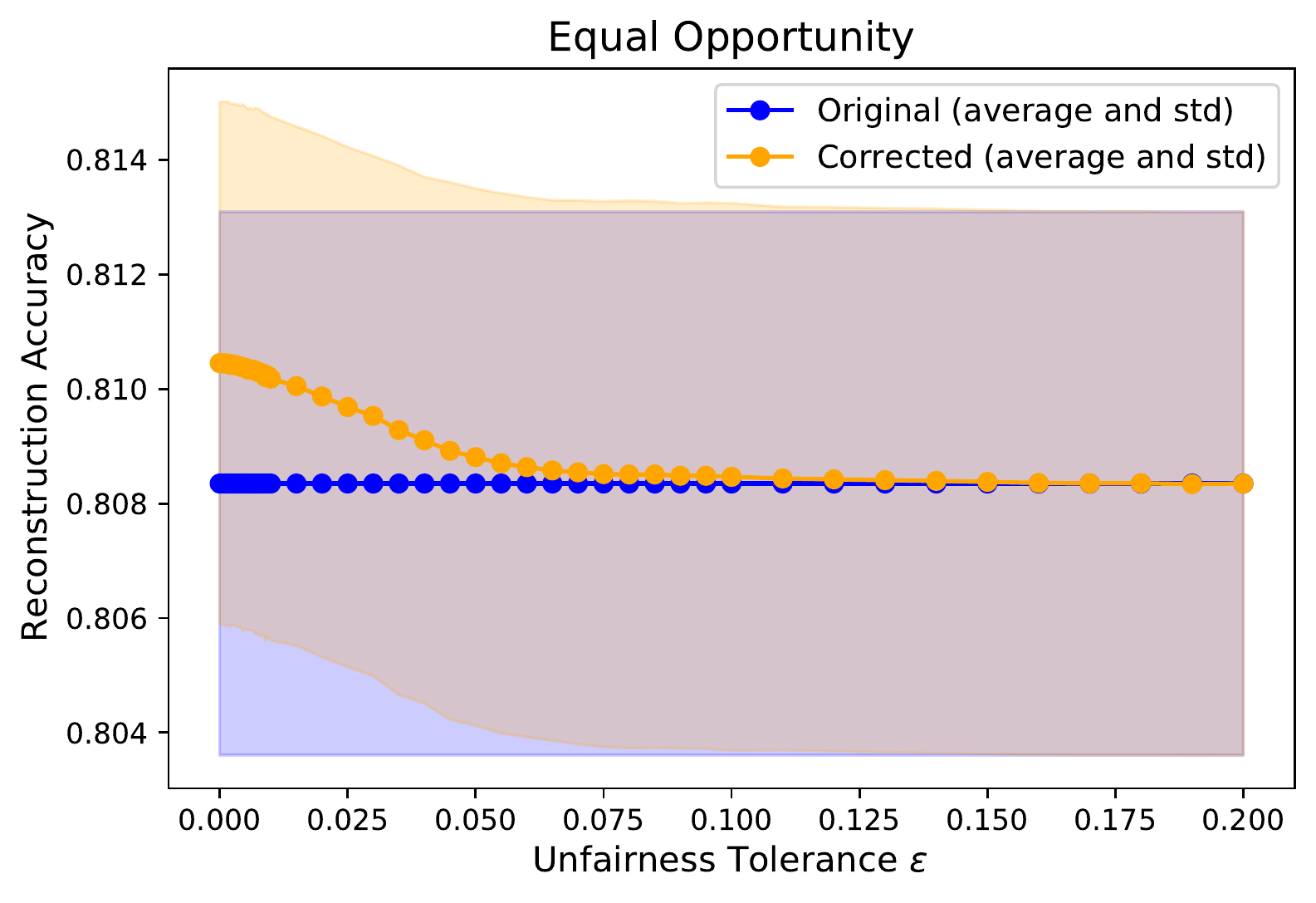}
    \includegraphics[width=\figwidth\textwidth]{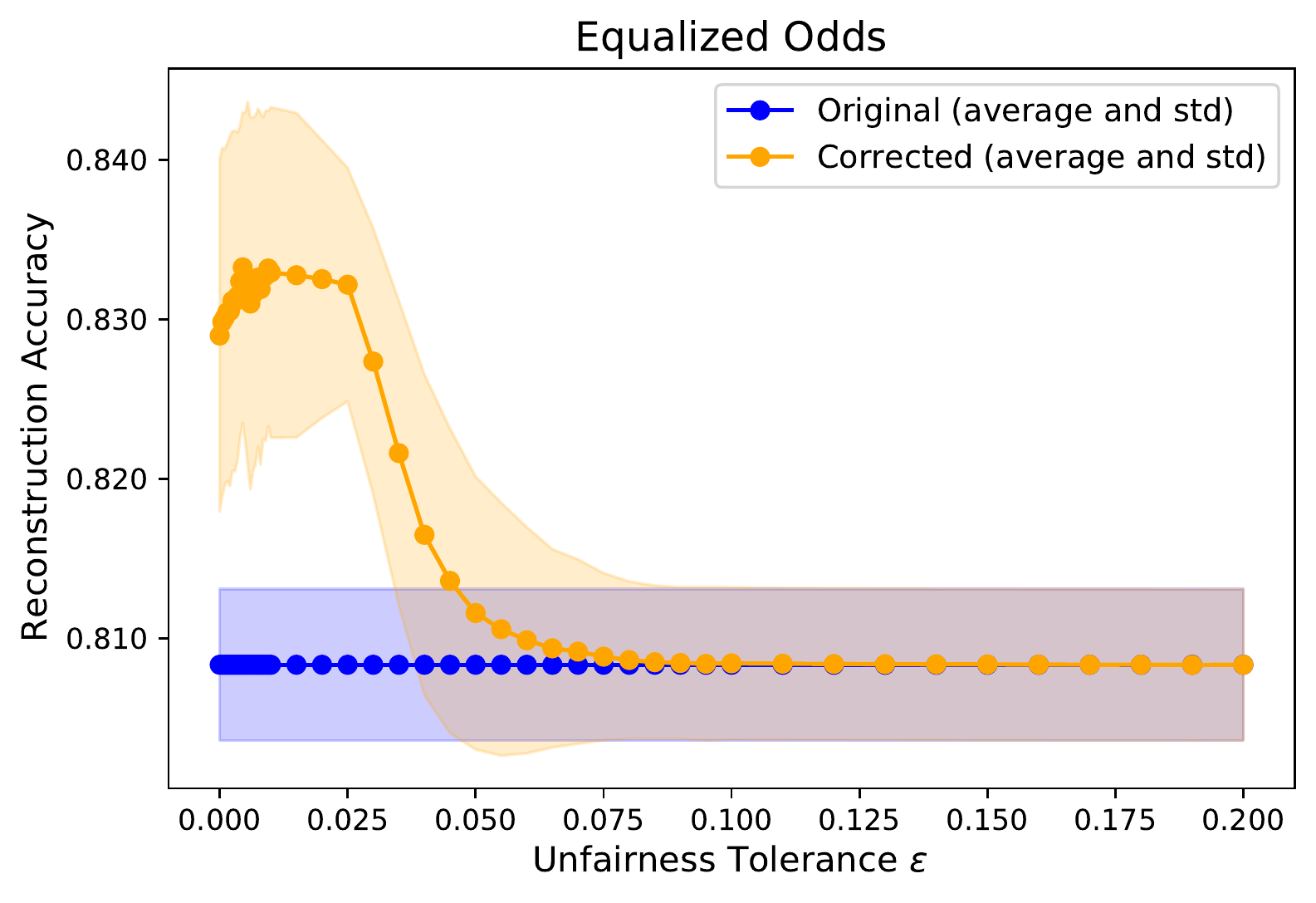}
    \end{center}
    \caption{Corrected and original (adversary $\attackersix$) reconstruction quality, for our experiments using the UCI Adult Income dataset.}
\label{fig:adult_results_inproc_attacker_6}
\end{figure*}

\begin{figure*}[htb]
    \begin{center}
    \includegraphics[width=\figwidth\textwidth]{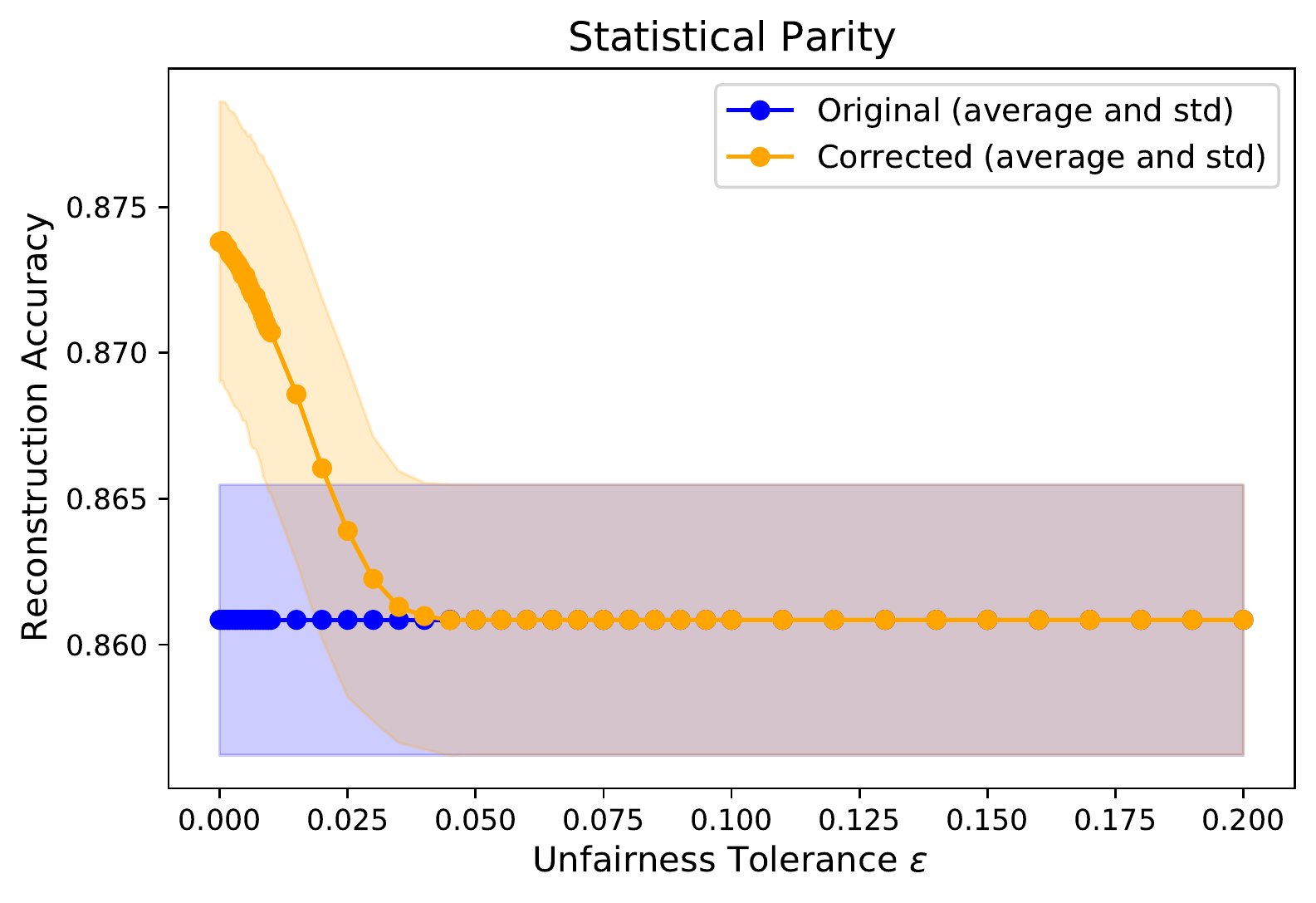} 
   \includegraphics[width=\figwidth\textwidth]{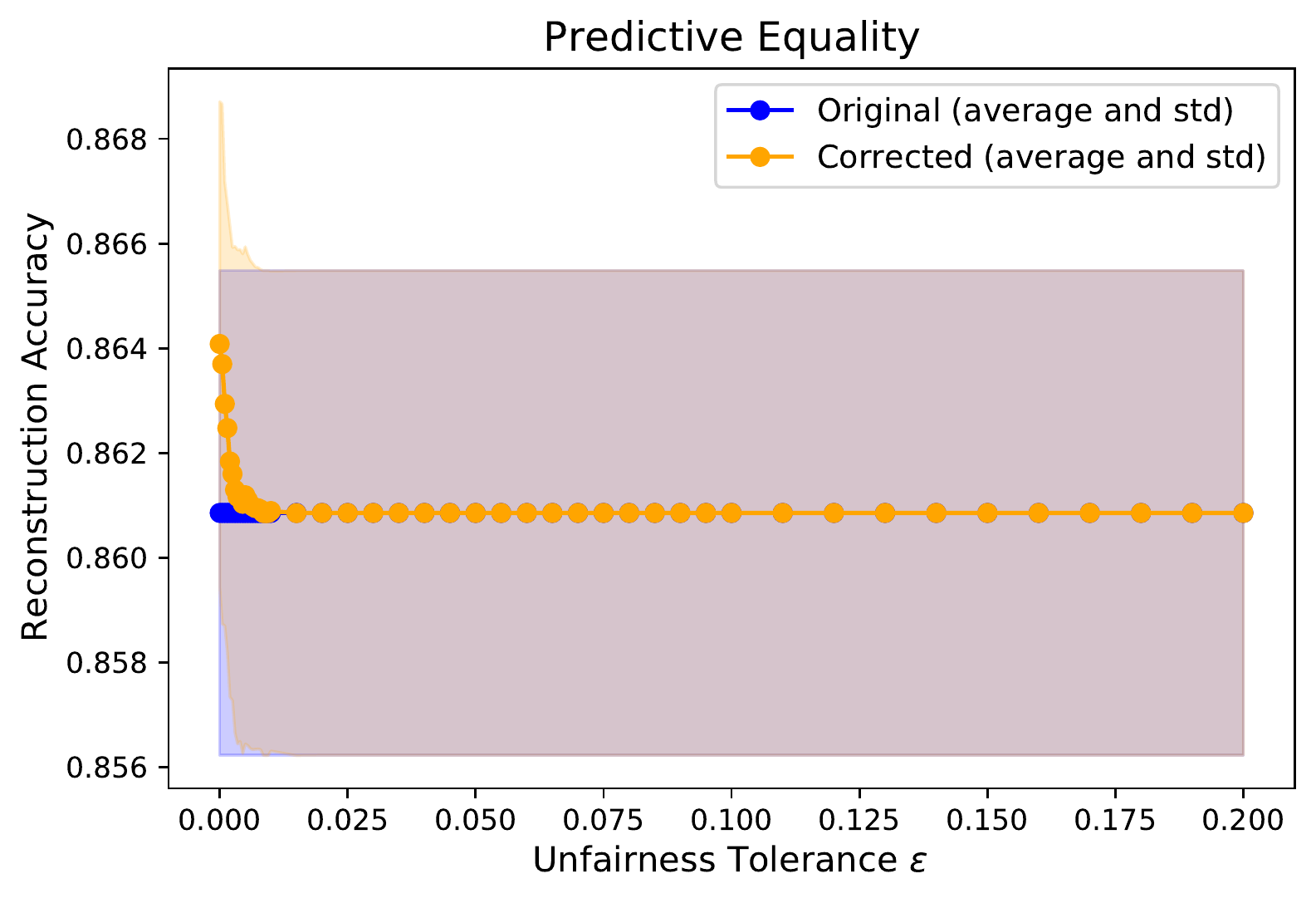}
    \includegraphics[width=\figwidth\textwidth]{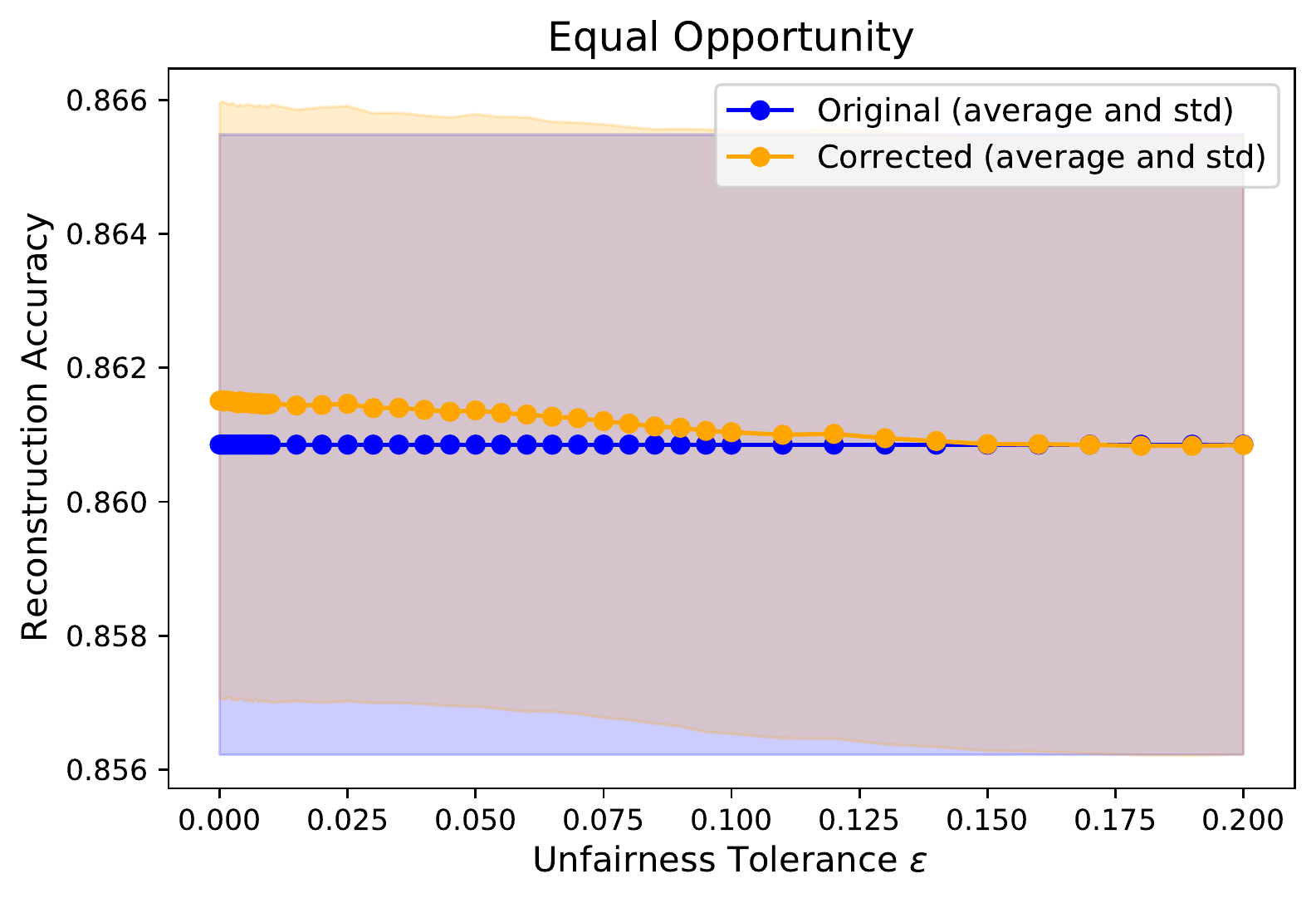}
    \includegraphics[width=\figwidth\textwidth]{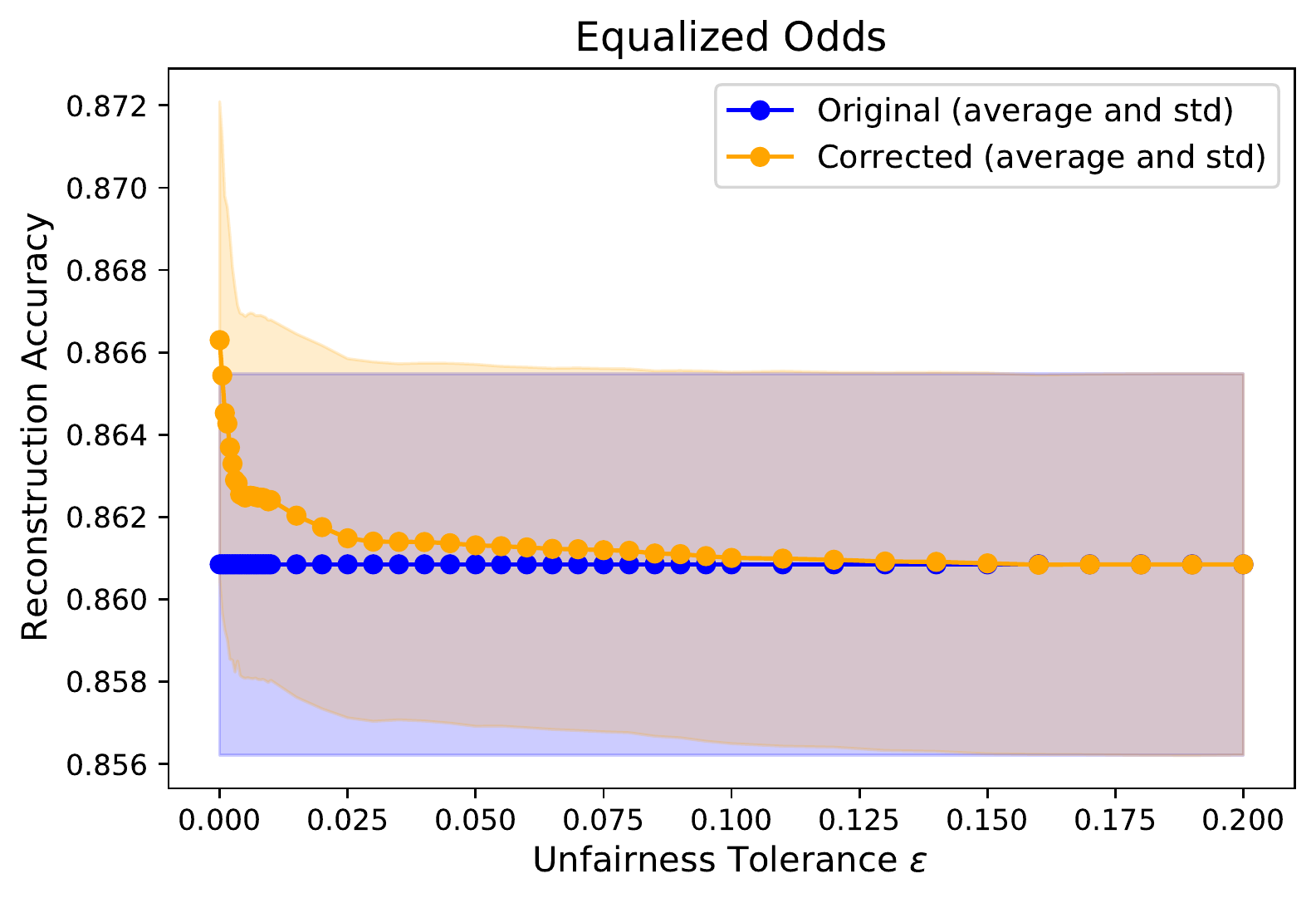}
    \end{center}
    \caption{Corrected and original (adversary $\attackersix$) reconstruction quality, for our experiments using the ACSPublicCoverage dataset}
\label{fig:acspubliccoverage_results_inproc_attacker_6}
\end{figure*}

\begin{figure*}[htb]
    \begin{center}
    \includegraphics[width=\figwidth\textwidth]{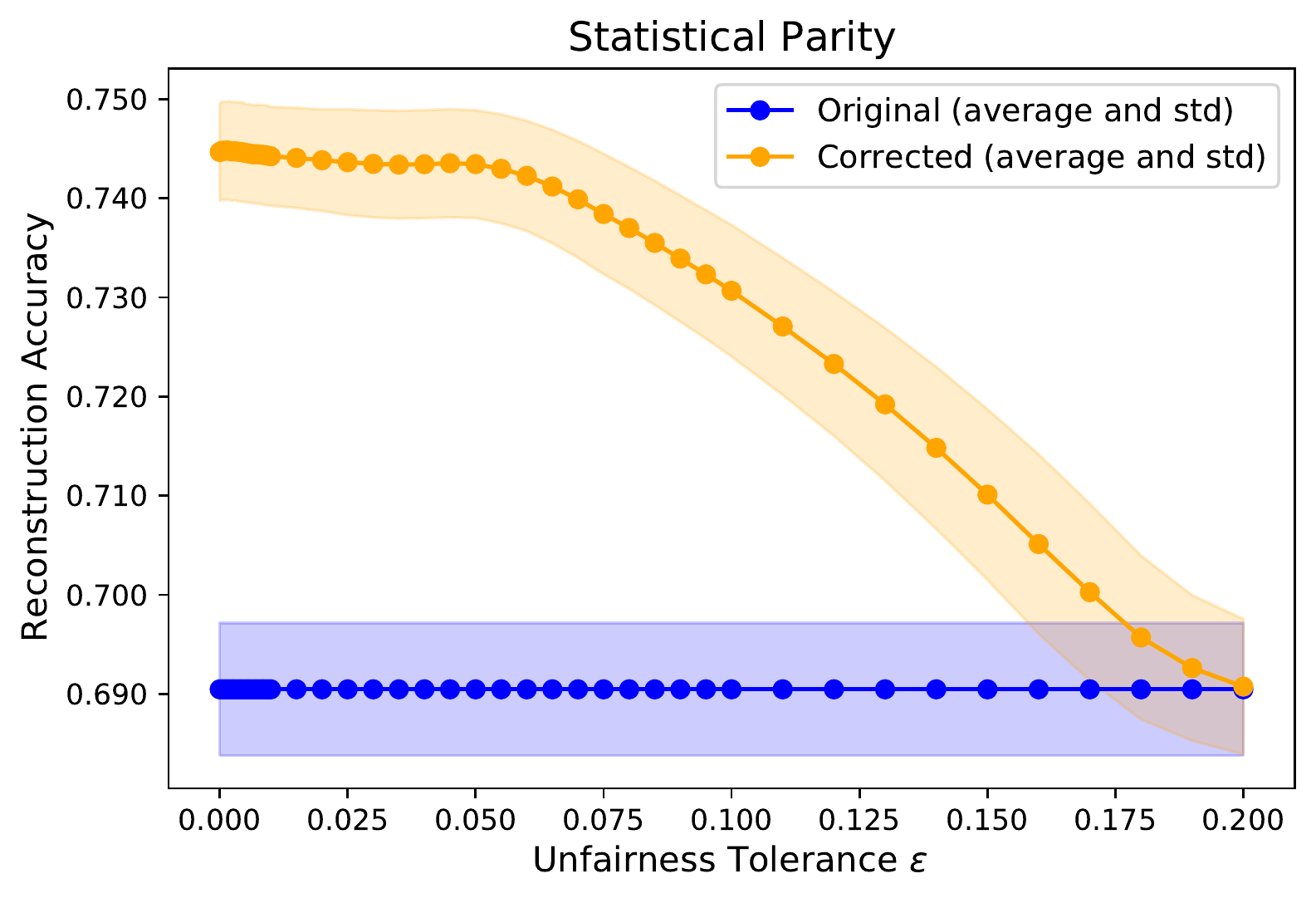} 
   \includegraphics[width=\figwidth\textwidth]{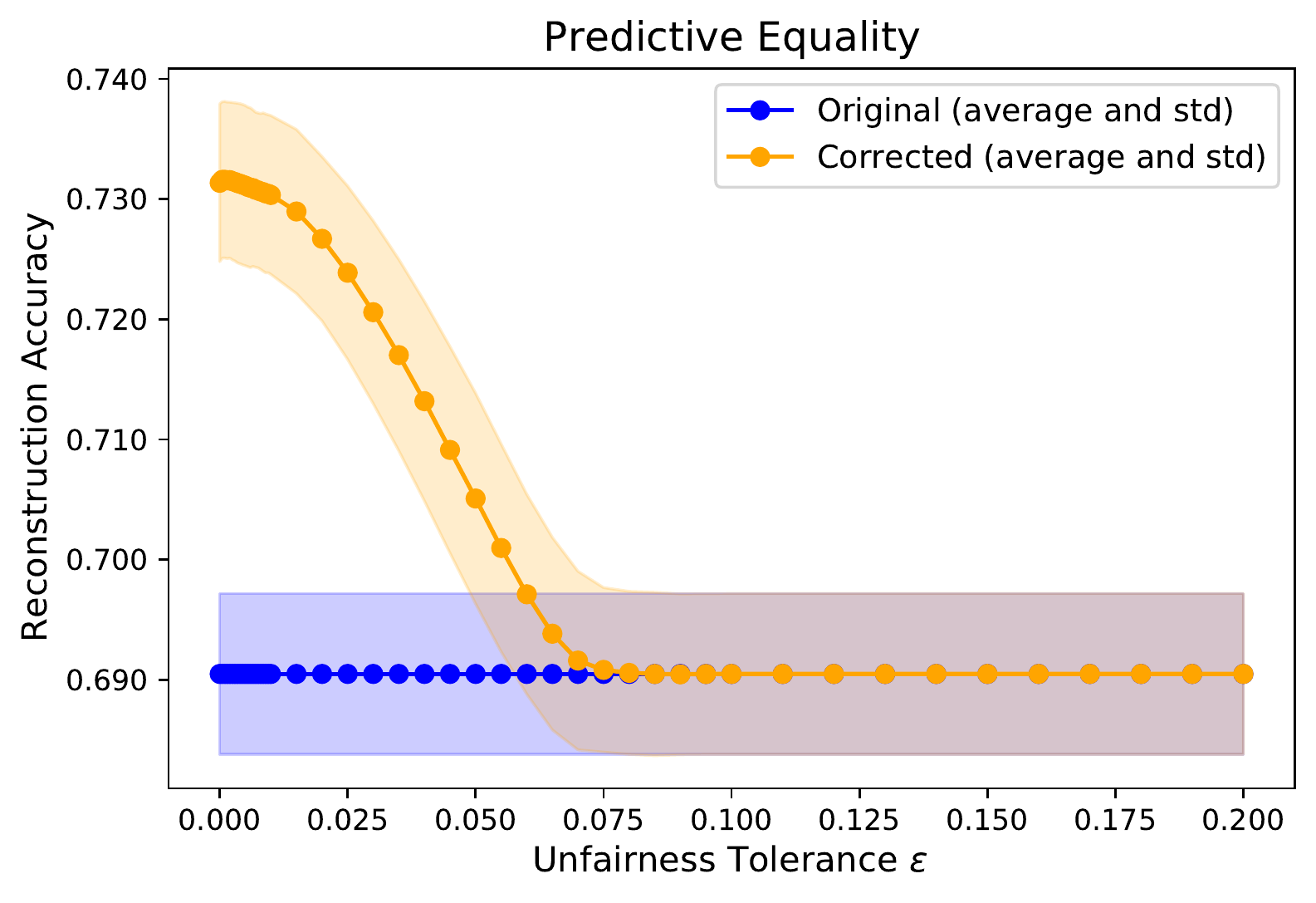}
    \includegraphics[width=\figwidth\textwidth]{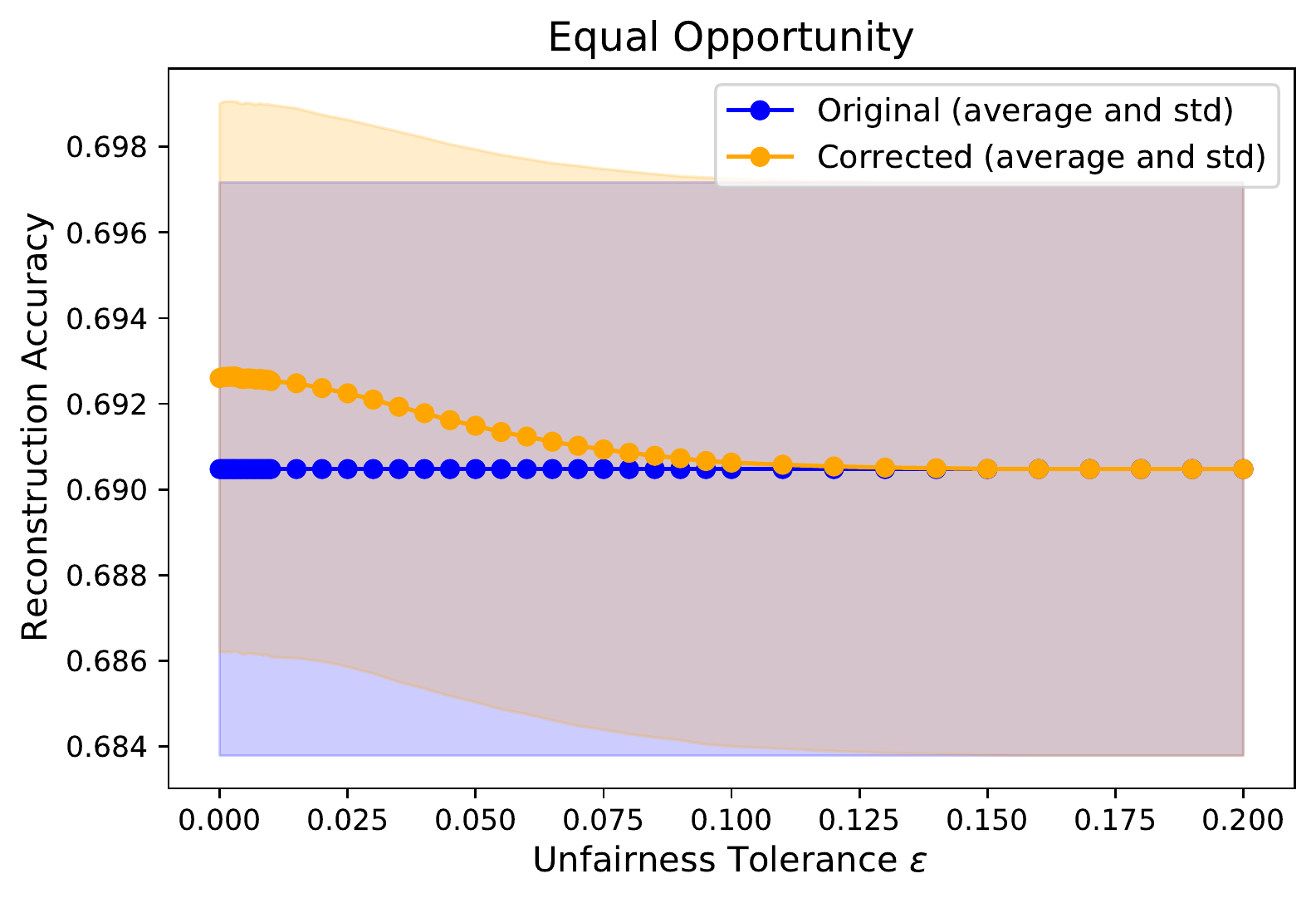}
    \includegraphics[width=\figwidth\textwidth]{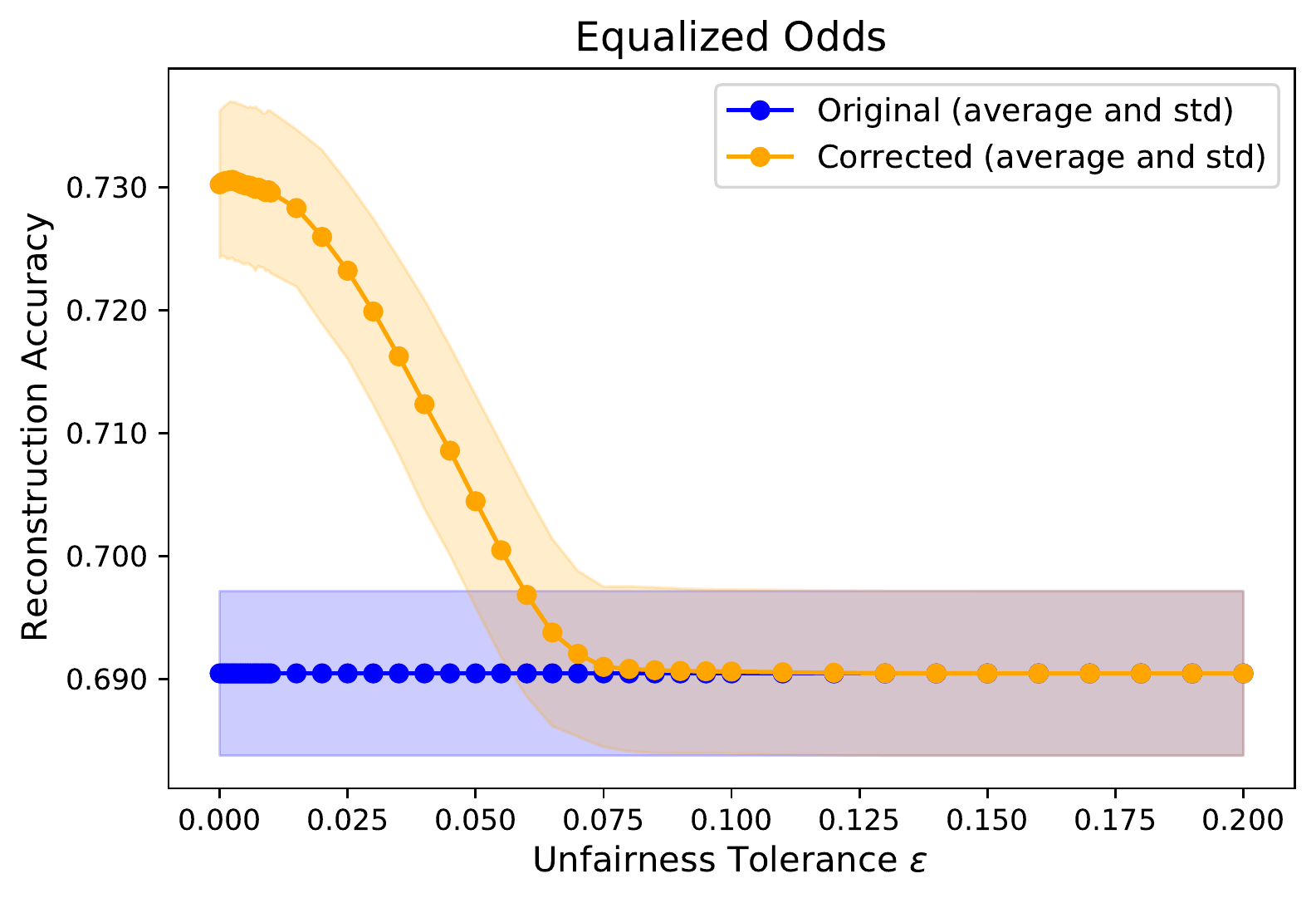}
    \end{center}
    \caption{Corrected and original (adversary $\attackersix$) reconstruction quality, for our experiments using the ACSIncome dataset}
\label{fig:acsincome_results_inproc_attacker_6}
\end{figure*}

These results are displayed in Fig.~\ref{fig:adult_results_inproc_attacker_6},~\ref{fig:acspubliccoverage_results_inproc_attacker_6} and~\ref{fig:acsincome_results_inproc_attacker_6}. 
They show the same trends as the experiments using the baseline adversary $\attackerseven$ (Fig.~\ref{fig:adult_results_inproc},~\ref{fig:acspubliccoverage_results_inproc} and~\ref{fig:acsincome_results_inproc}, presented in Section~\ref{expes:in_proc}). 
In particular, we observe that the corrected reconstruction always has better accuracy than the original one made by the baseline adversary. 
More precisely, the tighter the fairness constraint (\emph{i.e.}, the smaller the unfairness tolerance $\epsilon$), the greater the reconstruction correction step improvement.

\section{Detailed Results: Reconstruction Correction Performances from Estimated Fairness Constraints}
\label{appendix:countermeasures_other_expes}
In this section, we provide the reconstruction correction performances for our experiments using the ExponentiatedGradient~\cite{DBLP:conf/icml/AgarwalBD0W18} method, including a scenario in which the actual fairness constraint is not known. In such case, the adversary has to estimate it as described in Section~\ref{subsec:hiding_fairness}. Fig.~\ref{fig:adult_results_inproc_protection},~\ref{fig:acspubliccoverage_results_inproc_protection} and~\ref{fig:acsincome_results_inproc_protection} display these results for our experiments on the three datasets, for the four fairness metrics, with baseline adversary $\attackerseven$. Fig.~\ref{fig:adult_results_inproc_protection_attacker6},~\ref{fig:acspubliccoverage_results_inproc_protection_attacker6}, and~\ref{fig:acsincome_results_inproc_protection_attacker6} display these results for baseline adversary $\attackersix$. 
Results for both adversaries are very similar and show the same phenomenons.

\begin{table*}[htb!]
\caption{Summary of the results of our experiments using a pre-processing method for fairness, with the attacker inferring the fairness information. We report the accuracy performances of the trained (target) model, the results of the fairness constraint estimation process (inferred metrics and average inferred tolerance), and the reconstruction performances.}\label{tab:preproc}
\centering
\begin{tabular}{cccccccc}
\hline
\multicolumn{2}{c|}{\textbf{Target model (under attack)}}                                                                                                                                & \multicolumn{2}{c|}{\textbf{Estimated Constraint}}                                                                                                                                                        & \multicolumn{2}{c|}{\textbf{\begin{tabular}[c]{@{}c@{}}Baseline\\ Reconstructions\end{tabular}}}                & \multicolumn{2}{c}{\textbf{\begin{tabular}[c]{@{}c@{}}Corrected\\ Reconstructions\end{tabular}}} \\ \hline
\multicolumn{1}{c|}{\textit{\textbf{\begin{tabular}[c]{@{}c@{}}Train\\ Acc.\end{tabular}}}} & \multicolumn{1}{c|}{\textit{\textbf{\begin{tabular}[c]{@{}c@{}}Test\\ Acc.\end{tabular}}}} & \multicolumn{1}{c|}{\textit{\textbf{\begin{tabular}[c]{@{}c@{}}Estimated\\ Metric\end{tabular}}}} & \multicolumn{1}{c|}{\textit{\textbf{\begin{tabular}[c]{@{}c@{}}Estimated \\ Tolerance\end{tabular}}}} & \multicolumn{1}{c|}{\textit{\textbf{$\attackersix$}}} & \multicolumn{1}{c|}{\textit{\textbf{$\attackerseven$}}} & \multicolumn{1}{c|}{\textit{\textbf{$\attackersix$}}}    & \textit{\textbf{$\attackerseven$}}    \\ \hline
\multicolumn{8}{c}{\textbf{UCI Adult Income dataset}}                                                                                                                                                                                                                                                                                                                                                                                                                                                                                                                                                                     \\ \hline
$0.860 \pm 0.003$                                                                           & $0.848 \pm 0.003$                                                                          & PE (68\%), EO (32\%)                                                                              & $0.023 \pm 0.013$                                                                                     & $0.808 \pm 0.005$                                     & $0.806 \pm 0.005$                                       & $0.828 \pm 0.013$                                        & $\mathbf{0.827 \pm 0.014}$            \\ \hline
\multicolumn{8}{c}{\textbf{ACSPublicCoverage dataset}}                                                                                                                                                                                                                                                                                                                                                                                                                                                                                                                                                                    \\ \hline
$0.862 \pm 0.001$                                                                           & $0.852 \pm 0.002$                                                                          & PE (92\%), SP (8\%)                                                                               & $0.006 \pm 0.004$                                                                                     & $0.861 \pm 0.005$                                     & $0.860 \pm 0.006$                                       & $0.863 \pm 0.005$                                        & $\mathbf{0.872 \pm 0.010}$            \\ \hline
\multicolumn{8}{c}{\textbf{ACSIncome dataset}}                                                                                                                                                                                                                                                                                                                                                                                                                                                                                                                                                                            \\ \hline
$0.798 \pm 0.002$                                                                           & $0.785 \pm 0.003$                                                                          & PE (100\%)                                                                                        & $0.056 \pm 0.016$                                                                                     & $0.690 \pm 0.007$                                     & $0.685 \pm 0.008$                                       & $0.704 \pm 0.014$                                        & $\mathbf{0.763 \pm 0.009}$            \\ \hline
\end{tabular}
\end{table*}

As discussed in Section~\ref{subsec:hiding_fairness}, we observe several trends. 
When the fairness constraint is tight enough, the estimation process usually estimates it correctly. 
In this situation, the reconstruction correction step then exhibits slightly weaker performances as the provided tolerance estimation is not as tight as the actual constraint. 
However, when the unfairness tolerance $\tol$ is large enough, its actual value is not informative, while the estimated one is usually tighter leading to more accurate reconstruction results. 
Finally, the fairness constraint estimation process sometimes comes up with a fairness metric differing from the actual optimized one. 
When the proposed metric is more informative (in terms of number of examples involved in its computation), the reconstruction performance can even be better than with the actual constraint.

\def\figwidth{0.41}
 \begin{figure*}[htb]
    \begin{center}
    \includegraphics[width=\figwidth\textwidth]{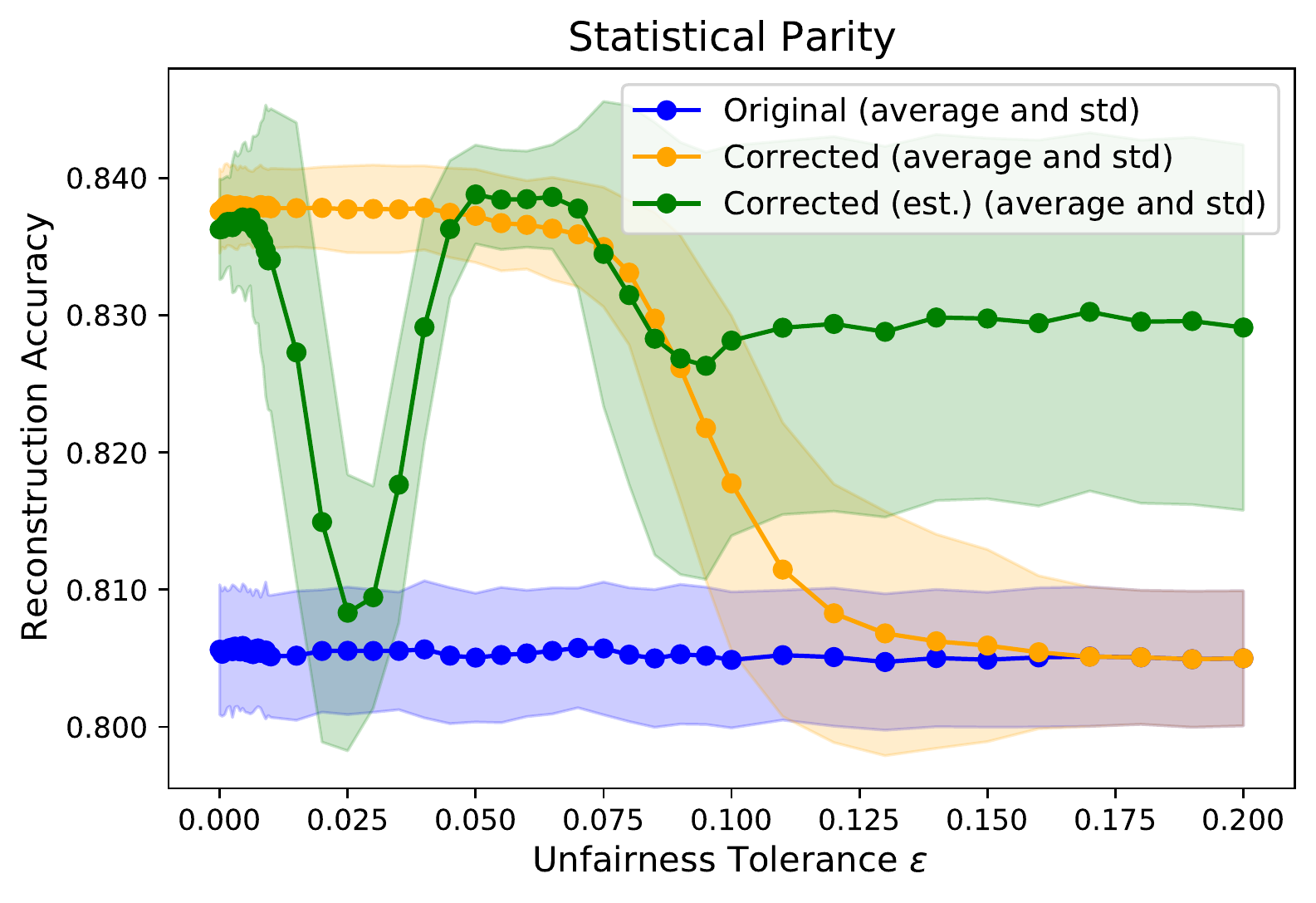} 
   \includegraphics[width=\figwidth\textwidth]{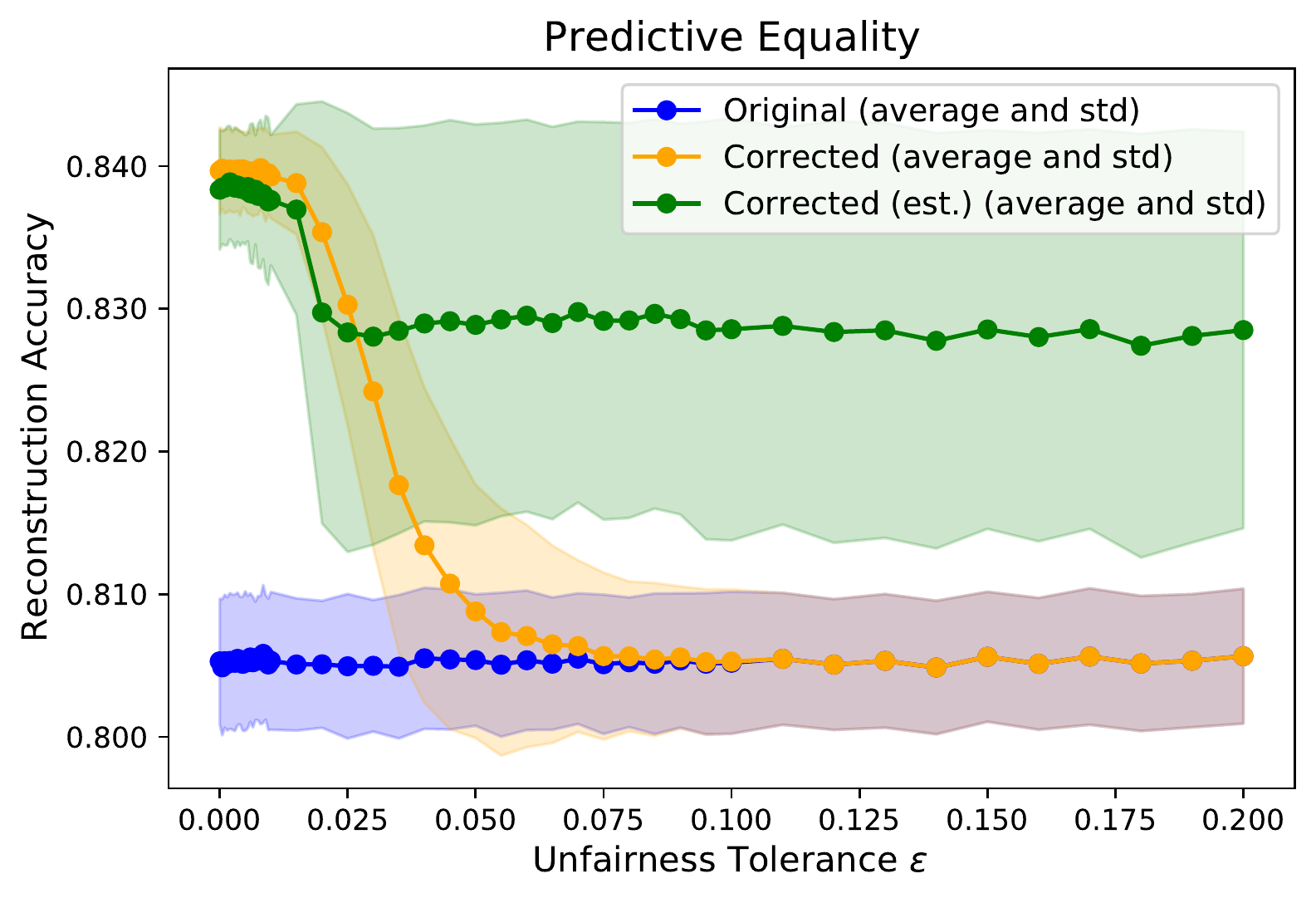}
    \includegraphics[width=\figwidth\textwidth]{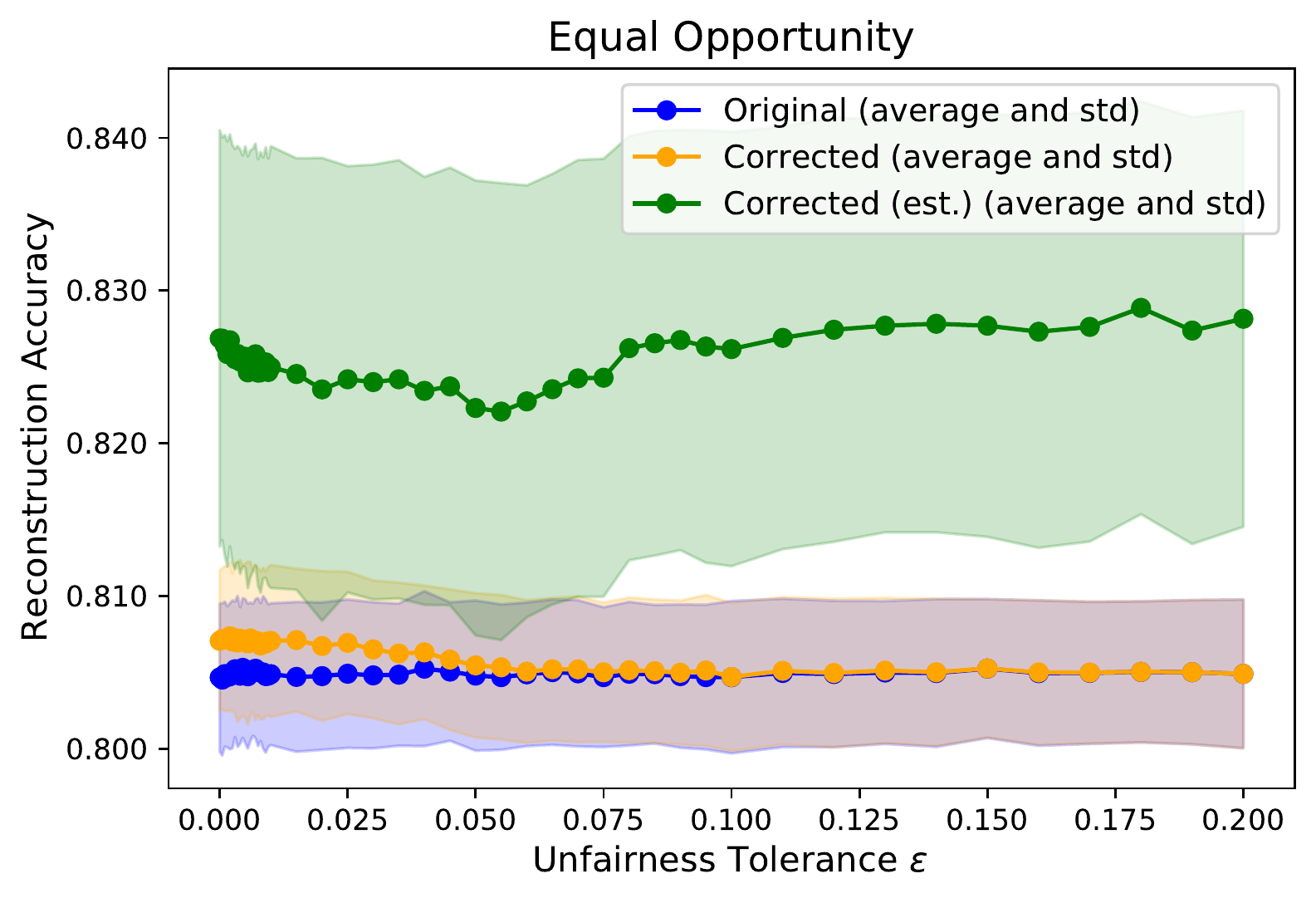}
    \includegraphics[width=\figwidth\textwidth]{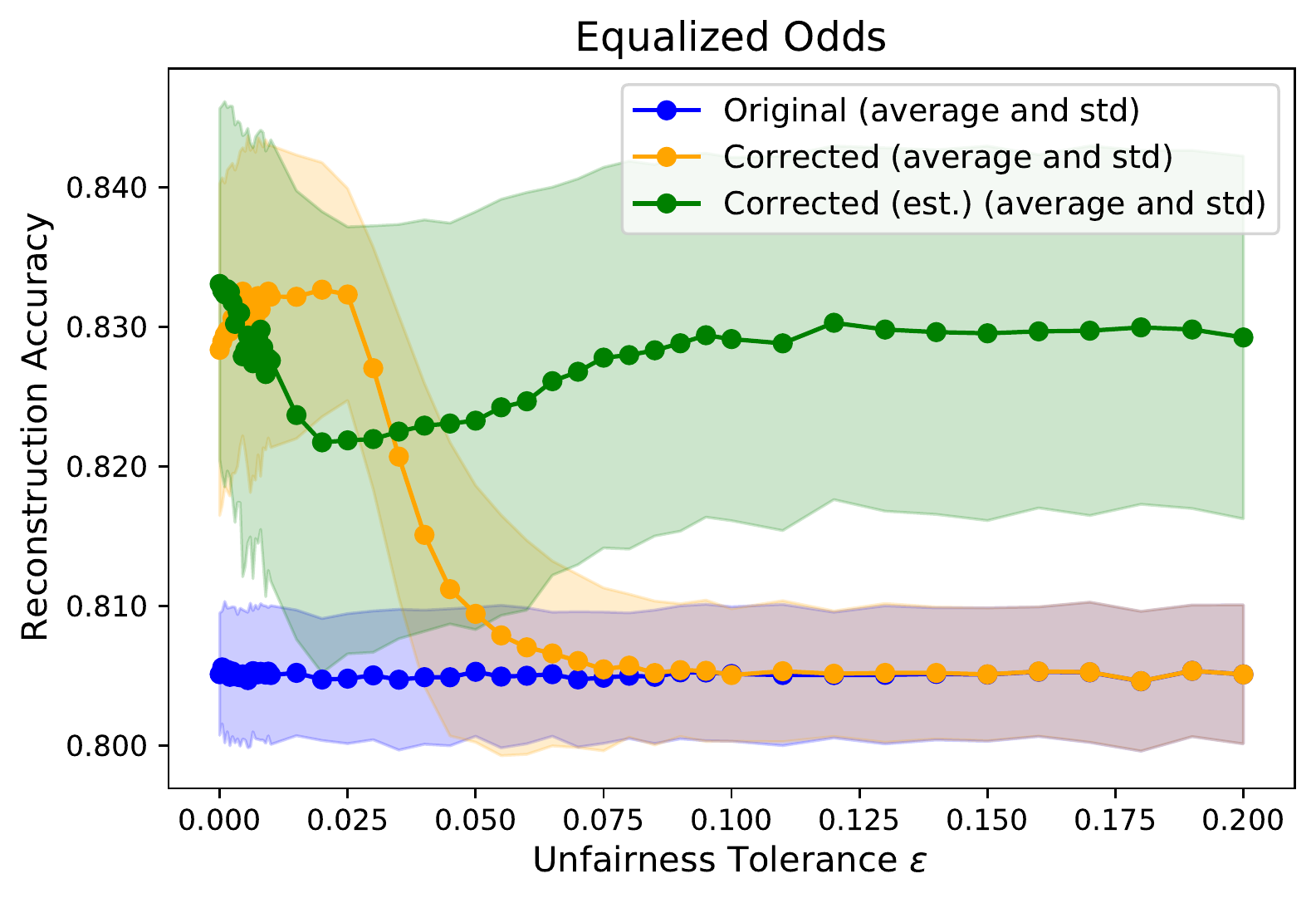}
    \end{center}
    \caption{Original (attacker $\attackerseven$), corrected (from actual fairness constraint, and from estimated one (\texttt{est.})) reconstruction quality, for our experiments using the UCI Adult Income dataset}
\label{fig:adult_results_inproc_protection}
\end{figure*}

 \begin{figure*}[htb]
    \begin{center}
    \includegraphics[width=\figwidth\textwidth]{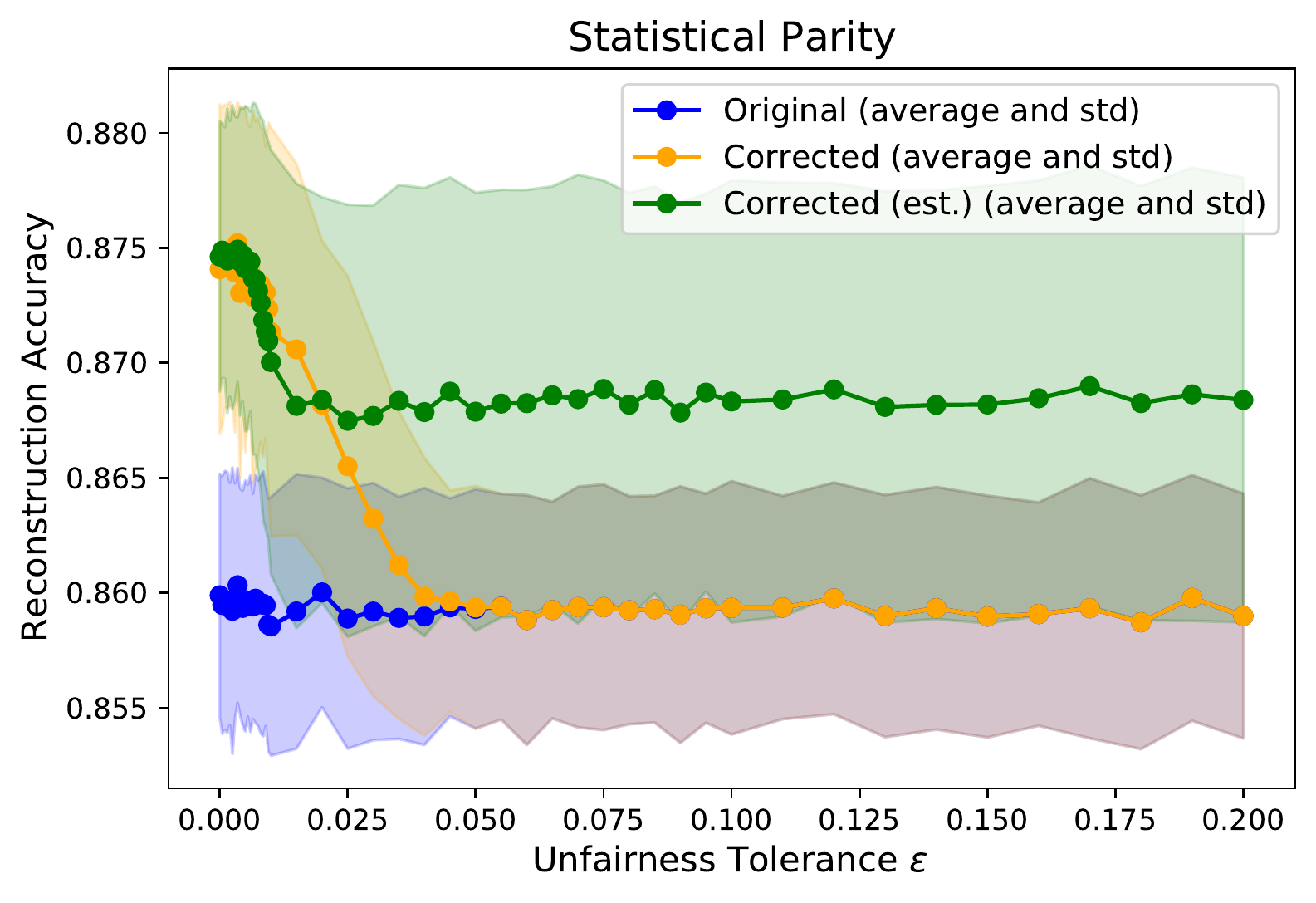} 
   \includegraphics[width=\figwidth\textwidth]{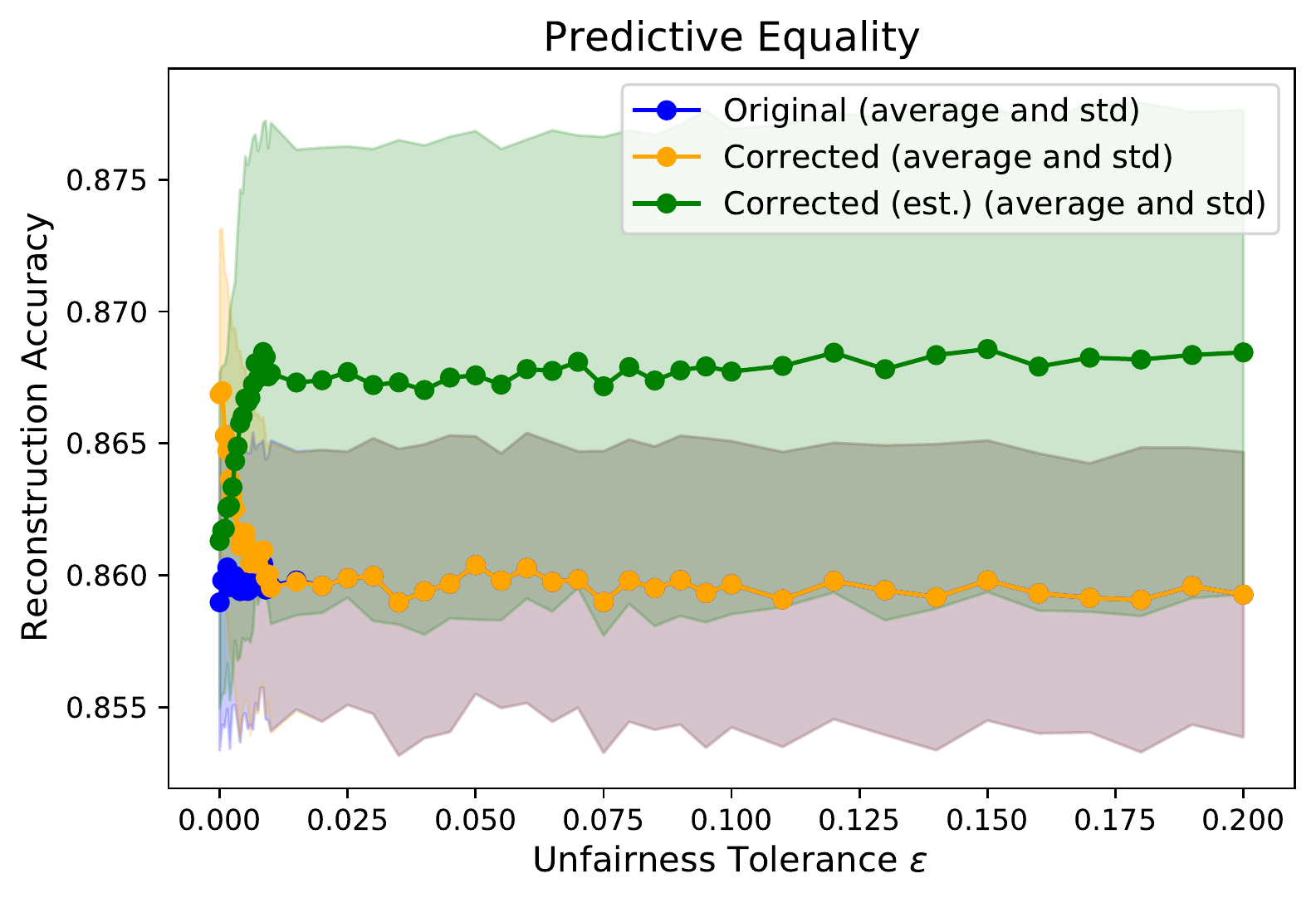}
    \includegraphics[width=\figwidth\textwidth]{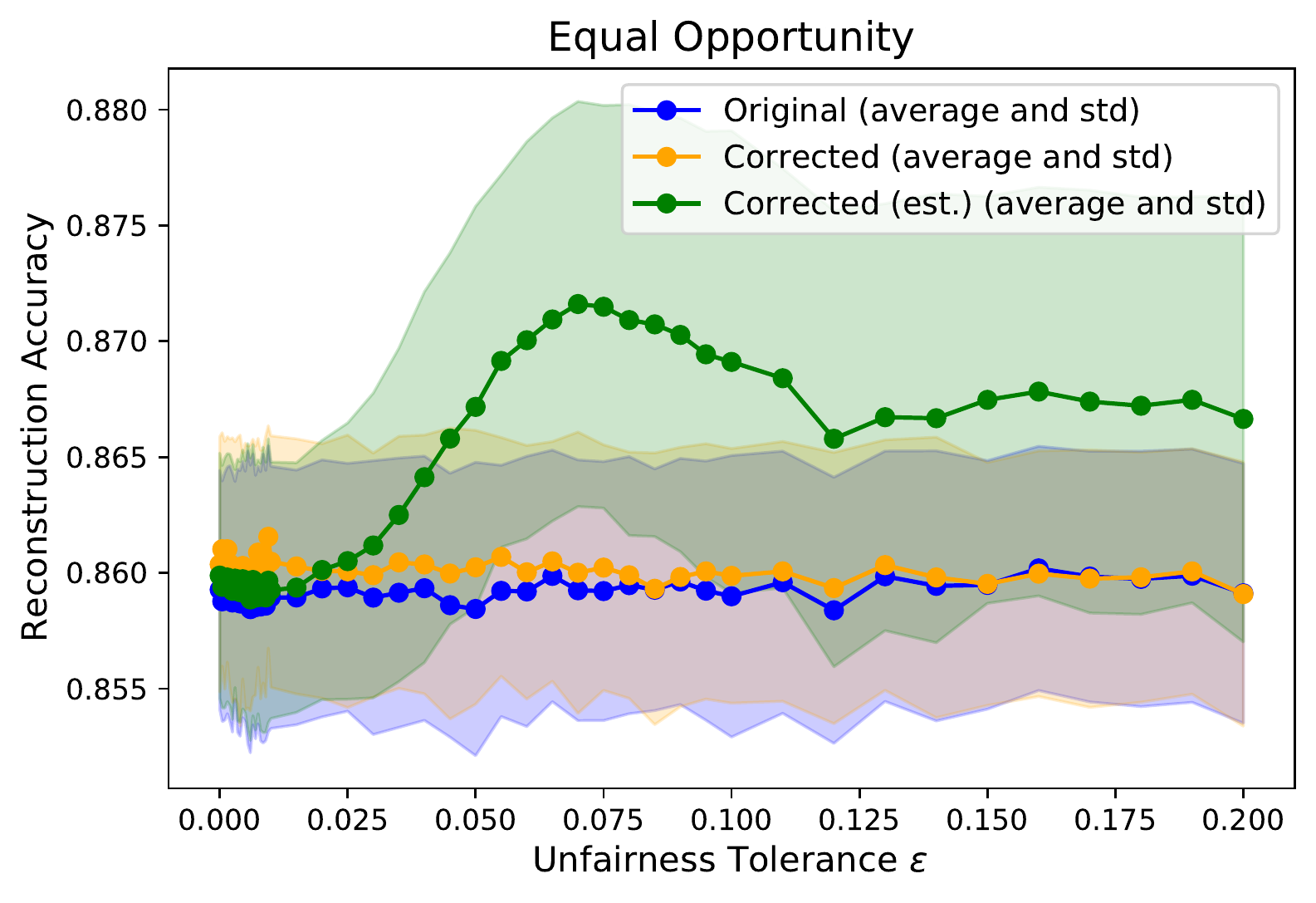}
    \includegraphics[width=\figwidth\textwidth]{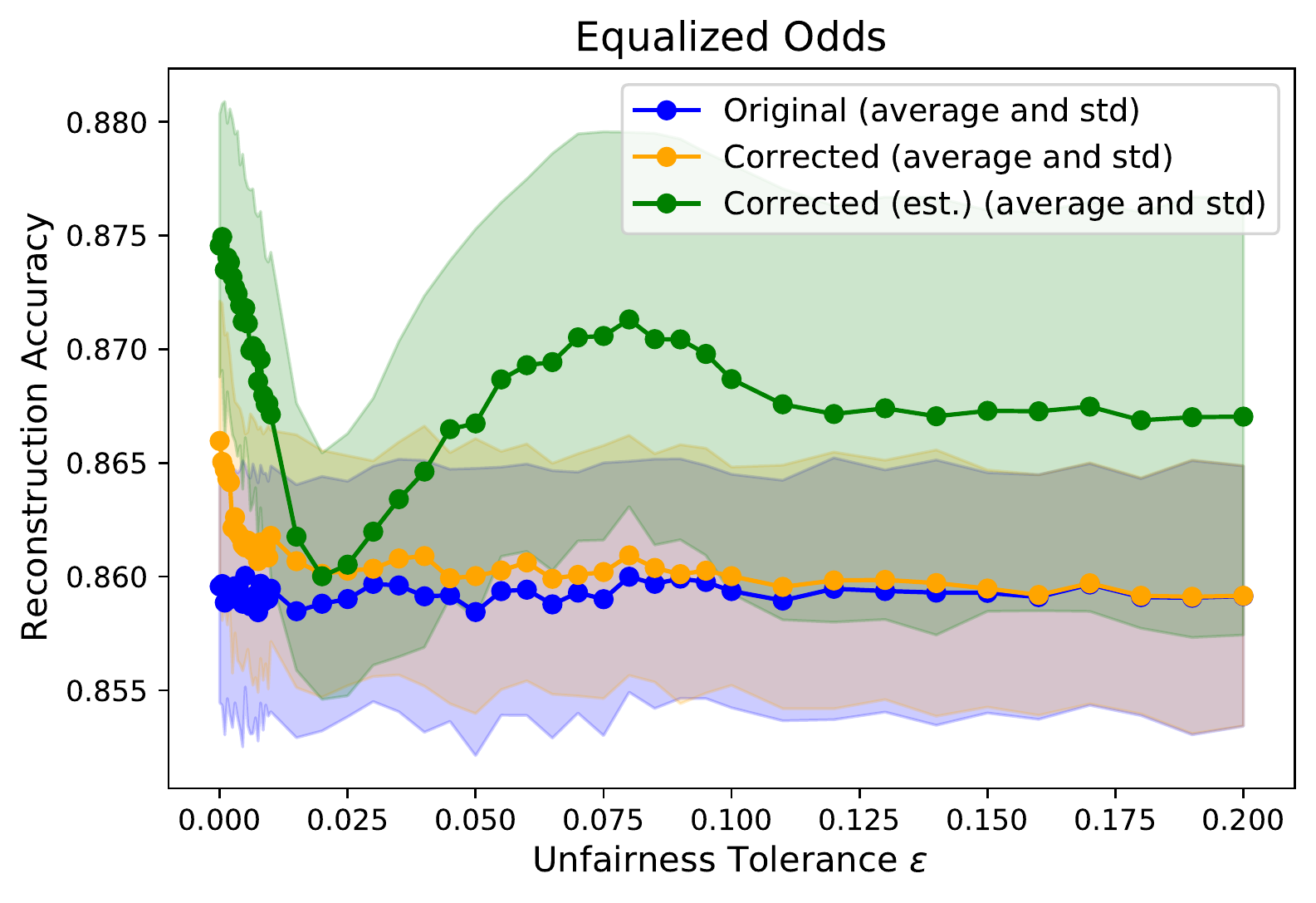}
    \end{center}
    \caption{Original (attacker $\attackerseven$), corrected (from actual fairness constraint, and from estimated one (\texttt{est.})) reconstruction quality, for our experiments using the ACSPublicCoverage dataset}
\label{fig:acspubliccoverage_results_inproc_protection}
\end{figure*}

 \begin{figure*}[htb]
    \begin{center}
    \includegraphics[width=\figwidth\textwidth]{protection_sensAttrReco_ACSIncome_race_1_attacker_7_reconstruction.pdf} 
   \includegraphics[width=\figwidth\textwidth]{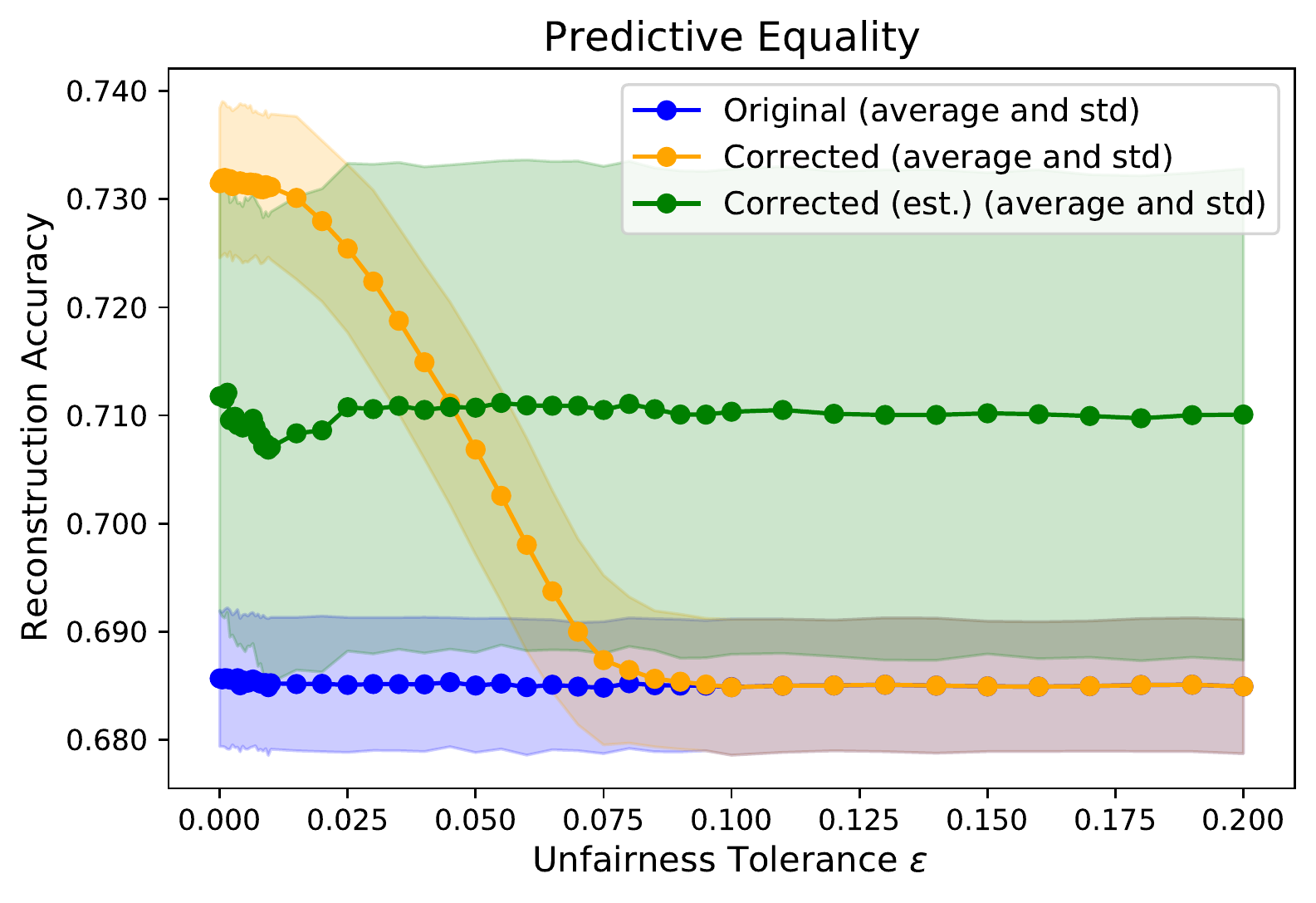}
    \includegraphics[width=\figwidth\textwidth]{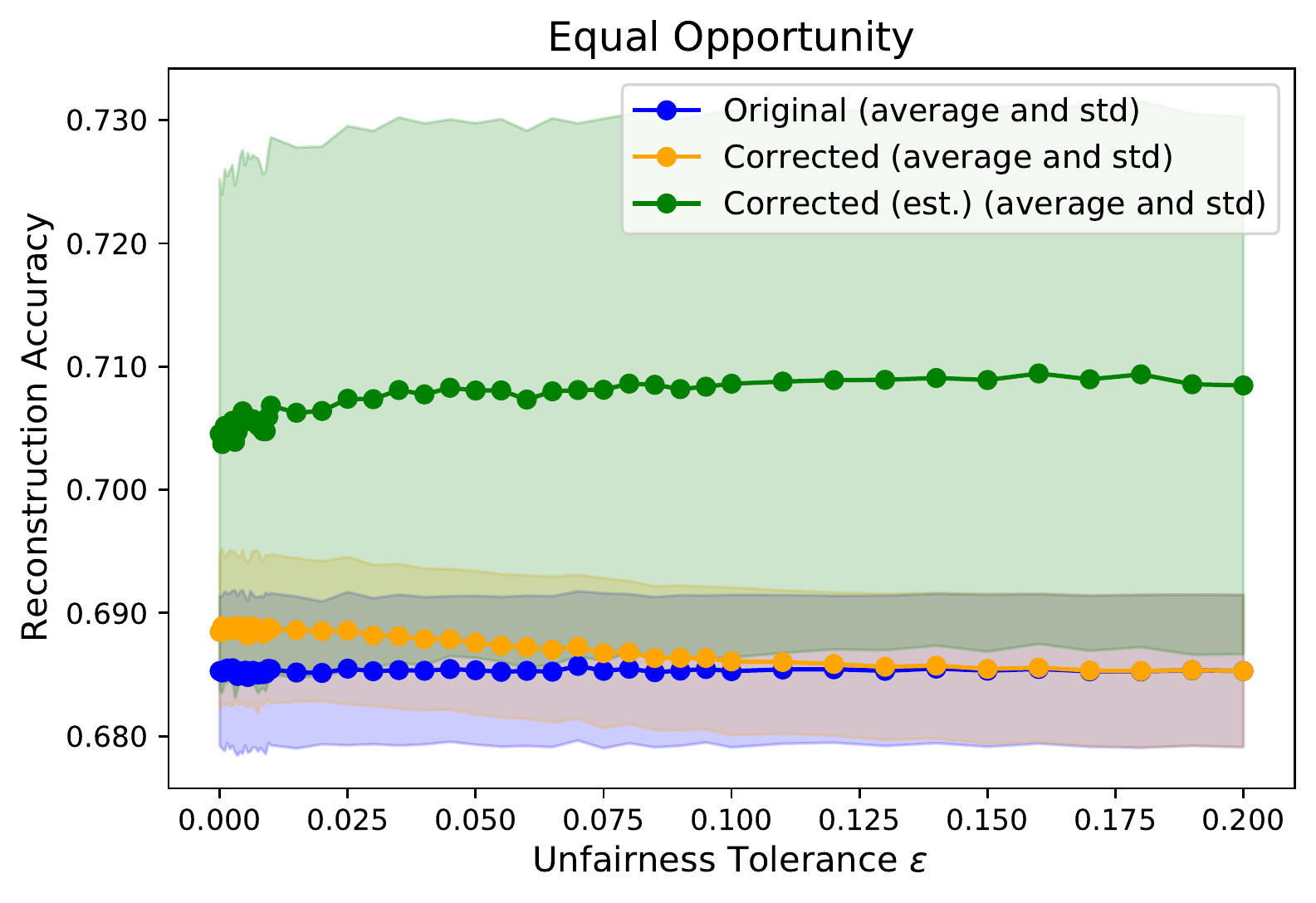}
    \includegraphics[width=\figwidth\textwidth]{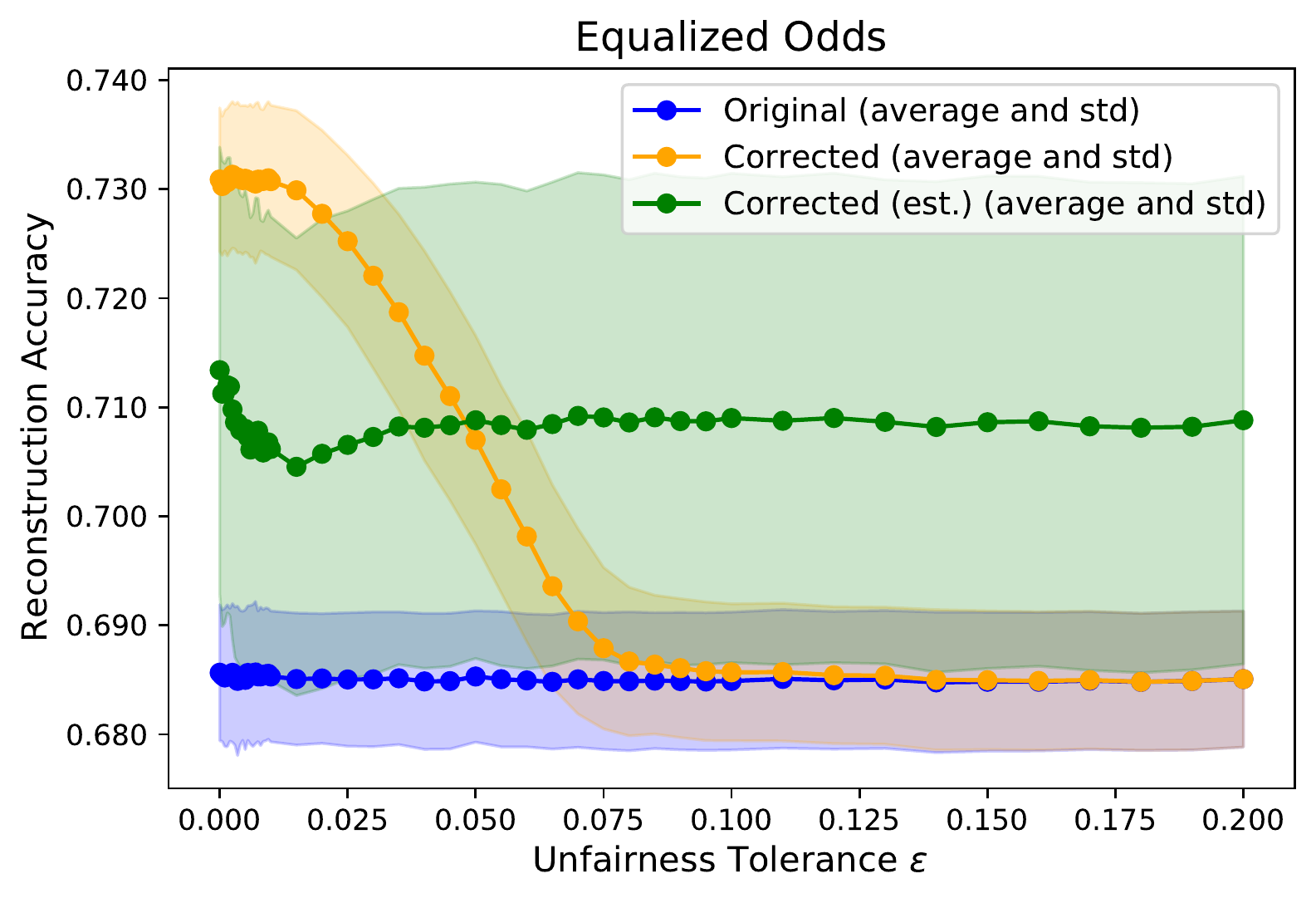}
    \end{center}
    \caption{Original (attacker $\attackerseven$), corrected (from actual fairness constraint, and from estimated one (\texttt{est.})) reconstruction quality, for our experiments using the ACSIncome dataset}
\label{fig:acsincome_results_inproc_protection}
\end{figure*}

 \begin{figure*}[htb]
    \begin{center}
    \includegraphics[width=\figwidth\textwidth]{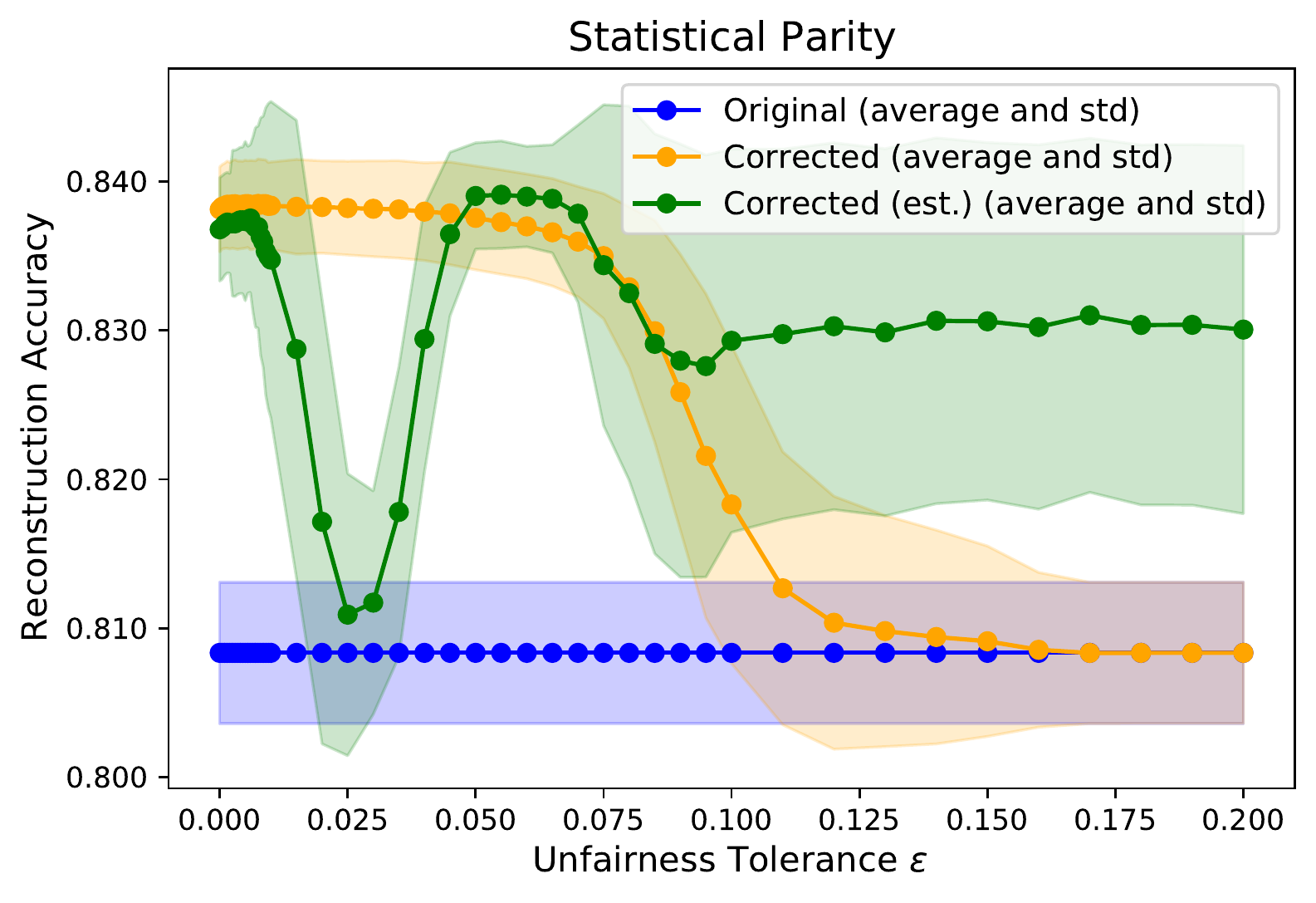} 
   \includegraphics[width=\figwidth\textwidth]{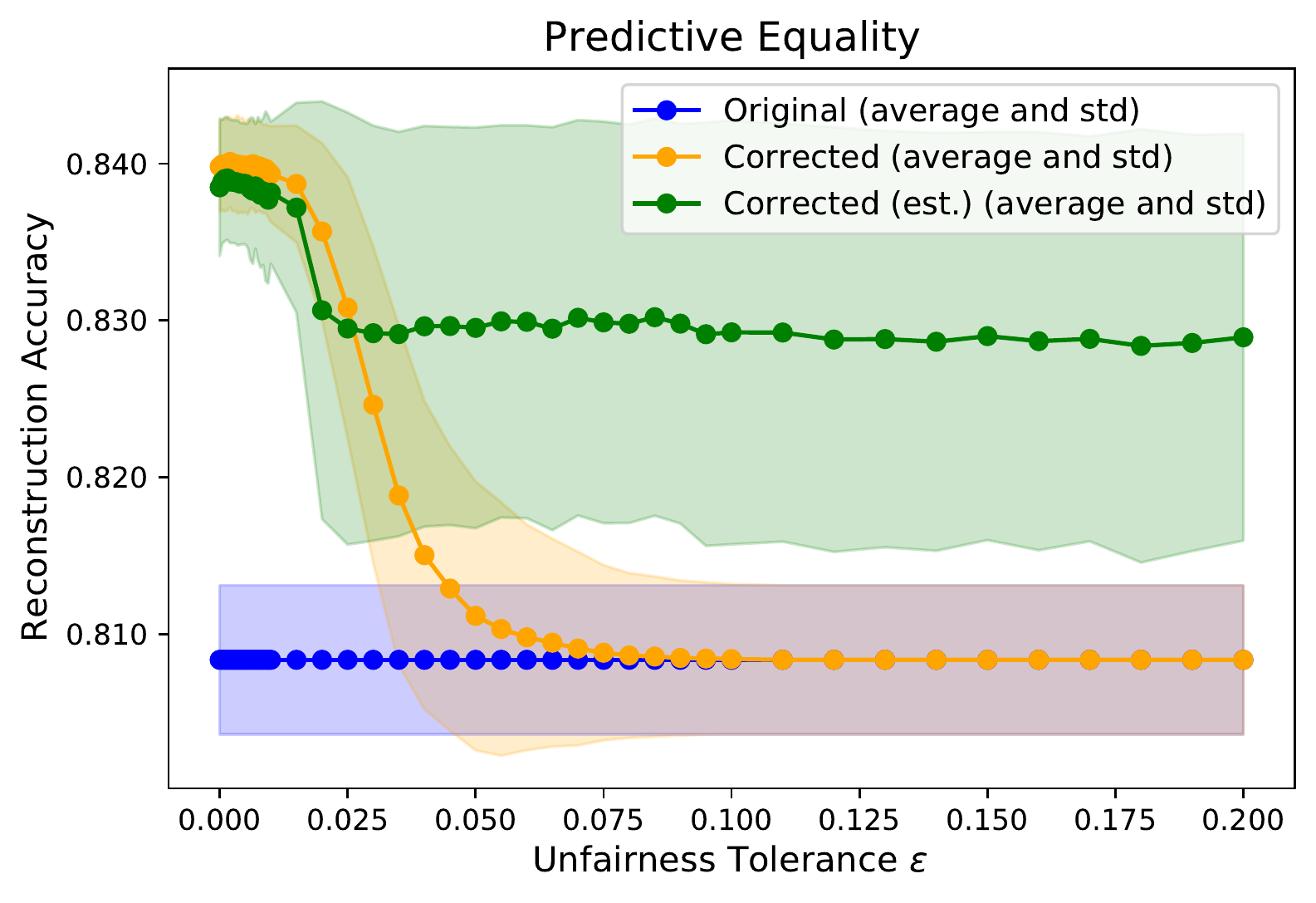}
    \includegraphics[width=\figwidth\textwidth]{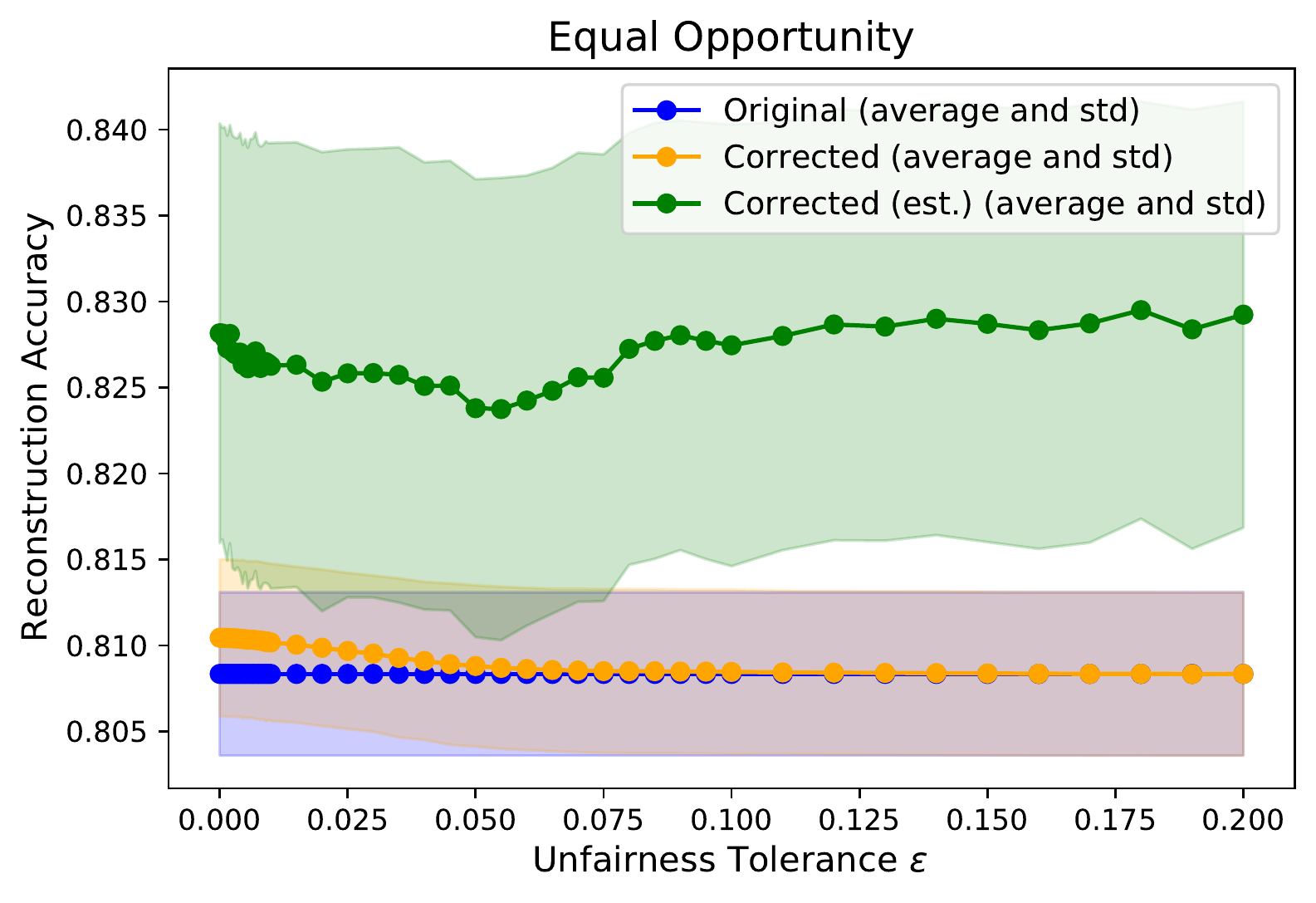}
    \includegraphics[width=\figwidth\textwidth]{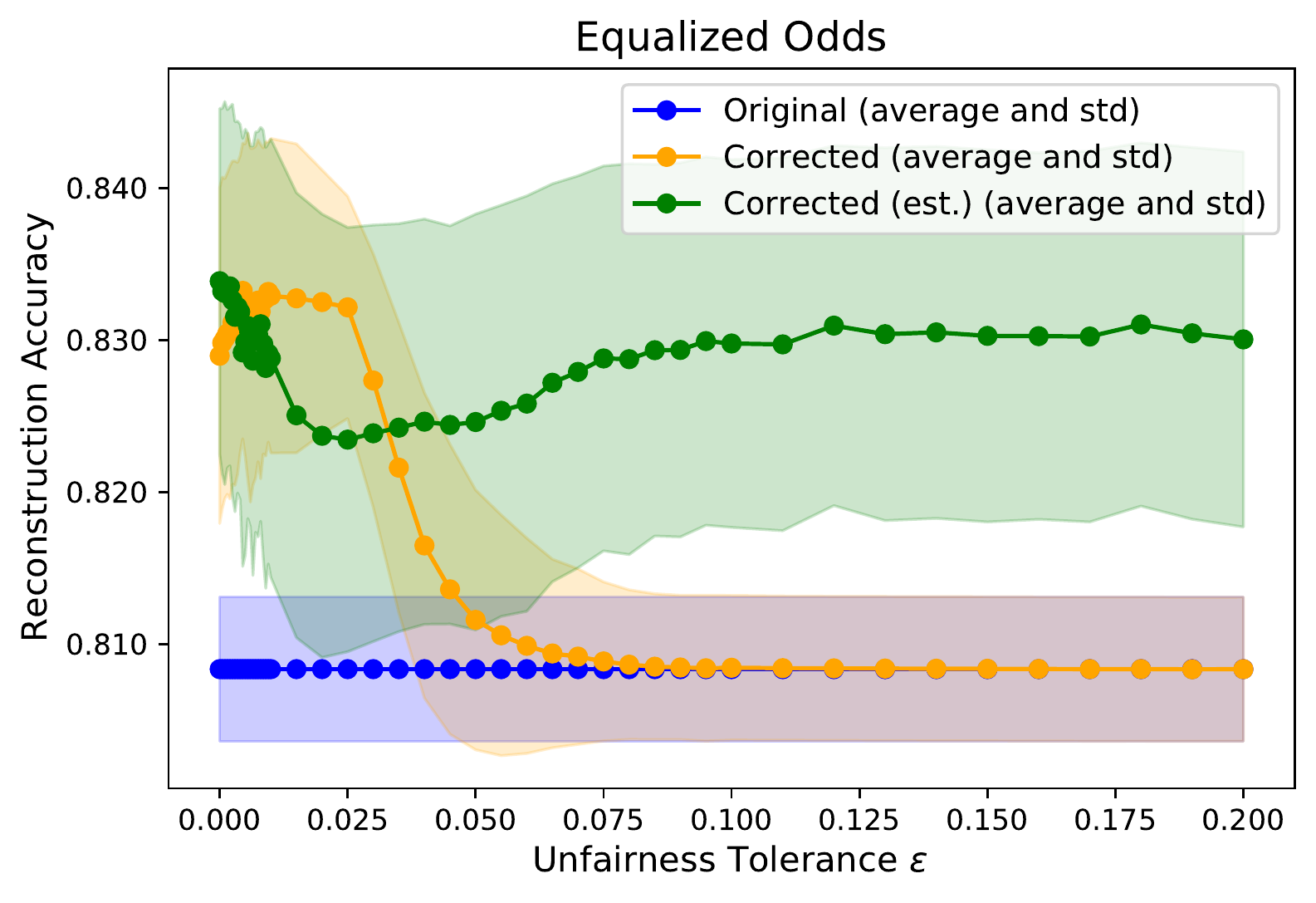}
    \end{center}
    \caption{Original (attacker $\attackersix$), corrected (from actual fairness constraint, and from estimated one (\texttt{est.})) reconstruction quality, for our experiments using the UCI Adult Income dataset}
\label{fig:adult_results_inproc_protection_attacker6}
\end{figure*}

 \begin{figure*}[htb]
    \begin{center}
    \includegraphics[width=\figwidth\textwidth]{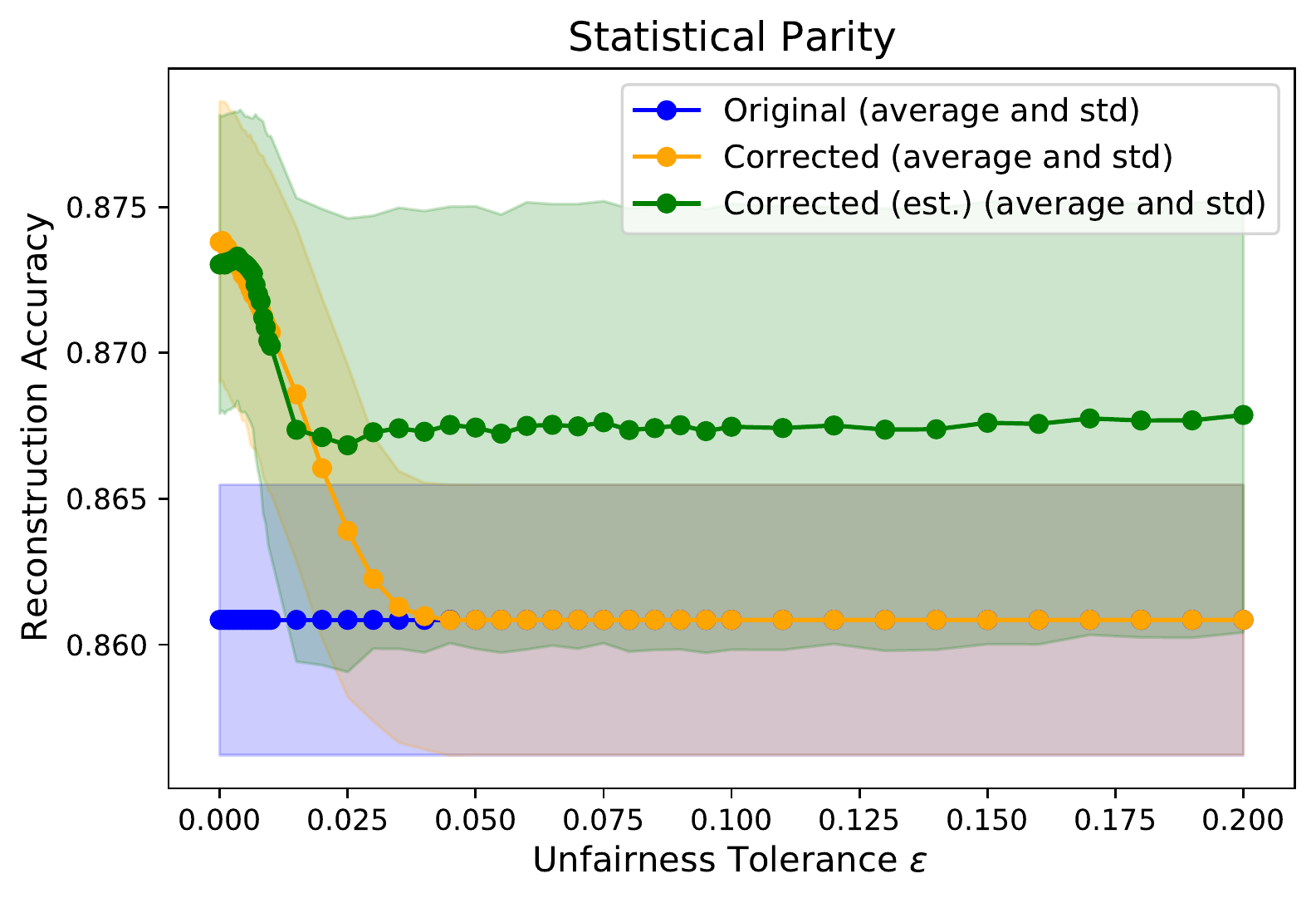} 
   \includegraphics[width=\figwidth\textwidth]{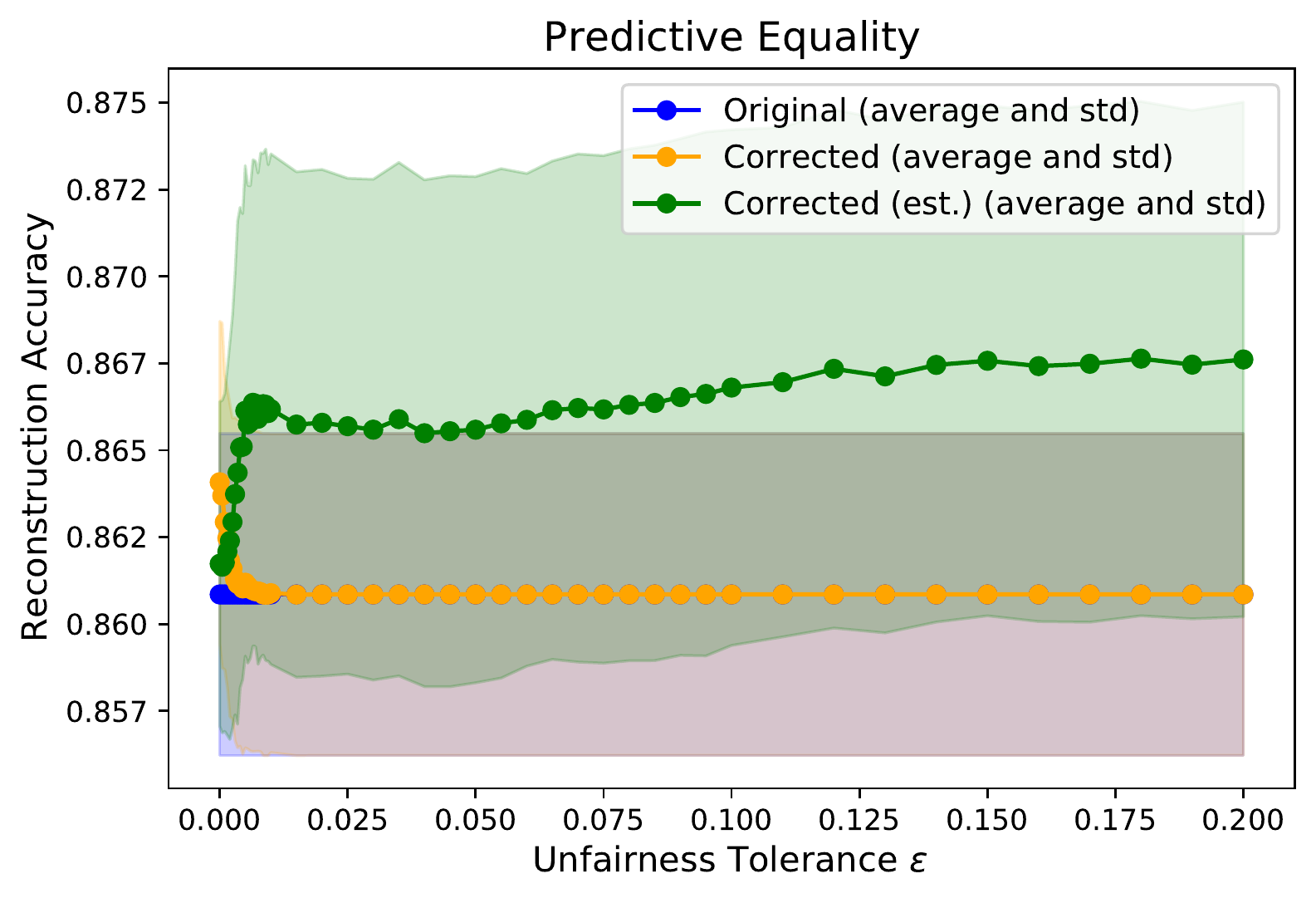}
    \includegraphics[width=\figwidth\textwidth]{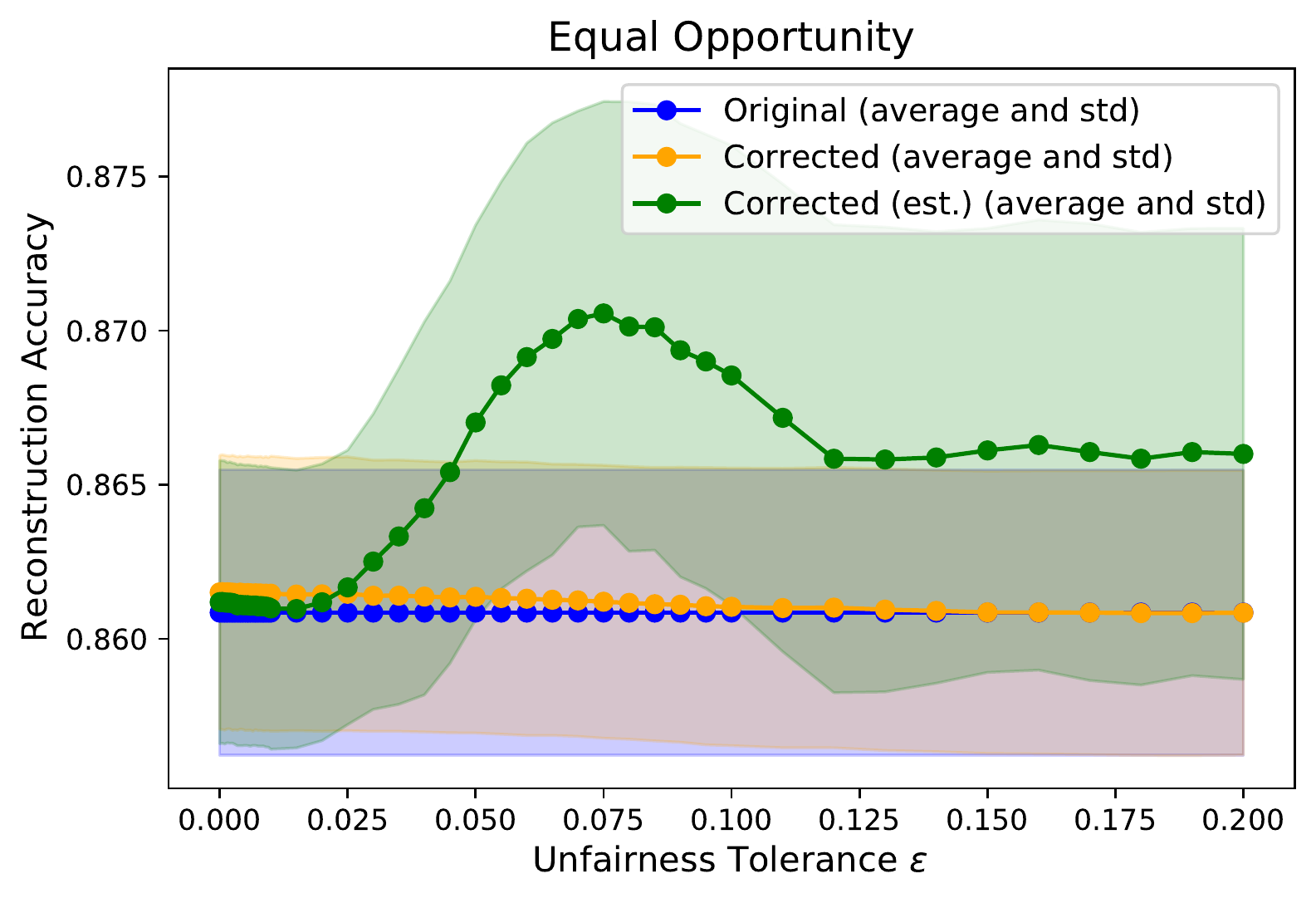}
    \includegraphics[width=\figwidth\textwidth]{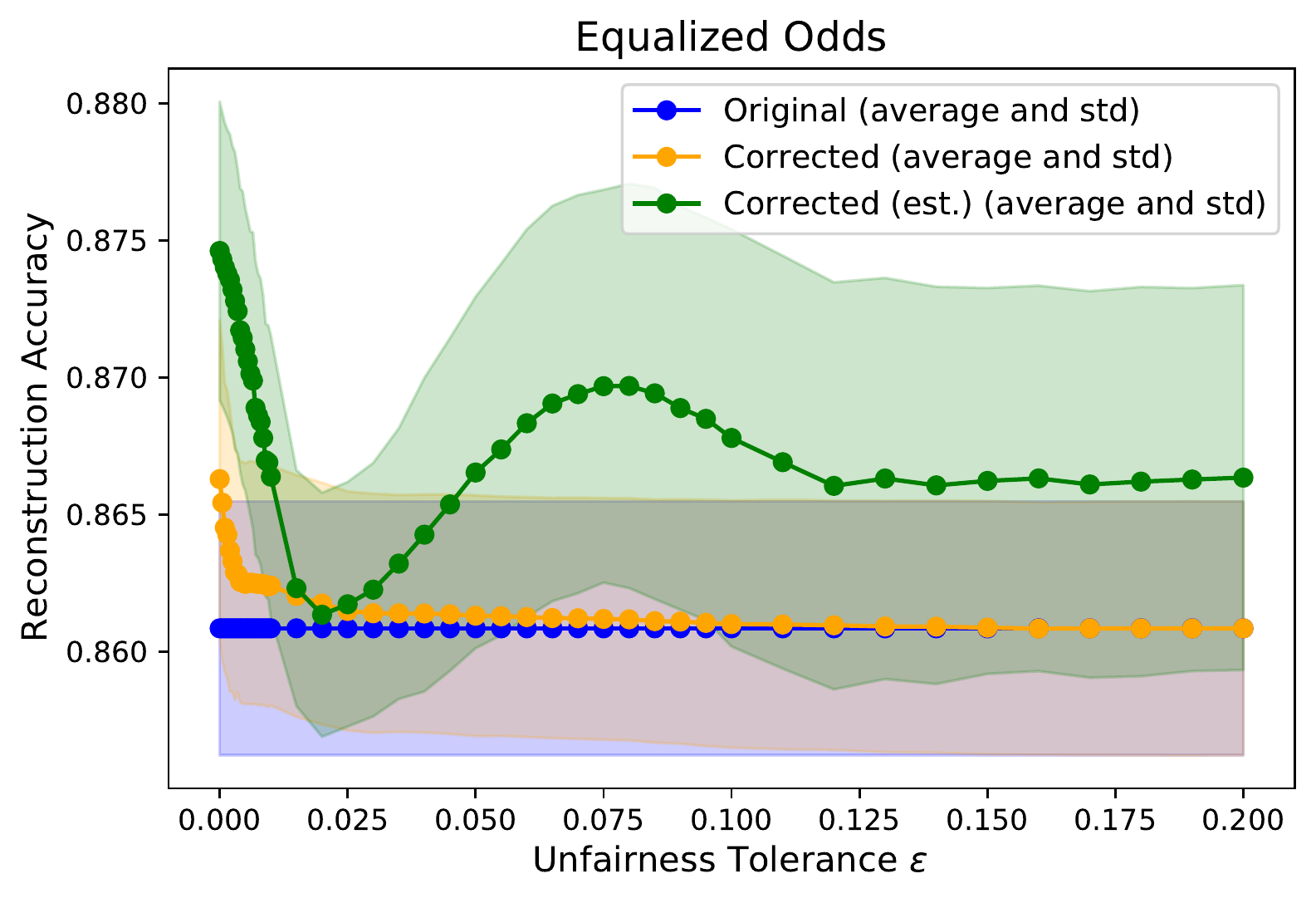}
    \end{center}
    \caption{Original (attacker $\attackersix$), corrected (from actual fairness constraint, and from estimated one (\texttt{est.})) reconstruction quality, for our experiments using the ACSPublicCoverage dataset}
\label{fig:acspubliccoverage_results_inproc_protection_attacker6}
\end{figure*}

 \begin{figure*}[htb]
    \begin{center}
    \includegraphics[width=\figwidth\textwidth]{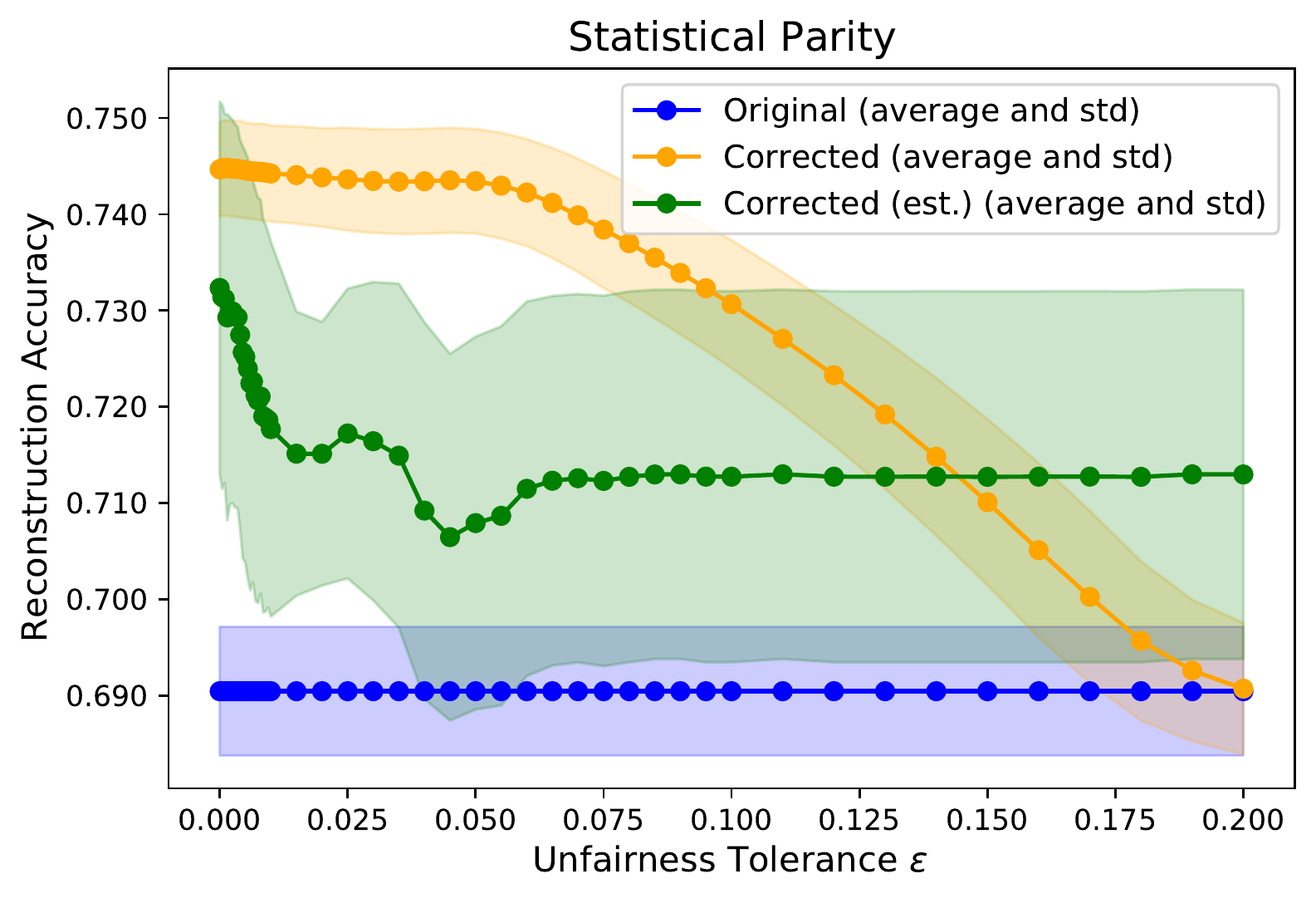} 
   \includegraphics[width=\figwidth\textwidth]{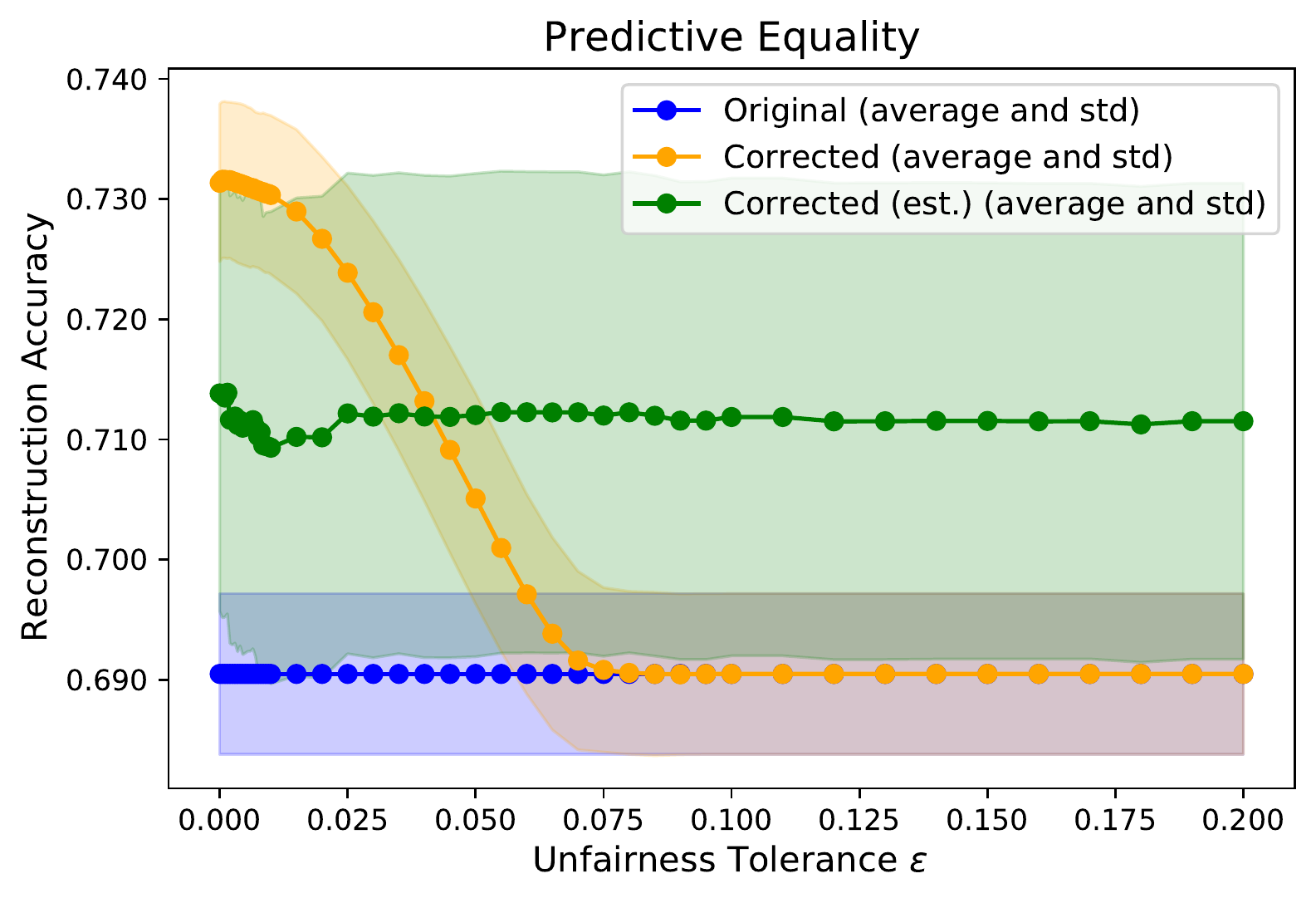}
    \includegraphics[width=\figwidth\textwidth]{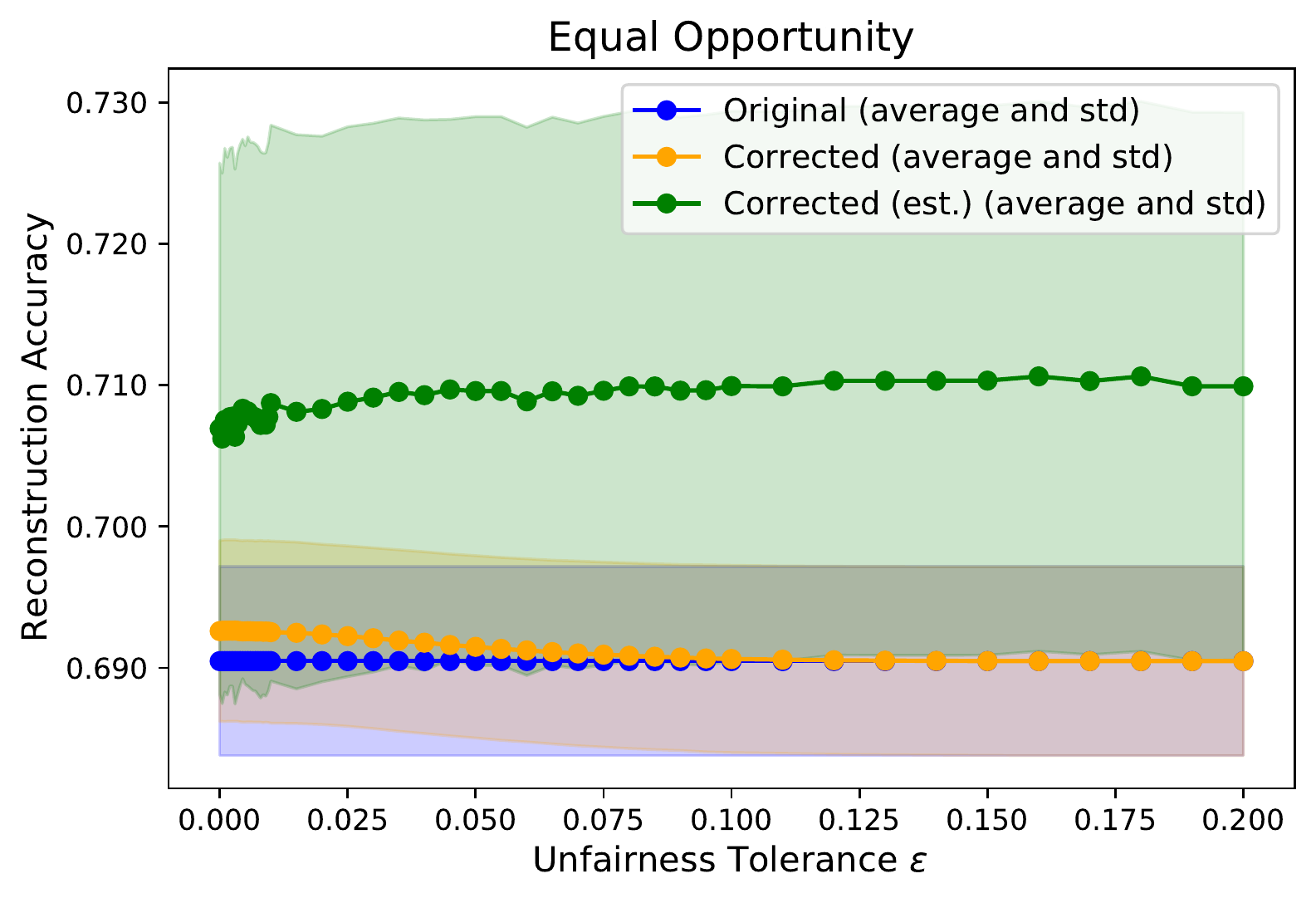}
    \includegraphics[width=\figwidth\textwidth]{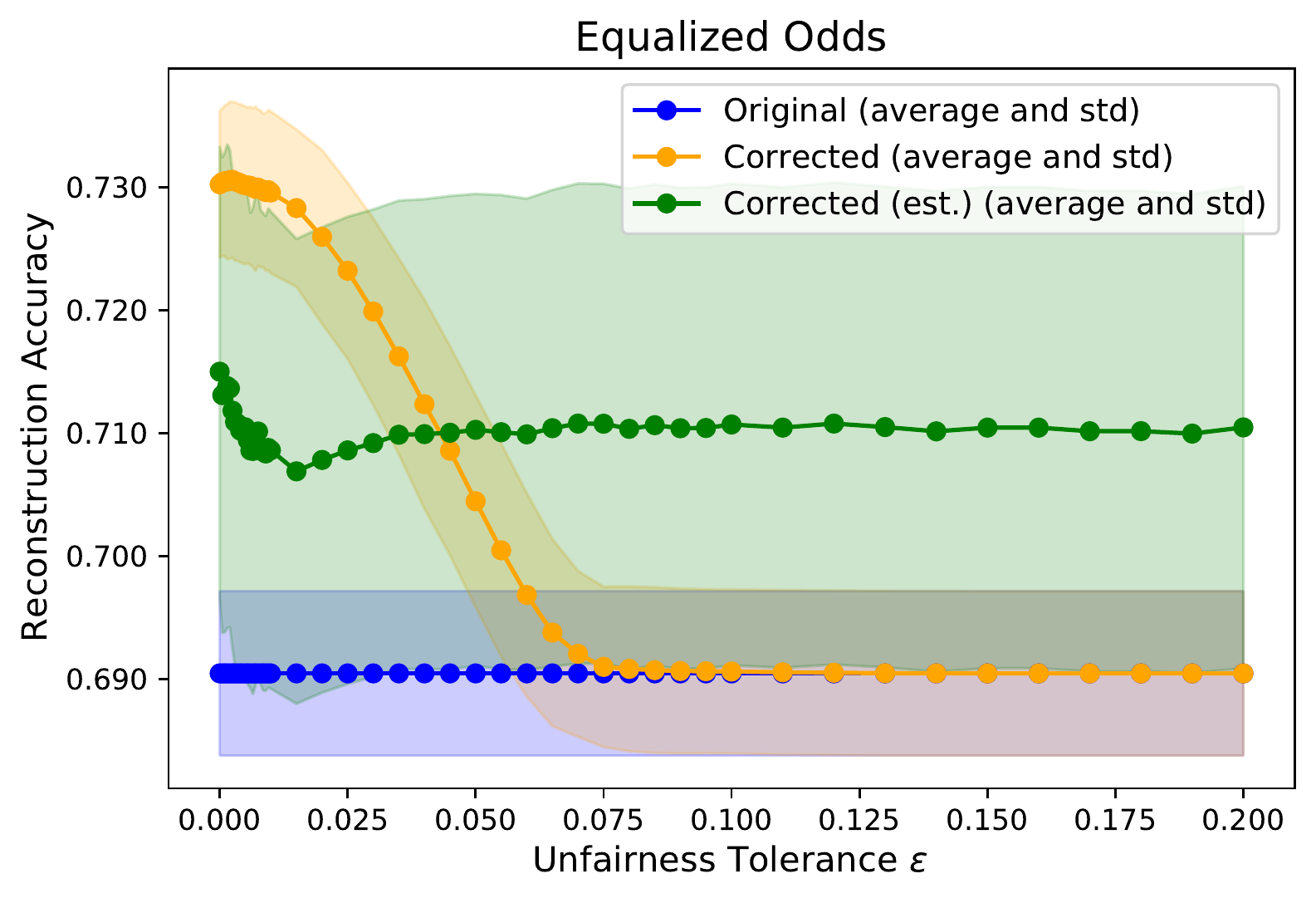}
    \end{center}
    \caption{Original (attacker $\attackersix$), corrected (from actual fairness constraint, and from estimated one (\texttt{est.})) reconstruction quality, for our experiments using the ACSIncome dataset}
\label{fig:acsincome_results_inproc_protection_attacker6}
\end{figure*}

\section{Additional Experiment: Reconstruction Performances Using a Pre-Processing Method for Fairness} \label{appendix:preproc_expes}

In this appendix section, we provide results for additional experiments using a pre-processing method for fairness: the CorrelationRemover method, implemented in the \texttt{Fairlearn} library~\cite{bird2020fairlearn}. 
In a nutshell, the CorrelationRemover transforms the training set unsensitive attributes in order to remove their correlations with the sensitive ones. 
A traditional machine learning algorithm is then used on the sanitized data (pre-processed unsensitive attributes) to produce a fair model. 

The CorrelationRemover does not optimize statistical fairness metrics explicitly. 
Indeed, bias against sensitive attributes is removed before training the model, in the data pre-processing step. 
Hence, in order to perform sensitive attributes reconstruction correction, one has to infer some fairness information. 
To do so, we use the strategy described in section~\ref{subsec:hiding_fairness}: the attacker measures the target model's unfairness on its own attack set, and chooses the metric with the smallest value. 
The experimental setup is similar to that of section~\ref{sec:expes_setup}. 
However, because the CorrelationRemover method does not optimize a particular fairness metric nor a particular tolerance value, we only perform one experiment for each dataset (repeated 100 times with different random seeds). 

The results presented in Table~\ref{tab:preproc} show that even in this context, the reconstruction correction step still provides significant reconstruction accuracy improvements. 
In all situations, the attacker was able to infer a valid fairness constraint and to leverage it to improve the initial sensitive attributes reconstruction.

Finally, these additional experiments confirm that the type of fairness intervention does not influence the performances of our proposed reconstruction correction step. 
The key factor for allowing reconstruction correction is that the predictions of the target model should be more fair than the original data. 
In this situation, the original attacker's reconstruction will likely be more biased than the (fair) target model's predictions, which will allow some reconstruction correction.

\end{document}